\documentclass[preprint,3p,times,twocolumn,sort&compress]{elsarticle}

\usepackage[modulo,switch]{lineno}

\usepackage{amsmath,graphicx}
\usepackage{amssymb,amsfonts,bm}
\usepackage[export]{adjustbox} 
\usepackage[ruled,vlined]{algorithm2e}
\usepackage{hyperref}
\usepackage[caption=false]{subfig}
\usepackage{color,soul}
\usepackage[table,xcdraw]{xcolor}
\usepackage{multirow}
\usepackage{siunitx}
\usepackage[utf8]{inputenc}
\usepackage[strings]{underscore} 
\usepackage[english]{babel} 
\usepackage{duckuments} 

\usepackage{pifont}
\newcommand{\cmark}{\text{\ding{51}}} 
\newcommand{\xmark}{\text{\ding{55}}} 

\usepackage[para]{threeparttable}

\usepackage{floatrow}
\usepackage[inline]{enumitem}
\newlist{inlinelist}{enumerate*}{1} 
\setlist[inlinelist]{label=(\roman*)}

\setlist[itemize]{nosep} 

\def\x{{\mathbf x}}

\DeclareMathOperator*{\argmin}{arg\,min}

\journal{Computer Methods and Programs in Biomedicine}

\begin{document}

\begin{frontmatter}



\title{MERA: Multimodal and Multiscale Self-Explanatory Model with Considerably Reduced Annotation for Lung Nodule Diagnosis}


\author[inst1,inst2]{Jiahao Lu\corref{cor1}}
\ead{lu@di.ku.dk}
\cortext[cor1]{Corresponding author}
\fntext[]{Postal address: Universitetsparken 1, København Ø, 2100, DK}

\author[inst3]{Chong Yin}
\author[inst2]{Silvia Ingala}
\author[inst1]{Kenny Erleben}
\author[inst2]{Michael Bachmann Nielsen}
\author[inst1]{Sune Darkner}

\affiliation[inst1]{organization={Department of Computer Science, University of Copenhagen},
            country={Denmark}}
\affiliation[inst2]{organization={Department of Diagnostic Radiology, Rigshospitalet, Copenhagen University Hospital},
            country={Denmark}}
\affiliation[inst3]{organization={Department of Computer Science, Hong Kong Baptist University},
            country={China}}

\begin{abstract}
\textbf{Background and objective:}
Lung cancer, a leading cause of cancer-related deaths globally, emphasizes the importance of early detection for improving patient outcomes. Pulmonary nodules, often early indicators of lung cancer, necessitate accurate and timely diagnosis. 
Despite advances in Explainable Artificial Intelligence (XAI), many existing systems struggle to provide clear, comprehensive explanations, especially in scenarios with limited labelled data. 
This study introduces MERA, a \textbf{M}ultimodal and \textbf{M}ultiscale self-\textbf{E}xplanatory model designed for lung nodule diagnosis with considerably \textbf{R}educed \textbf{A}nnotation requirements.

\textbf{Method:}
MERA integrates unsupervised and weakly supervised learning strategies, including self-supervised learning techniques and Vision Transformer architecture for unsupervised feature extraction, followed by a hierarchical prediction mechanism that leverages sparse annotations through semi-supervised active learning in the learned latent space.  
MERA explains its decisions on multiple levels: model-level global explanations through semantic latent space clustering, instance-level case-based explanations providing similar instances, local visual explanations via attention maps, and concept explanations based on critical nodule attributes. 

\textbf{Results:}
Evaluations on the public LIDC dataset underscore MERA’s superior diagnostic accuracy and self-explainability. 
With only $1\%$ of annotated samples, MERA achieves diagnostic accuracy comparable to or exceeding that of state-of-the-art methods that require full annotation. 
The model's inherent design delivers comprehensive, robust, multilevel explanations that align closely with clinical practices, thereby enhancing the trustworthiness and transparency of the diagnostic process.

\textbf{Conclusions:}
MERA addresses critical gaps in XAI and lung nodule diagnosis by providing a self-explanatory framework with multimodal and multiscale explanations and significantly reduced annotation needs. 
The demonstrated viability of unsupervised and weakly supervised learning in this context lowers the barrier to deploying diagnostic AI systems in broader medical domains. 
MERA represents a significant step toward more transparent, understandable, and trustworthy AI systems in healthcare.

Our complete code is open-source available: \url{https://github.com/diku-dk/credanno}.
\end{abstract}


\end{frontmatter}



\section{Introduction}
\label{sec:intro}

\begin{figure*}[tbp]
    \centering
    \includegraphics[width=\textwidth]{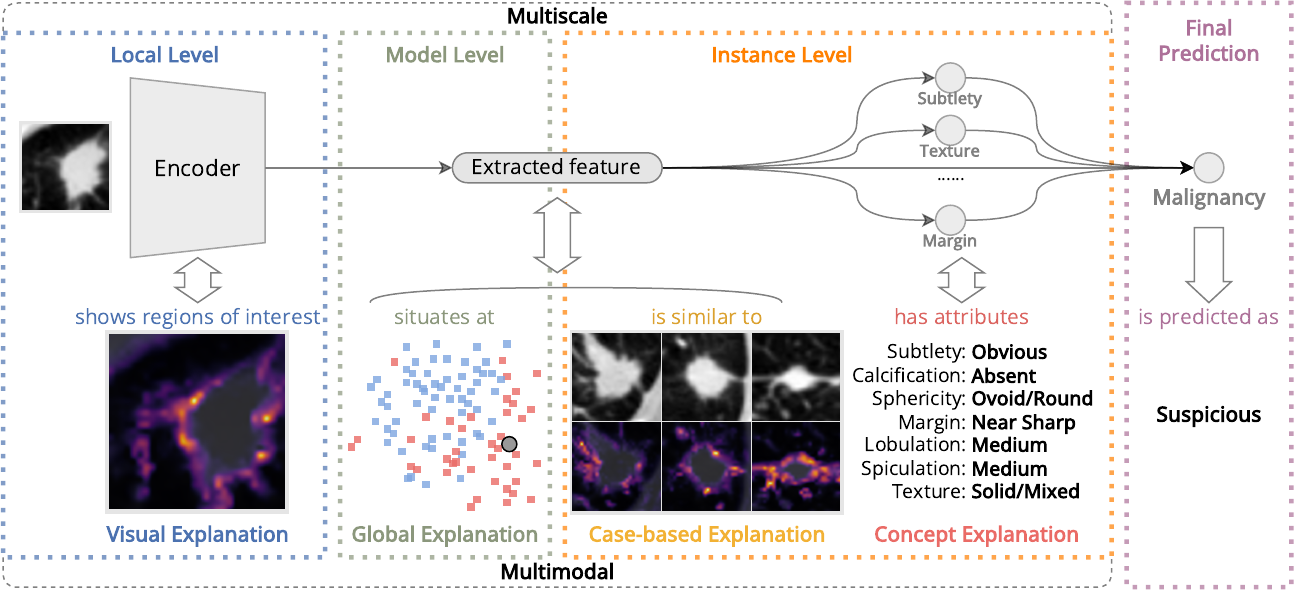}

    \caption{\textbf{Multimodal and multiscale explanations as an intrinsic driving force of the decision}: local visual explanations via attention maps, model-level global explanations through semantic latent space clustering, instance-level case-based explanations providing similar instances, and concept explanations based on critical nodule attributes.
    }
    \label{fig:intro}
\end{figure*}

Lung cancer remains a leading cause of cancer-related deaths worldwide, with early detection being critical for improving patient outcomes \cite{cielloMissedLungCancer2017}. Pulmonary nodules, which are small masses of tissue in the lungs, are often the first indicators of potential lung cancer. Accurate and timely diagnosis of these nodules can significantly impact the prognosis and treatment strategies for affected individuals \cite{vlahosLungCancerScreening2018, mazzoneEvaluatingPatientPulmonary2022}.
Lung nodule diagnosis is a critical step in the early detection and treatment of lung cancer \cite{vlahosLungCancerScreening2018, mazzoneEvaluatingPatientPulmonary2022}, which remains a leading cause of cancer-related deaths worldwide \cite{cielloMissedLungCancer2017}.
In this domain, not only the high prediction accuracy but also transparency in decision-making processes aligning with diagnostic practices are of paramount importance for clinical acceptance and trust.
Despite notable advances in Explainable Artificial Intelligence (XAI) \cite{shenInterpretableDeepHierarchical2019, lalondeEncodingVisualAttributes2020, chenEndtoEndMultiTaskLearning2021, liuMultiTaskDeepModel2020, joshiLungNoduleMalignancy2021}, many existing XAI systems struggle to provide clear, comprehensive multi-level explanations \cite{rudinStopExplainingBlack2019, vanderveldenExplainableArtificialIntelligence2022}, particularly in scenarios where labelled data is scarce or even absent \cite{salahuddinTransparencyDeepNeural2022}.

This work introduces MERA, a \textbf{M}ultimodal and \textbf{M}ultiscale self-\textbf{E}xplanatory model for lung nodule diagnosis with considerably \textbf{R}educed \textbf{A}nnotation needs. 
MERA is designed to be primarily unsupervised, requiring little annotated data, and stands out through its capacity to elucidate the reasoning behind its decisions in a multimodal and multiscale manner (illustrated in Fig.~\ref{fig:intro}):
\begin{itemize}[label=\textbullet]
    \item Model-level Global Explanations: MERA's decision procedure is designed to align with radiologists' guidelines \cite{macmahonGuidelinesManagementIncidental2017}. It learns a semantic latent space, where the clustering reveals underlying correlations between nodule attributes and malignancy, to support a global understanding of model decisions.
    \item Instance-level Case-based Explanations: By presenting cases most similar to the one under consideration, akin to methodologies like ProtoPNet \cite{chenThisLooksThat2019}, MERA aids radiologists by contextualising decisions with comparable instances.
    \item Local Visual Explanations: The model generates attention maps that provide visual explanations, highlighting the regions of interest in the imaging data that contribute to the diagnosis.
    \item Concept Explanations: MERA predicts a set of nodule attributes defined in consensus by radiologists \cite{macmahonGuidelinesManagementIncidental2017}, which are then used to inform the final malignancy prediction.
\end{itemize}
It is important to note that most of these explanatory mechanisms operate independently of annotations, with the sole exception being the concept explanation, which necessitates as little as $1\%$ of annotation.

Methodologically, MERA employs a sophisticated integration of unsupervised and weakly supervised learning strategies and original components. 
The model utilises self-supervised learning (SSL) techniques \cite{caronEmergingPropertiesSelfSupervised2021} and a Vision Transformer (ViT) architecture \cite{dosovitskiyImageWorth16x162020} for unsupervised feature extraction, followed by a hierarchical prediction mechanism that leverages sparse annotations through semi-supervised active learning \cite{wangCostEffectiveActiveLearning2017} in the learned latent space. 
This approach not only exploits the use of available data and annotations, 
but also empowers the model with explainability across the outlined levels.

Extensive evaluations on the public LIDC dataset demonstrate MERA's effectiveness and superiority over existing methods in diagnostic accuracy and explainability. 
It achieves comparable or better diagnostic accuracy using only $1\%$ annotated samples compared to state-of-the-art methods requiring full annotation. 
The model’s intrinsic design provides comprehensive, robust, multilevel explanations that align closely with clinical practices, significantly enhancing the trustworthiness and transparency of the diagnostic process. 

Our contributions are twofold: First, we address a critical gap between XAI and lung nodule diagnosis by proposing a self-explanatory framework that offers multimodal and multiscale explanations with considerably limited annotation. Second, we demonstrate the feasibility of unsupervised and weakly supervised learning in this context, significantly lowering the barrier to entry for deploying diagnostic AI systems in broader medical domains. MERA represents a significant step toward more transparent, understandable, and trustworthy AI systems in healthcare.

\section{Related work}
\label{sec:background}

\subsection{Lung nodule analysis}

Optimal treatment of an individual with a pulmonary nodule is crucial as it can lead to the early detection of cancer  \cite{mazzoneEvaluatingPatientPulmonary2022}. The relevance of lung nodule analysis using machine learning methods lies in its potential to enhance the accuracy and efficiency of early lung cancer diagnosis. Deep learning methods, in particular, have demonstrated a superior capability to automatically extract features from limited annotated datasets, significantly improving lung nodule detection and classification \cite{liangPerformanceDeepLearningSolutions2023, balciSeriesBasedDeepLearning2023}.

The inherent demand for transparency and comprehensibility in medical diagnostics, especially in lung nodules analysis, makes the explainability of algorithms essential. Explainability is not merely a desirable attribute but a fundamental requirement to ensure clinical efficacy and user trust.

Features critical for the algorithms' decision-making are often emphasised in international guidelines \cite{macmahonGuidelinesManagementIncidental2017}, which highlight their significance in lung nodule classification. These include:
\begin{itemize}[label=\textbullet]
    \item \textbf{Malignancy}: Indicative of potential cancerous nature, assessed through size, growth patterns, and cellular abnormalities.
    \item \textbf{Subtlety}: The visibility of nodules in medical imaging, which can range from clearly visible to subtle and easily missed.
    \item \textbf{Calcification}: Calcium deposits within the nodule provide insights into its nature, with distinct patterns suggesting different conditions.
    \item \textbf{Sphericity} and \textbf{Margin}: These factors aid in assessing the overall shape, where a spherical shape is often linked to benign conditions, and irregular margins may indicate malignancy.
    \item \textbf{Lobulation}: Rounded projections on the nodule's periphery, suggesting a potentially aggressive nature.
    \item \textbf{Spiculation}: Spicules extending from the nodule, associated with a higher risk of malignancy due to their infiltrative nature.
    \item \textbf{Texture}: Internal composition variations, offering additional diagnostic clues.
\end{itemize}

Analysing these characteristics collectively enables healthcare professionals to develop a detailed and nuanced profile for each lung nodule, facilitating accurate classification and informed decision-making in patient care. This multi-faceted approach to nodule assessment underscores the complexity of lung nodule characterisation and highlights the need for a thorough understanding of each characteristic to ensure precise and reliable diagnostic outcomes.

Continuous advancements have been made using the public LIDC dataset \cite{armatoLungImageDatabase2011} to predict these attributes along with malignancy classification. 
Notable efforts include:
HSCNN\cite{shenInterpretableDeepHierarchical2019} proposed a hierarchical network using 3D volume data. 
MTMR\cite{liuMultiTaskDeepModel2020} and MSN-JCN\cite{chenEndtoEndMultiTaskLearning2021} approached this task as a multitask learning problem, where MTMR\cite{liuMultiTaskDeepModel2020} also used all 2D slices in 3D volumes as input, and MSN-JCN\cite{chenEndtoEndMultiTaskLearning2021} even used additional segmentation masks, annotated diameter information, as well as two other traditional methods to assist training. 
X-caps\cite{lalondeEncodingVisualAttributes2020} is the most similar work as ours, but as above, still relies heavily on annotations.
WeakSup\cite{joshiLungNoduleMalignancy2021} attempted to reduce the dependency on feature annotations to $25\%$ through multi-stage training and multi-scale 3D volume data, although its performance degrades with further reductions in annotations. 
In contrast, our methodology not only preserves robust performance with a minimal annotation requirement of only $1\%$ for both malignancy classification and feature annotations, but it also introduces additional forms of explainability absent in previous models.

\subsection{Interpretable methods}

\subsubsection{Visual explanations.}
In the domain of Explainable Artificial Intelligence (XAI), visual explanations have emerged as a pivotal approach for elucidating the decision-making processes of deep learning models, particularly in critical applications such as medical image analysis. Traditional methodologies predominantly leverage post-hoc techniques to generate attribution maps or saliency maps, which highlight regions of an input image that are most influential for the model's prediction. These techniques, including activation-based methods such as Class Activation Mapping (CAM) \cite{zhouLearningDeepFeatures2016} and Score-CAM \cite{wangScoreCAMScoreWeightedVisual2020}, as well as gradient-based methods (Grad-CAM \cite{selvarajuGradCAMVisualExplanations2019}, Grad-CAM++ \cite{chattopadhayGradCAMGeneralizedGradientBased2018}, Smoothed Grad-CAM++ \cite{omeizaSmoothGradCAMEnhanced2019}, XGrad-CAM \cite{fuAxiombasedGradCAMAccurate2020}), and Layer-wise Relevance Propagation (LRP) \cite{bachPixelWiseExplanationsNonLinear2015, bohleLayerWiseRelevancePropagation2019}, offer insights after the model's decision has been made, serving as a tool for interpreting complex models' behaviours retrospectively. However, these post-hoc explanations are often criticised for their lack of transparency and potential to mislead by highlighting irrelevant features or multiple classes can have the same regions of interest \cite{rudinStopExplainingBlack2019, salahuddinTransparencyDeepNeural2022}. These methods, while useful, do not inherently integrate interpretability into the decision-making process itself.

Our method distinguishes itself from existing strategies by leveraging the model's inherent mechanisms to generate visual explanations that are integral to the decision-making process, rather than retrospectively interpreting these decisions. This is in contrast to traditional attribution methods.

By integrating visual explanations directly into the decision-making process, our approach aligns with the clinical workflow, enabling practitioners to make informed decisions based on transparent and interpretable AI-driven insights. This paradigm shift addresses the critical need for trust and reliability in AI applications for lung nodule diagnosis, offering a pathway to more widespread adoption and acceptance of AI tools in medical diagnostics.

\subsubsection{Concept explanations.}
Concept learning models, also known as self-explanatory methods, function by initially predicting high-level clinical concepts, and then leverage them to inform the final classification decision \cite{kohConceptBottleneckModels2020, stammerRightRightConcept2021}. This approach is especially valuable in medical applications where the reliability of AI is improved by explicitly incorporating clinically relevant concepts into the decision-making pathway, and therefore gaining favour in contrast with posthoc approaches attempting to explain ``black box" models \cite{rudinStopExplainingBlack2019}. However, a critical limitation of this approach is the additional burden of annotation cost. This is particularly significant in medical applications where expert annotations are costly and time-consuming to obtain. 

In contrast, our work addresses this critical limitations by integrating self-supervised learning (SSL) techniques to considerably reduce the dependency on annotated data. By doing so, we mitigate the annotation burden commonly associated with self-explainable models, thus enhancing the practical applicability of our approach for lung nodule diagnosis. This innovation not only aligns with the foundational goals of XAI but also pushes the boundaries of what is achievable in the context of medical image analysis with very limited annotations.

\subsubsection{Case-based explanations.}
Case-based models, fundamentally grounded in the comparison of new instances with class discriminative prototypes, offer a transparent mechanism for decision-making and represent a significant stride towards self-explainability. Specifically, these models operate by learning discriminative prototypes and then performing classification by comparing the extracted features from input images against these prototypes \cite{chenThisLooksThat2019}. 
Despite their explainability advantages and successful usage in medical field such as Alzheimer’s Disease Classification from MRI \cite{mohammadjafariUsingProtoPNetInterpretable2021} and diagnosis in chest X-ray images \cite{kimXProtoNetDiagnosisChest2021, singhTheseNotLook2021}, case-based models are notably susceptible to corruption by noise and compression artefacts, which can significantly impair their performance \cite{hoffmannThisLooksThat2021}. Moreover, the training process for these models is notably challenging, given the intricacy of learning discriminative and robust prototypes, as highlighted by the scarcity of available techniques for their effective training \cite{rudinStopExplainingBlack2019}.

Our research introduces an innovative approach by leveraging SSL to learn a semantically meaningful latent space. This advancement enables our model to identify and present the most similar samples as references to the users, providing an additional layer of interpretability without directly influencing the prediction process. By doing so, we address the challenges associated with traditional case-based models, notably the difficulties in training and the susceptibility to noise and artifacts, thereby enhancing the practical utility and robustness of our approach. Importantly, our work stands distinct in its ability to reduce reliance on extensive labeled datasets, a common challenge in deploying case-based reasoning systems, especially in domains requiring significant expert knowledge, such as medical diagnosis.


\subsection{Contrastive learning}
In recent years, contrastive learning \cite{heMomentumContrastUnsupervised2020, chenExploringSimpleSiamese2021, grillBootstrapYourOwn2020, caronEmergingPropertiesSelfSupervised2021, chenSimpleFrameworkContrastive2020} has shown great potential for self-supervised learning in computer vision area. Efforts have been made to extend its success towards the domain of medical images \cite{sowrirajanMoCoPretrainingImproves2021, vuMedAugContrastiveLearning2021}.

\subsection{Vision Transformers (ViT)}
Vision Transformers\cite{dosovitskiyImageWorth16x162020} has significantly advanced the frontier of computer vision. However, its impressive performance is shown to be heavily dependent on large-scale data and high-capacity models -- even the ImageNet dataset\cite{russakovskyImageNetLargeScale2015} with $1$ M samples is not enough to unleash its full potential, let alone the medical image domain where such large-scale data is rare, as well as annotations.
Recent works \cite{chenEmpiricalStudyTraining2021, caronEmergingPropertiesSelfSupervised2021} show the successful combination of ViT and contrastive learning in the self-supervised paradigm. 
In this work, we demonstrate a successful usage of ViT with only hundreds of training samples.

\section{Method}
\label{sec:method}

\begin{figure}[tbp]
    \centering
    \includegraphics[width=\textwidth]{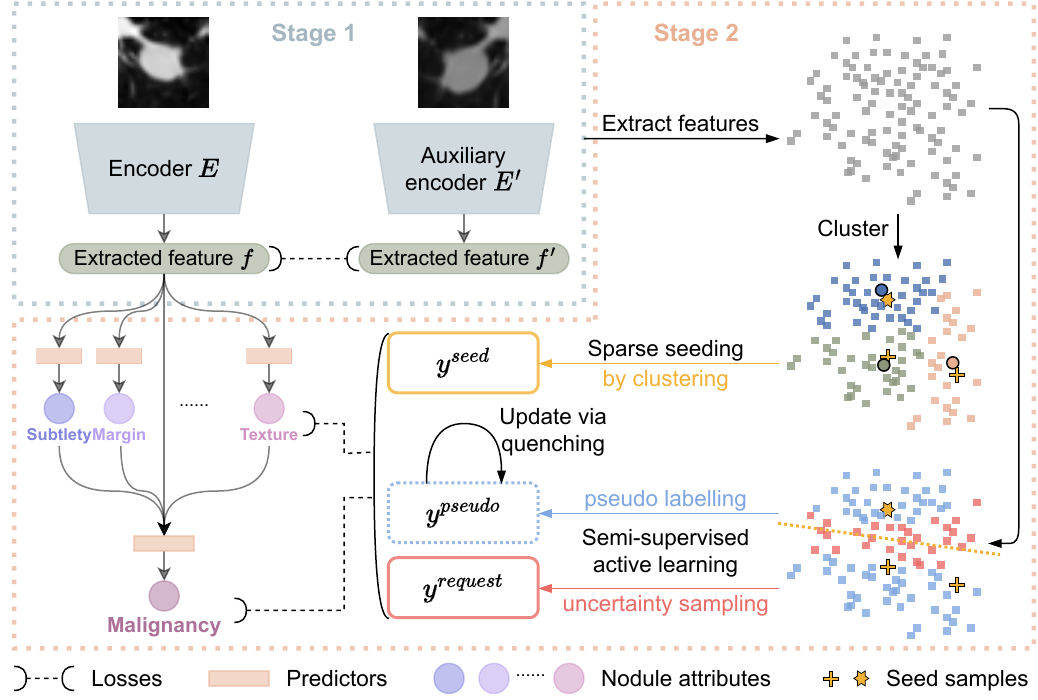}

    \caption{\textbf{Method overview of the data-/annotation-efficient training.}
    In Stage 1, an encoder is trained using self-supervised contrastive learning to map the input nodule images to a semantically meaningful latent space.
    In Stage 2, the proposed annotation exploitation mechanism conducts semi-supervised active learning with sparse seeding and training quenching in the learned space, to jointly exploit the extracted features, annotations, and unlabelled data.
    }
    \label{fig:overview}
\end{figure}

As illustrated in Fig.~\ref{fig:overview}, the proposed approach only requires a relatively small amount of unlabelled data and down to 1\% of their annotation in the two-stage training. In the inference phrase, in contrast to post-hoc explanations, comprehensive multimodal and multiscale intrinsic explanations are generated in prior to the model's predictions. It is worth mentioning that some of the explanations can be generated even without any annotation in training.

\subsection{Data-/annotation-efficient training}

The proposed training consists two stages: unsupervised training of the feature encoder and weakly supervised training to predict malignancy with human-interpretable nodule attributes as concept explanations.
As a result, most of the parameters are learned during the first stage in a self-supervised manner. Therefore, in the second stage, only few annotations are needed to train the predictors.

\subsubsection{Stage1: Unsupervised feature extraction}
\label{sec:method_recap}

In Stage 1, the majority of parameters are trained using self-supervised contrastive learning as an encoder to map the input nodule images to a latent space that complies with radiologists' reasoning for nodule malignancy. 

Due to the outstanding results exhibited by DINO \cite{caronEmergingPropertiesSelfSupervised2021}, we adopt their framework for unsupervised feature extraction, which trains 
\begin{inlinelist}
    \item a primary branch $\{E, H\}_{\theta_\text{pri}}$, composed by a feature encoder $E$ and a multi-layer perceptron (MLP) prediction head $H$, parameterised by $\theta_\text{pri}$;
    \item an auxiliary branch $\{E, H\}_{\theta_\text{aux}}$, which is of the same architecture as the primary branch, while parameterised by $\theta_\text{aux}$.
\end{inlinelist}
After training only the primary encoder $E_{\theta^\text{E}_\text{pri}}$ is used for feature extraction.

The branches are trained using augmented image patches of different scales to grasp the core feature of a nodule sample.
For a given input image $x$, different augmented global views $V^g$ and local views $V^l$ are generated \cite{caronUnsupervisedLearningVisual2020}:
$x \rightarrow v \in V^g \cup V^l$.
The primary branch is only applied to the global views $v_\text{pri} \in V^g$, producing $K$ dimensional outputs $z_\text{pri}=E_{\theta^\text{E}_\text{pri}} \circ H_{\theta^\text{H}_\text{pri}} (v_\text{pri})$;  while the auxiliary branch is applied to all views $v_\text{aux} \in V^g \cup V^l$, producing outputs $z_\text{aux}=E_{\theta^\text{E}_\text{aux}} \circ H_{\theta^\text{H}_\text{aux}} (v_\text{aux})$ to predict $z_\text{pri}$.
To compute the loss, the output in each branch is passed through a Softmax function scaled by temperature $\tau_\text{pri}$ and $\tau_\text{aux}$:
$ p_\text{aux} = \texttt{softmax}(z_\text{aux} / \tau_\text{aux})$,\quad
$p_\text{pri} = \texttt{softmax}((z_\text{pri}-\mu) / \tau_\text{pri})$,
where a bias term $\mu$ is applied to $z_\text{pri}$ to avoid collapse\cite{caronEmergingPropertiesSelfSupervised2021}, and updated at the end of each iteration using the exponential moving average (EMA) of the mean value of a batch with batch size $N$ using momentum factor $\lambda \in [0, 1)$: $\mu \leftarrow \lambda \mu + (1 - \lambda) \frac{1}{N} \sum_{s=1}^{N} z_\text{pri}^{(s)}$.

The parameters $\theta_\text{aux}$ are learned by minimising the cross-entropy loss between the two branches via back-propagation \cite{heMomentumContrastUnsupervised2020}:
\begin{equation}
    \theta_\text{aux} \leftarrow 
    \arg\min_{\theta_\text{aux}} \sum_{v_\text{pri} \in V^g} \sum_{\substack{v_\text{aux} \in V^g \cup V^l \\ v_\text{aux} \neq v_\text{pri}}} \mathcal{L}\left( p_\text{pri}, p_\text{aux} \right) ,
\end{equation}
where $\mathcal{L}({p}_1, {p}_2) = - \sum_{c=1}^C {p}_1^{(c)} \log {p}_2^{(c)}$ for $C$ categories. 
The parameters $\theta_\text{pri}$ of the primary branch are updated by the EMA of the parameters $\theta_\text{aux}$ with momentum factor $m \in [0, 1)$:
\begin{equation}
    \theta_\text{pri} \leftarrow m\theta_\text{pri} + (1 - m)\theta_\text{aux} .
\end{equation}

In our implementation, the feature encoders $E$ use Vision Transformer (ViT)\cite{dosovitskiyImageWorth16x162020} as the backbone for their demonstrated ability to learn more generalisable features. 
The created views are image patches of size $16 \times 16$ and projected by a linear layer to get patch embeddings of $d = 384$ dimensions and then concatenated with an extra class token of the same dimensionality. 
Following the basic implementation in \texttt{DeiT-S}\cite{touvronTrainingDataefficientImage2021}, 
our ViTs consist of $12$ layers of standard Transformer encoders \cite{vaswaniAttentionAllYou2017} with $h = 6$ attention heads each. 
The MLP heads $H$ consist of three linear layers (with GELU activation 
) with $2048$ hidden dimensions, followed by a bottleneck layer of $256$ dimensions, $l_2$ normalisation and a weight-normalised layer \cite{salimansWeightNormalizationSimple2016} 
to output predictions of $K = 65536$ dimensions, as suggested by \cite{caronEmergingPropertiesSelfSupervised2021}.


\subsubsection{Stage 2: Weakly supervised hierarchical prediction}
\label{sec:method_annoexp}

In Stage 2, 
a small random portion of labelled samples is used to train a hierarchical set of predictors $\{G_\text{cls}, G_\text{exp}\}$, including a predictor $G_\text{exp}^{(i)}$ for each nodule attribute $i$, and a malignancy predictor $G_\text{cls}$ whose input is the concatenation of extracted features $f$ and the predicted human-understandable nodule attributes.


To jointly utilise the extracted features, annotations, and unlabelled data, the set of predictors $G = \{G_\text{cls}, G_\text{exp}\}$ is trained with an annotation exploitation mechanism (Fig.~\ref{fig:overview}), consisting of three key components elaborated as follows. 



\textbf{Sparse seeding.} 
To mitigate potential bias and randomness, we select seed samples by clustering\cite{huGoodStartUsing2010} the extracted features in the learned space via the primary encoder. The extracted features are clustered into $n$ clusters, where $n$ equals the number of seed samples to select. Then the sample closest to each cluster centroid (based on cosine similarity) is selected as $f^{seed}$, whose annotations $y^{seed} = \{y^{seed}_\text{cls}, y^{seed}_\text{exp}\}$ are used to train the predictors from the initial status $\text{st}_0$ to the seeded status $\text{st}_1$: 
\begin{equation}
    G^{\text{st}_1} = \argmin_{\{G_\text{cls}, G_\text{exp}\}} \mathcal{L} \Bigl( y^{seed}, \texttt{softmax}\left( G^{\text{st}_0}(f^{seed}) \right) \Bigr),
\end{equation}
where $\mathcal{L}$ denotes the cross-entropy loss.

\textbf{Semi-supervised active learning.}
Semi-supervised learning and active learning are conducted simultaneously\cite{wangCostEffectiveActiveLearning2017} to exploit the available data. We adopt the classic yet effective uncertainty sampling by least confidence as acquisition strategy \cite{settlesUncertaintySampling2012} to request annotations $y^{request}$ for the uncertain/informative samples $f^{request}$. Concurrently, other samples with relatively high confidence are assigned with pseudo annotations $y^{pseudo(\text{st}_t)}$ by the prediction of $G^{\text{st}_t}$ at status $\text{st}_t$.

\textbf{Quenching.}
To facilitate training under the restrictions of limited samples and complex annotation space, we propose "quenching" as a training technique. 
Similar to Curriculum Pseudo Labelling\cite{cascante-bonillaCurriculumLabelingRevisiting2021, zhangFlexMatchBoostingSemiSupervised2021}, at a certain status $\text{st}_t$ since $\text{st}_1$, the predictor weights are reinitialised to $G^{\text{st}_0}$ to avoid potential confirmation bias\cite{arazoPseudoLabelingConfirmationBias2020}. Meanwhile, the pseudo annotations are updated to the current prediction results: 
\begin{equation}
\begin{split}
    G^{\text{st}_{t+1}} &= \argmin_{\{G_\text{cls}, G_\text{exp}\}} \mathcal{L} \Bigl( \{y^{request}, y^{pseudo(\text{st}_t)}\}, \\
    &\texttt{softmax}\left( G^{\text{st}_0}\left( \{f^{request}, f^{pseudo}\}\right) \right) \Bigl),
\end{split}
\end{equation}
to preserve the learned information and resume training.

\subsection{Inference from multimodal and multiscale explanations}
During inference, the proposed self-explanatory lung nodule diagnosis generates multimodal and multiscale explanations prior to using them to diagnose nodule malignancy. 

Our model-level global explanation is an intrinsic property of proposed method; 
instance-level case-based explanation and local visual explanation are generated in an unsupervised manner; only concept explanation requires as little as 1\% of annotation. 

\textbf{Model-level global explanation} 
includes both model perspective and data perspective. 
From the model perspective, the model's decision procedure is self-explanatory in design and accords with radiologists', which considers both nodule images and their observed nodule attributes. 
Given a nodule image $x^\text{test}$ with malignancy annotation $y_\text{cls}$ and explanation annotation $y_\text{exp}^{(i)}$ for each nodule attribute $i=1,\cdots, M$, its feature $f$ is extracted via the primary encoder: $f = E_{\theta^\text{E}_\text{pri}} (x^\text{test})$. 
The predicted human-understandable nodule attribute $i$ is generated by a simple predictor $G_\text{exp}^{(i)}$:
$z_\text{exp}^{(i)} = G_\text{exp}^{(i)} (f)$.
Then the malignancy $z_\text{cls}$ is predicted by a predictor $G_\text{cls}$ from the concatenation ($\oplus$) of extracted features $f$ and predicted nodule attributes:
\begin{equation}
    z_\text{cls} = G_\text{cls} (f \oplus z_\text{exp}^{(1)} \oplus \cdots \oplus z_\text{exp}^{(M)}).
    \label{eq:infer}
\end{equation}
From the data perspective, the explanation is provided qualitatively by the separability of malignancy and its correlation to the clustering of each nodule attribute in the learned latent space. 

\textbf{Instance-level case-based explanation}
is given by the k-Nearest Neighbour (k-NN) instances in the learned latent space. For a testing nodule image $x^\text{test}$ with its extracted feature $f^\text{test}$, a set $S^\ast$ of $k$ similar samples from the training data are retrieved according to the cosine similarities between their extracted features:
\begin{equation}
    S^\ast = \argmin_S  \sum_{f^\text{train} \in S} \frac{f^\text{test} \cdot f^\text{train}}{\|f^\text{test}\| \cdot \|f^\text{train}\|}.
    \label{eq:knn}
\end{equation}

\textbf{Local visual explanation} takes directly from the self-attention mechanism with $h$ attention heads: 
\begin{equation}
    V_\text{exp} = \frac{1}{h} \sum_{i=1}^h \texttt{softmax}\left(\frac{Q_i \cdot K_i^T}{\sqrt{d / h}}\right),
\end{equation}
where $Q_i$ and $K_i$ are the corresponding splits from $Q$ and $K$.

\textbf{Concept explanation} is given as a prediction $z_\text{exp}^{(i)}$ of human-understandable nodule attributes $i$ by the aforementioned set of predictors $G_\text{exp}^{(i)}$: $z_\text{exp}^{(i)} = G_\text{exp}^{(i)} (f)$.

\section{Experiments and discussion}
\label{sec:experiments}

\subsection{Implementation details}
Since our complete code for data pre-processing, implementation and all experiments is open-source available, here we briefly state our main settings and refer to our code repository at \url{https://github.com/diku-dk/credanno} for further details.

\subsubsection{Data pre-processing}
Although several previous works\cite{shenInterpretableDeepHierarchical2019, lalondeEncodingVisualAttributes2020, chenEndtoEndMultiTaskLearning2021, liuMultiTaskDeepModel2020, joshiLungNoduleMalignancy2021} have been evaluated on the publicly available LIDC dataset\cite{armatoLungImageDatabase2011}, their discrepancies in sample selection and pre-processing increase the difficulty in making direct comparisons. Therefore, we follow the common procedure summarised in \cite{baltatzisPitfallsSampleSelection2021}. 
Scans with slice thickness larger than $2.5\,mm$ are discarded for being unsuitable for lung cancer screening according to clinical guidelines \cite{kazerooniACRSTRPractice2014}, and the remaining scans are resampled to the resolution of $1\,mm^3$ isotropic voxels. Only nodules annotated by at least three radiologists are retained. Annotations for both malignancy and nodule attributes of each nodule are aggregated by the median value among radiologists. Malignancy score is binarised by a threshold of $3$: nodules with median malignancy score larger than $3$ are considered malignant, smaller than $3$ are considered benign, while the rest are excluded\cite{baltatzisPitfallsSampleSelection2021}. 

For each annotation, 
only a 2D patch of size $32 \times 32\,px$ is extracted from the central axial slice. Although an image is extracted for each annotation, our training($70\%$)/testing($30\%$) split is on nodule level to ensure no image of the same nodule exists in both training and testing sets. This results in $276/242$ benign/malignant nodules for training and $108/104$ benign/malignant nodules for testing.

\subsubsection{Training settings}
The network is trained in two separate stages.

In Stage 1, 
the training of the feature extraction follows the suggestions in \cite{caronEmergingPropertiesSelfSupervised2021}. The encoders and prediction heads are trained for $300$ epochs with an AdamW optimiser \cite{loshchilovFixingWeightDecay2018} 
and batch size $128$, starting from the weights pretrained unsupervisedly on ImageNet\cite{russakovskyImageNetLargeScale2015}. The learning rate is linearly scaled up to $0.00025$ during the first 10 epochs and then follows a cosine scheduler \cite{loshchilovSGDRStochasticGradient2016} 
to decay till $10^{-6}$. The temperatures for the two branches are set to $\tau_\text{pri} = 0.04$, $\tau_\text{aux} = 0.1$. The momentum factor $\lambda$ is set to $0.9$, while $m$ is increased from $0.996$ to $1$ following a cosine scheduler. 
The data augmentation for encoder training adapts from BYOL\cite{grillBootstrapYourOwn2020} and includes multi-crop \cite{caronUnsupervisedLearningVisual2020}.

In Stage 2, K-means are used for clustering to select $1\%$ annotation as seed samples. The predictors $G_\text{exp}^{(i)}$ and $G_\text{cls}$, each consisting of one linear layer, are first jointly trained using the seed samples for $100$ epochs with SGD optimisers with momentum $0.9$ and batch size $128$. The learning rate follows a cosine scheduler with initial value $0.00025$. 
After reaching the seeded status $G^{\text{st}_1}$, the predictors and optimisers are quenched for the first time. The training then resumes using the requested and dynamic pseudo annotations for $50$ more epochs, where quenching happens every $10$ epochs.
The input images are augmented following previous works\cite{al-shabiLungNoduleClassification2019, baltatzisPitfallsSampleSelection2021} on the LIDC dataset, including random scaling within $[0.08, 1]$ and resizing to $224 \times 224\,px$, random horizontal/vertical flip ($p=0.5$), random rotation by a random integer multiple of $\ang{90}$, and Gaussian blur with kernel size $1$.

\begin{figure*}[tbp]
    \centering
    \subfloat[Malignancy]{
        \includegraphics[width=0.229\textwidth]{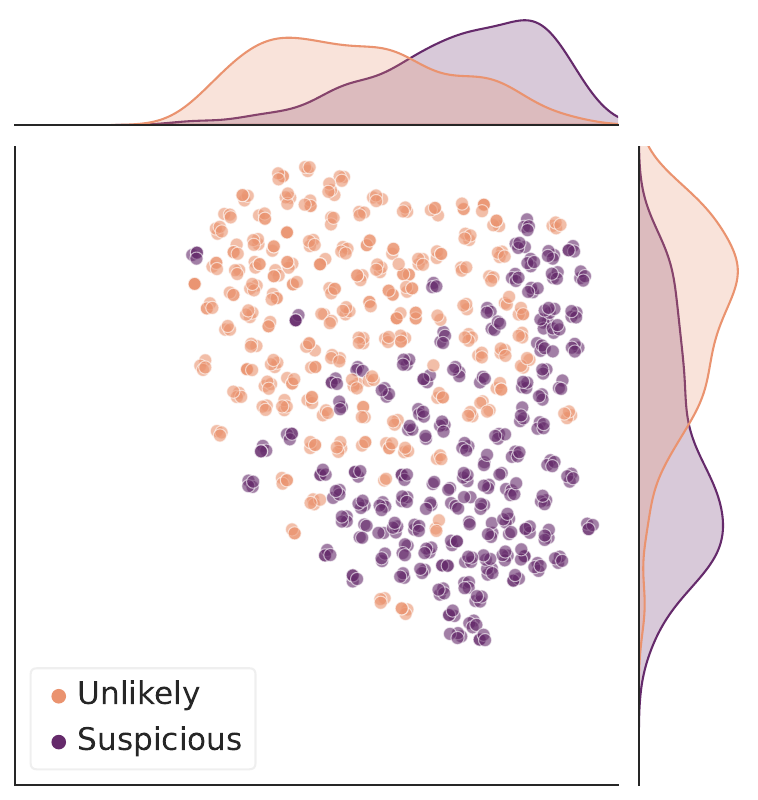}
        \label{fig:tsne_mal}
        }\hfil
    \subfloat[Subtlety]{
        \includegraphics[width=0.229\textwidth]{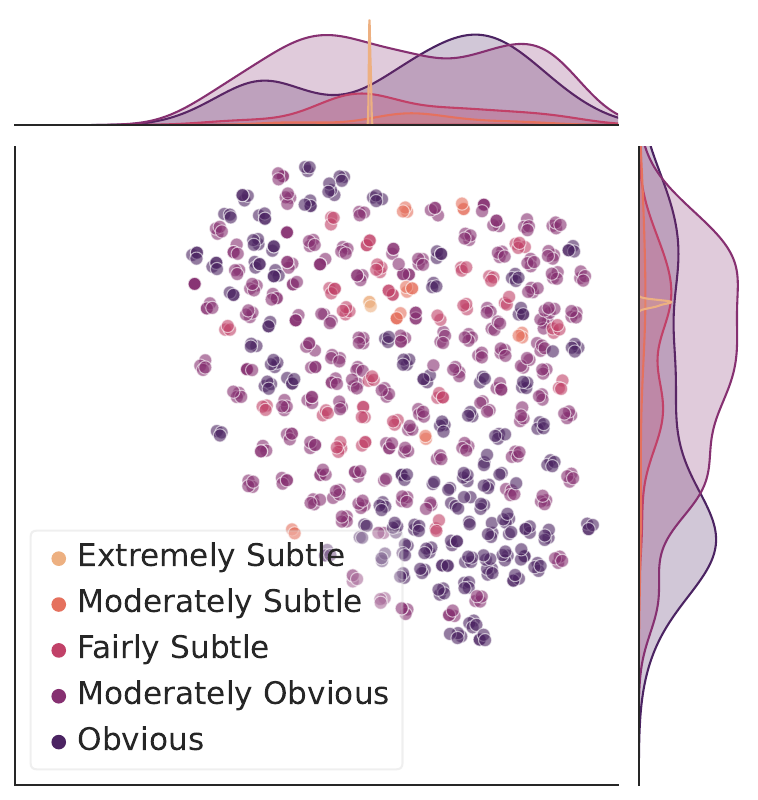}
        \label{fig:tsne_sub}
        }\hfil
    \subfloat[Calcification]{
        \includegraphics[width=0.229\textwidth]{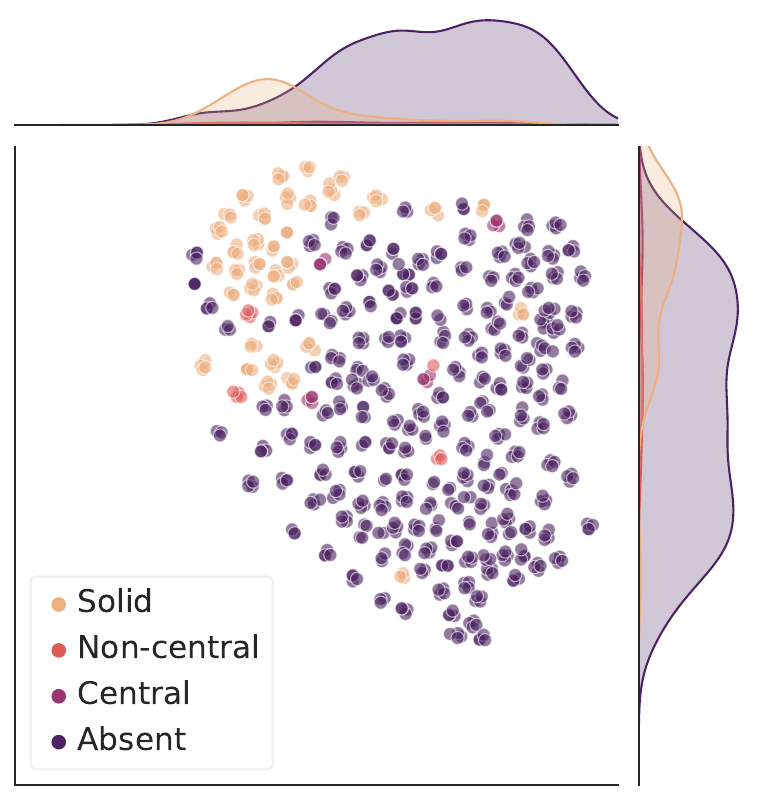}
        \label{fig:tsne_cal}
        }\hfil
    \subfloat[Sphericity]{
        \includegraphics[width=0.229\textwidth]{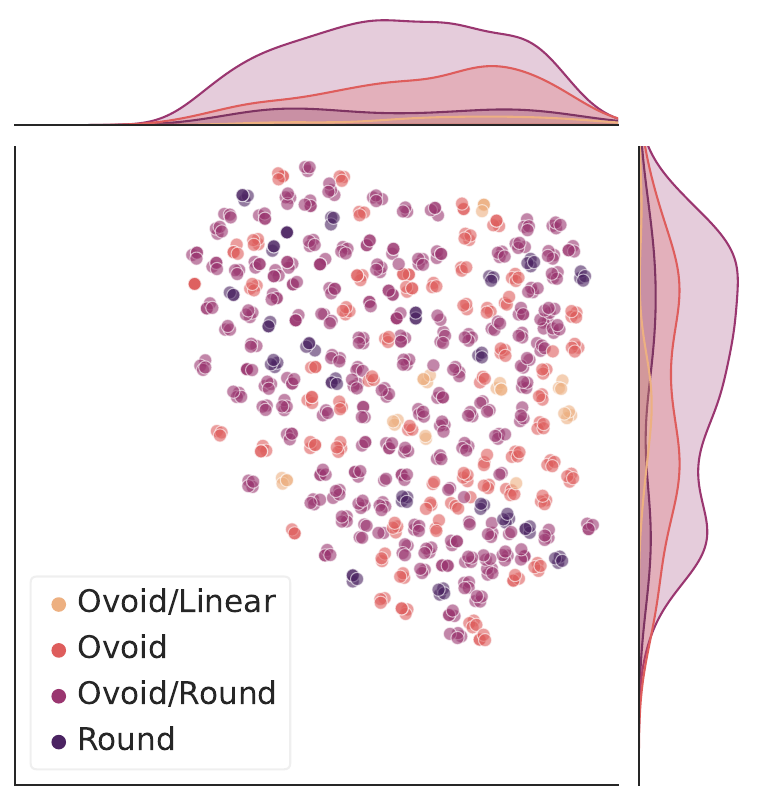}
        \label{fig:tsne_sph}
        }\\[-1ex]

    \subfloat[Margin]{
        \includegraphics[width=0.229\textwidth]{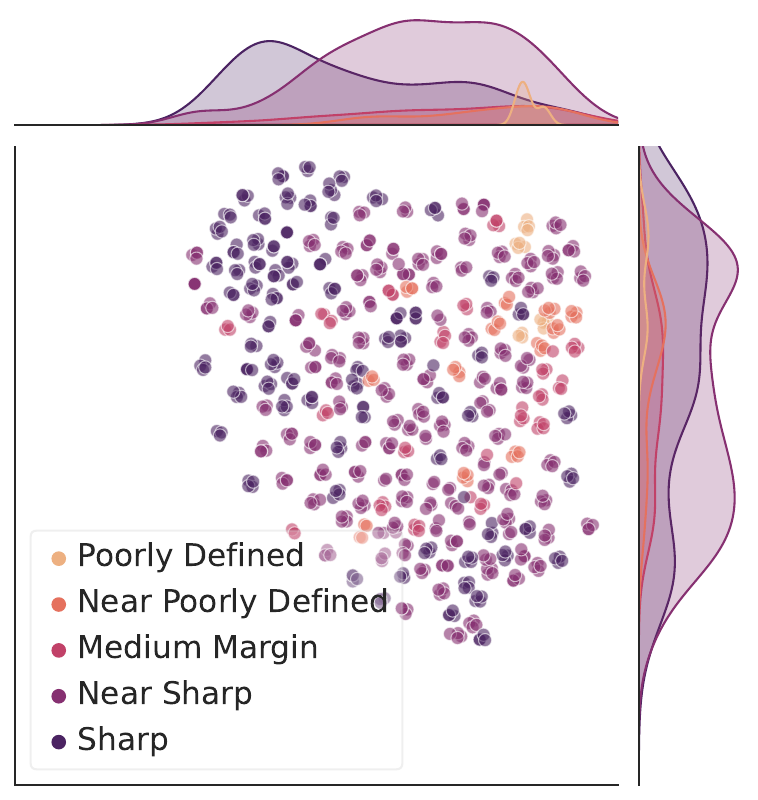}
        \label{fig:tsne_mar}
        }\hfil
    \subfloat[Lobulation]{
        \includegraphics[width=0.229\textwidth]{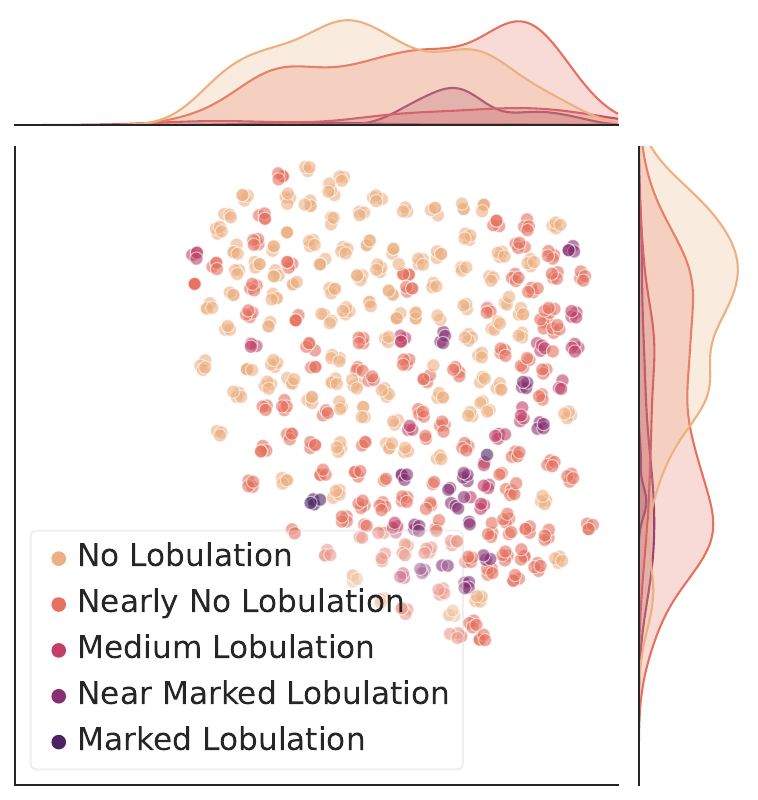}
        \label{fig:tsne_lob}
        }\hfil
    \subfloat[Spiculation]{
        \includegraphics[width=0.229\textwidth]{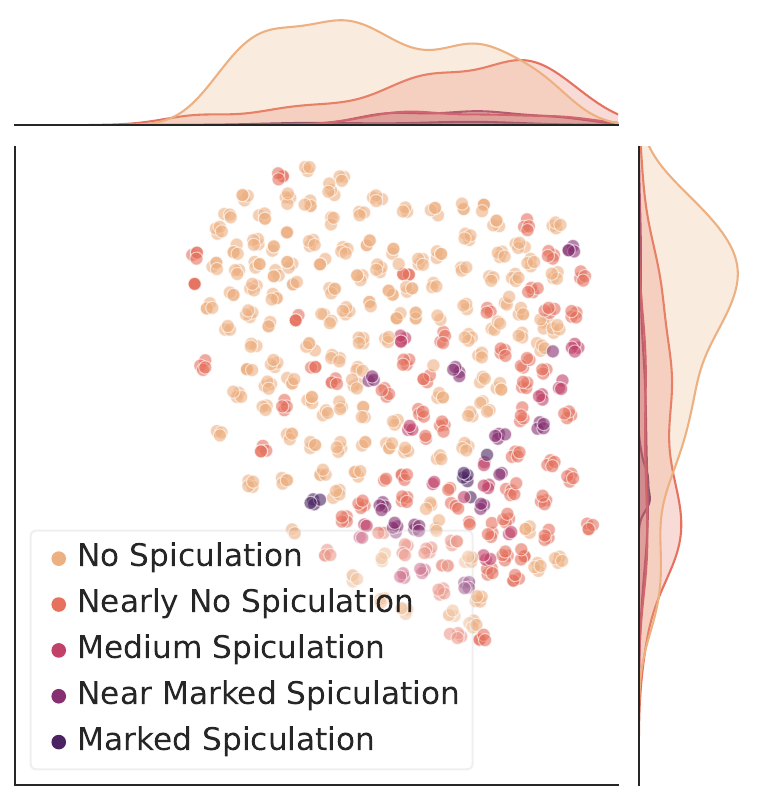}
        \label{fig:tsne_spi}
        }\hfil
    \subfloat[Texture]{
        \includegraphics[width=0.229\textwidth]{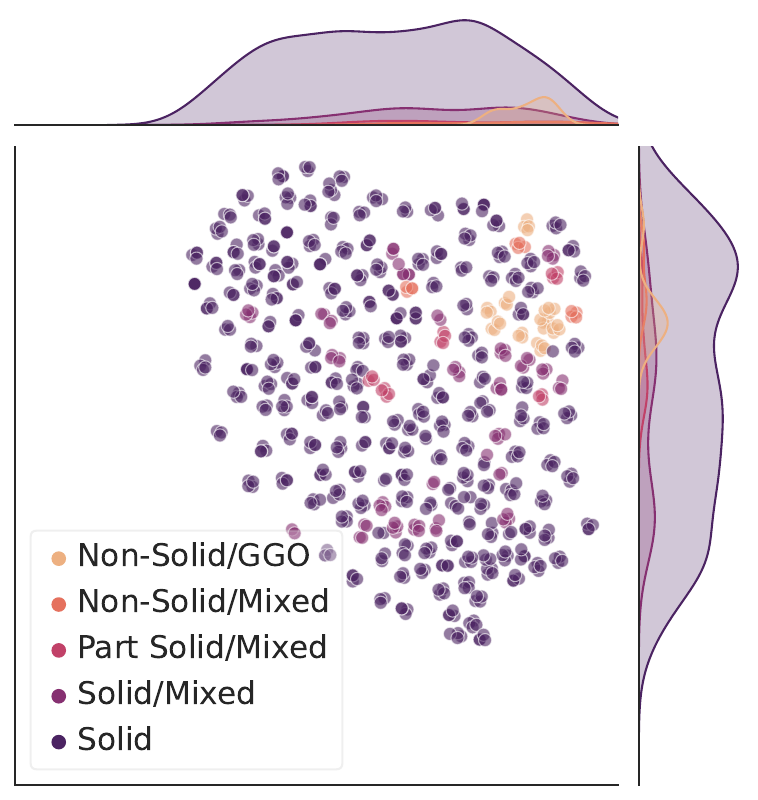}
        \label{fig:tsne_tex}
        }\\[-0.7ex]

    \caption{\textbf{t-SNE visualisation of features extracted from testing images.}
    Data points are coloured using ground truth annotations.
    Malignancy shows highly separable in the learned space, and semantically correlates with the clustering in each nodule attribute.
    }
    \label{fig:res_tsne}
\end{figure*}

\subsection{Model-level global explanation}

The reduction of annotations relies heavily on the separability and semantic information of the learned feature space. We use t-SNE 
to visualise the learned feature as a qualitative evaluation. Feature $f$ extracted from each testing image is mapped to a data point in 2D space. Fig.~\ref{fig:tsne_mal} to \ref{fig:tsne_tex} correspond to these data points coloured by the ground truth annotations of malignancy to nodule attribute ``texture", respectively. 

Fig.~\ref{fig:res_tsne} intuitively demonstrates the underlying correlation between malignancy and nodule attributes.
For instance, the cluster in Fig.~\ref{fig:tsne_cal} indicates that solid calcification negatively correlates with nodule malignancy. 
Similarly, Fig.~\ref{fig:tsne_lob} and Fig.~\ref{fig:tsne_spi} indicate that lobulation is associated with spiculation, both of which are positively correlated with malignancy.
These semantic correlations coincide with the radiologists' diagnostic process\cite{vlahosLungCancerScreening2018} and thereby further support the potential of the proposed approach as a trustworthy decision support system.

More importantly, Fig.~\ref{fig:tsne_mal} shows that even in this 2D space for visualisation, the samples show reasonably separable in both malignancy and nodule attributes. This provides the possibility to train the initial predictors using only a very small number of seed annotations, provided they are sufficiently dispersed and informative. 

\subsection{Instance-level case-based explanation}
\label{sec:res_knn}

In this subsection, we present a comprehensive evaluation of MERA's instance-level case-based explanation performance, analysing both qualitative and quantitative metrics to assess its effectiveness and robustness in clinical decision-making.

\begin{figure*}[tbp]
    \centering
    \subfloat[CT image and visual explanation]{
        \includegraphics[height=0.35\textwidth]{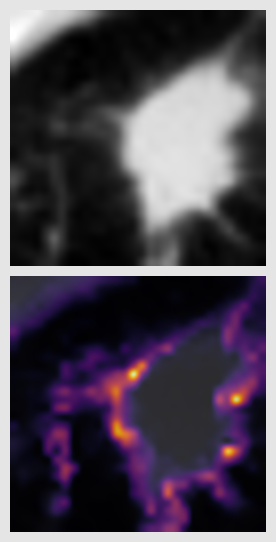}
        \label{fig:retri_malig_q}}\hfil
    \subfloat[Nearest training samples with groundtruth annotation (if available)]{
        \includegraphics[height=0.35\textwidth]{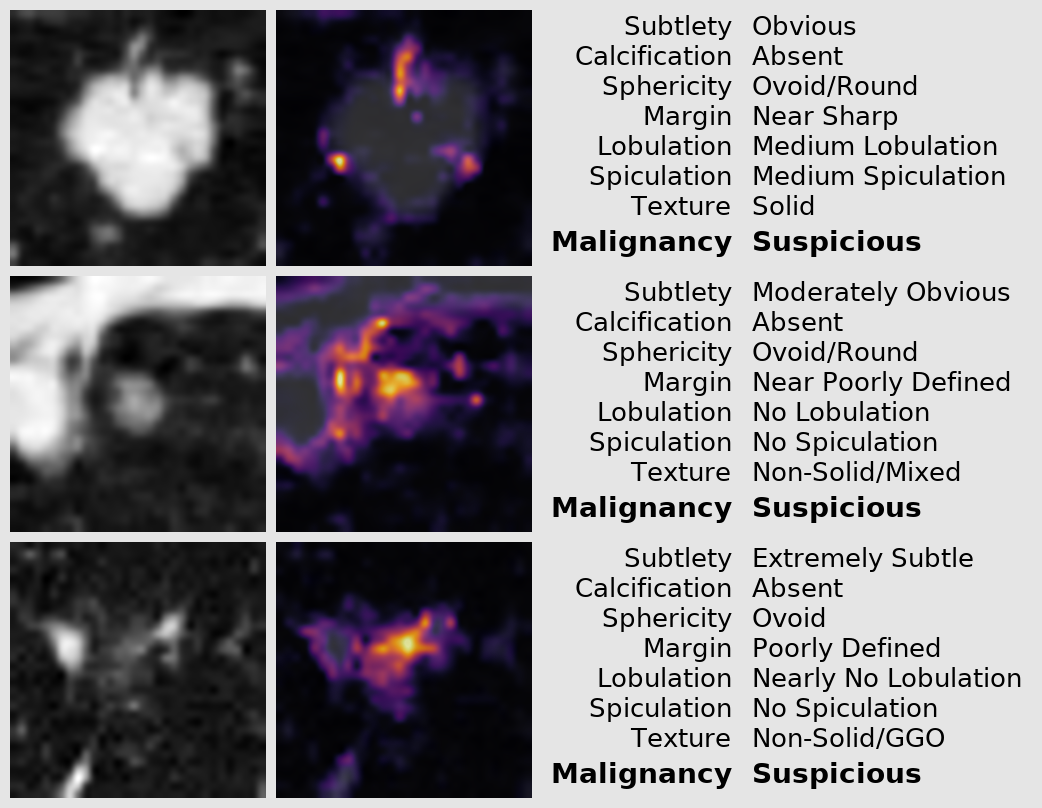}
        \label{fig:retri_malig_train}}
    \subfloat[Nearest testing samples]{
        \includegraphics[height=0.35\textwidth]{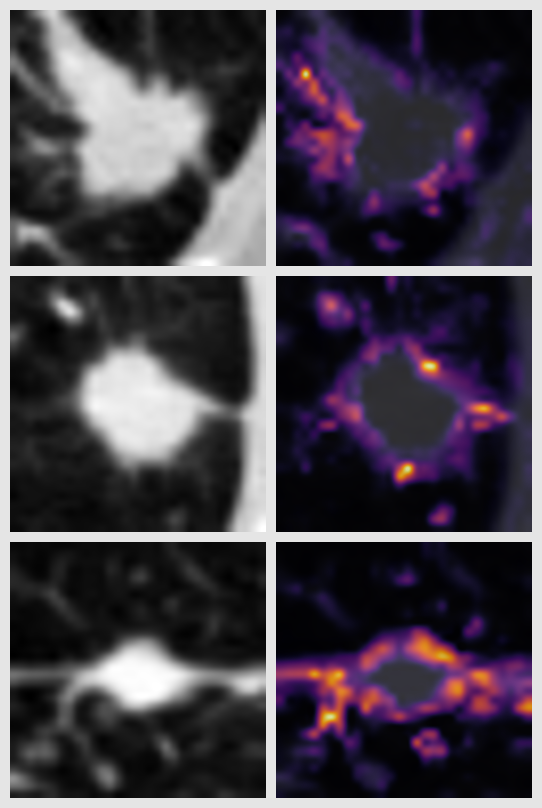}
        \label{fig:retri_malig_test}}\\[-0.7ex]
    \caption{\textbf{Case-based explanation of a malignant nodule}.
    Nearest training and testing samples illustrating similarities in poorly defined margins, lobulation, and spiculation, which are typical morphological characteristics suggesting malignancy.
    }
    \label{fig:retri_malig}
\end{figure*}

\begin{figure*}[tbp]
    \centering
    \subfloat[CT image and visual explanation]{
        \includegraphics[height=0.35\textwidth]{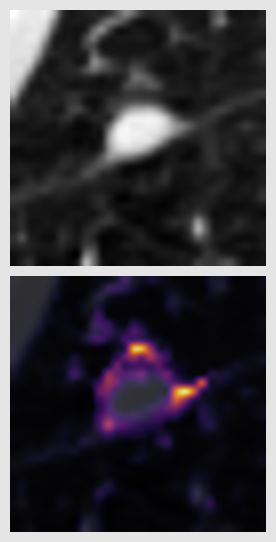}
        \label{fig:retri_benign_q}}\hfil
    \subfloat[Nearest training samples with groundtruth annotation (if available)]{
        \includegraphics[height=0.35\textwidth]{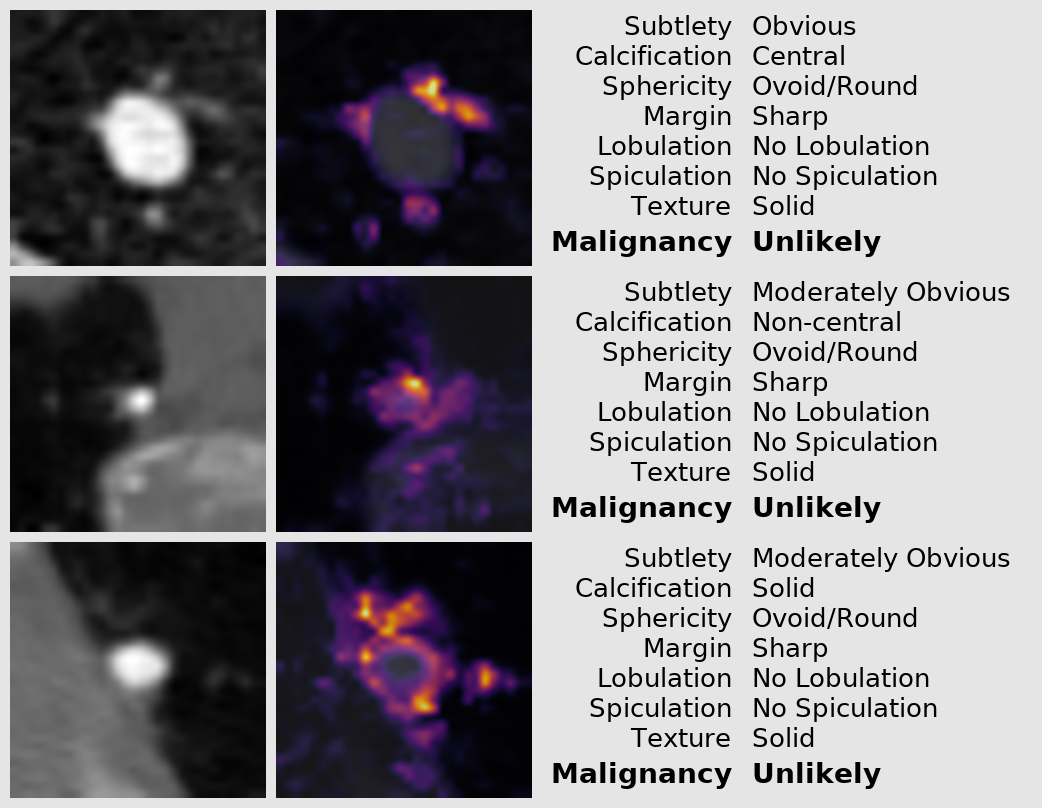}
        \label{fig:retri_benign_train}}
    \subfloat[Nearest testing samples]{
        \includegraphics[height=0.35\textwidth]{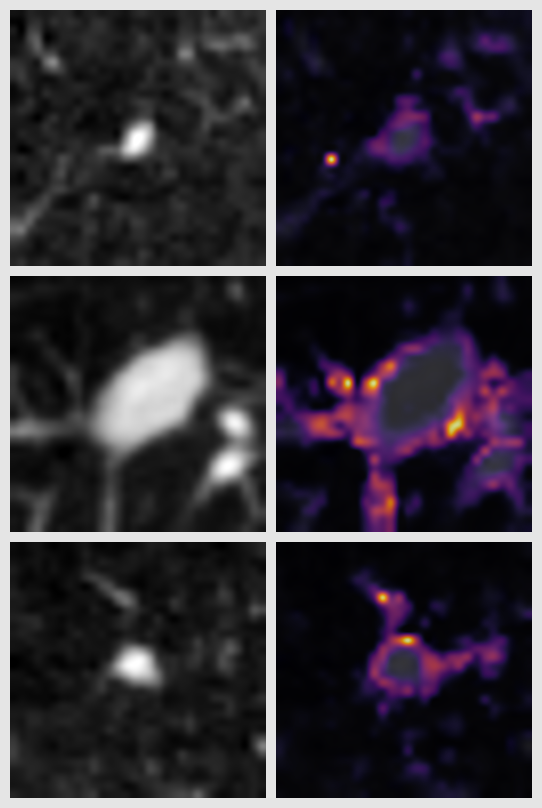}
        \label{fig:retri_benign_test}}\\[-0.7ex]
    \caption{\textbf{Case-based explanation of a benign nodule.}.
    Nearest training and testing samples display the same solid texture, smooth and sharp margins, and lack of lobulations or spiculations, mirroring the features of the target nodule.
    }
    \label{fig:retri_benign}
\end{figure*}


Fig.~\ref{fig:retri_malig} presents a detailed case-based explanation for a suspicious lung nodule, emphasising the utility of our model in identifying diagnostically relevant features. The target nodule in Fig.~\ref{fig:retri_malig_q} exhibits a partly solid structure with increased density and no calcification, marked by irregular margins and lobulation. These features are crucial indicators often associated with malignancy.
The training samples in Fig.~\ref{fig:retri_malig_train} display varying degrees of morphological similarity to the target case in Fig.~\ref{fig:retri_malig_q}, particularly in terms of poorly defined margins, lobulation, and spiculation, all of which enhance the malignancy suspicion. Notably, the second training sample also shares a mixed texture characteristic, albeit less pronounced, which aligns with the texture seen in the target case. 
Moreover, the testing samples in Fig.~\ref{fig:retri_malig_test} manifest a consistent pattern of lobulation and spiculation, alongside potentially similar relationships with pleura and poorly defined margins, mirroring the target case's characteristics. 
This resemblance in visual features across the training and testing examples reinforces the model's consistency in detecting and highlighting critical features across similar cases, supporting the model's decision process.

Similarly, Fig.~\ref{fig:retri_benign} presents a detailed case-based explanation of a benign lung nodule. The target nodule in Fig.~\ref{fig:retri_benign_q} exhibits classic characteristics of a benign lesion, with a solid texture, smooth and sharp margins, and absence of lobulation or spiculation. 
The nearest training samples in Fig.~\ref{fig:retri_benign_train} and testing samples in Fig.~\ref{fig:retri_benign_test} exhibit similar benign characteristics, such as solid texture, sphericity, sharply defined margins, which align closely with the features observed in the target nodule. The absence of malignant indicators like lobulation or spiculation across these examples further reinforces the benign nature of the lesion. 
This consistency demonstrates the model's ability to recognise and focus on features that are typical of intrapulmonary lymph nodes, aiding in accurate and reliable benign diagnoses.

By aligning its focus with radiological best practices, the model consistently matched the target cases with visually and diagnostically similar training and testing samples, underscoring its reliability and generalisation in instance-level explanations. 

\begin{table*}[t]
    \centering
    \caption{\textbf{Prediction accuracy ($\%$) of nodule attributes and malignancy.}
    The best in each column is \textbf{bolded} for full/partial annotation respectively.
    Dashes (-) denote values not reported by the compared methods. 
    Results of our proposed \colorbox[HTML]{E2EFD9}{MERA} are highlighted.
    Observe that by simply assigning labels of nearest training samples, MERA reaches over $90\%$ accuracy simultaneously in predicting all nodule attributes and competitive accuracy in malignancy prediction, meanwhile using the fewest nodules and no additional information.
    }
    \label{tab:res_knn}
    \resizebox{\textwidth}{!}{%
    \begin{threeparttable}
        \begin{tabular}{lcccccccccc}
            \hline
             & \multicolumn{7}{c}{\textbf{Nodule attributes}} & \multicolumn{1}{l}{} & \multicolumn{1}{l}{} & \multicolumn{1}{l}{} \\ \cline{2-8}
             & \textbf{Sub} & \textbf{Cal} & \textbf{Sph} & \textbf{Mar} & \textbf{Lob} & \textbf{Spi} & \textbf{Tex} & \multicolumn{1}{l}{\multirow{-2}{*}{\textbf{Malignancy}}} & \multicolumn{1}{l}{\multirow{-2}{*}{\textbf{\#nodules}}} & \multicolumn{1}{l}{\multirow{-2}{*}{\begin{tabular}[c]{@{}c@{}}\textbf{No additional} \\ \textbf{information}\end{tabular}}} \\ \hline

            \multicolumn{11}{l}{Full annotation} \\
            \textbf{HSCNN}\cite{shenInterpretableDeepHierarchical2019} & 71.90 & 90.80 & 55.20 & 72.50 & - & - & 83.40 & 84.20 & 4252 & \xmark\tnote{c} \\
            \textbf{X-Caps}\cite{lalondeEncodingVisualAttributes2020} & 90.39 & - & 85.44 & 84.14 & 70.69 & 75.23 & 93.10 & 86.39 & 1149 & \cmark \\
            \textbf{MSN-JCN}\cite{chenEndtoEndMultiTaskLearning2021} & 70.77 & \textbf{94.07} & 68.63 & 78.88 & \textbf{94.75} & \textbf{93.75} & 89.00 & 87.07 & 2616 & \xmark\tnote{d} \\
            \textbf{MTMR}\cite{liuMultiTaskDeepModel2020} & - & - & - & - & - & - & - & \textbf{93.50} & 1422 & \xmark\tnote{e} \\
            \rowcolor[HTML]{E2EFD9} 
            \textbf{MERA (20-NN)} & 94.28 & 92.07 & 95.45 & \textbf{94.54} & 91.81 & 92.33 & 94.54 & 86.35 & \cellcolor[HTML]{E2EFD9} & \cellcolor[HTML]{E2EFD9} \\
            \rowcolor[HTML]{E2EFD9} 
            \textbf{MERA (50-NN)} & 94.93 & 92.72 & 95.58 & 93.76 & 91.29 & 92.72 & \textbf{94.67} & 87.52 & \cellcolor[HTML]{E2EFD9} & \cellcolor[HTML]{E2EFD9} \\
            \rowcolor[HTML]{E2EFD9} 
            \textbf{MERA (150-NN)} & 95.32 & 92.59 & 96.10 & 94.28 & 90.90 & 92.33 & 93.63 & 88.30 & \cellcolor[HTML]{E2EFD9} & \cellcolor[HTML]{E2EFD9} \\
            \rowcolor[HTML]{E2EFD9} 
            \textbf{MERA (250-NN)} & \textbf{96.36} & 92.59 & \textbf{96.23} & 94.15 & 90.90 & 92.33 & 92.72 & 88.95 & \multirow{-4}{*}{\cellcolor[HTML]{E2EFD9}\textbf{730}} & \multirow{-4}{*}{\cellcolor[HTML]{E2EFD9}\cmark} \\ \hline

            \multicolumn{11}{l}{Partial annotation} \\
            \textbf{WeakSup}\cite{joshiLungNoduleMalignancy2021} \textbf{(1:5\tnote{a} )} & 43.10 & 63.90 & 42.40 & 58.50 & 40.60 & 38.70 & 51.20 & 82.40 &  &  \\
            \textbf{WeakSup}\cite{joshiLungNoduleMalignancy2021} \textbf{(1:3\tnote{a} )} & 66.80 & 91.50 & 66.40 & 79.60 & 74.30 & 81.40 & 82.20 & \textbf{89.10} & \multirow{-2}{*}{2558} & \multirow{-2}{*}{\xmark\tnote{f}} \\
            \rowcolor[HTML]{E2EFD9} 
            \textbf{MERA (10\%\tnote{b}, 20-NN)} & 94.54 & 90.90 & 96.23 & 93.76 & 91.03 & \textbf{91.42} & \textbf{94.41} & 87.13 & \cellcolor[HTML]{E2EFD9} & \cellcolor[HTML]{E2EFD9} \\
            \rowcolor[HTML]{E2EFD9} 
            \textbf{MERA (10\%\tnote{b}, 50-NN)} & 94.93 & \textbf{92.07} & 96.75 & \textbf{94.28} & \textbf{92.59} & 91.16 & 94.15 & 87.13 & \cellcolor[HTML]{E2EFD9} & \cellcolor[HTML]{E2EFD9} \\
            \rowcolor[HTML]{E2EFD9} 
            \textbf{MERA (10\%\tnote{b}, 150-NN)} & \textbf{95.32} & 89.47 & \textbf{97.01} & 93.89 & 91.81 & 90.51 & 92.85 & 88.17 & \multirow{-3}{*}{\cellcolor[HTML]{E2EFD9}\textbf{730}} & \multirow{-3}{*}{\cellcolor[HTML]{E2EFD9}\cmark} \\ \hline
        \end{tabular}%
        \begin{tablenotes}
            \item[a] $1:N$ indicates that $\frac{1}{1+N}$ of training samples have annotations on nodule attributes. (All samples have malignancy annotations.)
            \item[b] The proportion of training samples that have annotations on nodule attributes and malignancy.
            \item[c] 3D volume data are used.
            \item[d] Segmentation masks and nodule diameter information are used. Two other traditional methods are used to assist training.
            \item[e] All 2D slices in 3D volumes are used.
            \item[f] Multi-scale 3D volume data are used.
        \end{tablenotes}
    \end{threeparttable}
    }
    \vspace{-0.7ex}
\end{table*}

In addition, a quantitative analysis is also conducted to provide a comprehensive understanding of the model's effectiveness and robustness.

The instance-level case-based explanation performance is quantified in terms of prediction accuracy of nodule attributes and malignancy by directly assigning the majority labels of the set $S^\ast$ of $k$ nearest training samples, as formulated in Eq.~\ref{eq:knn}. 

The evaluation procedure follows previous works\cite{luReducingAnnotationNeed2022, luCRedAnnoAnnotationExploitation2023}:
each annotation is considered independently \cite{shenInterpretableDeepHierarchical2019}; the predictions of nodule attributes are considered correct if within $\pm1$ of aggregated radiologists' annotation \cite{lalondeEncodingVisualAttributes2020}; attribute ``internal structure" is excluded from the results because its heavily imbalanced classes are not very informative \cite{shenInterpretableDeepHierarchical2019, lalondeEncodingVisualAttributes2020, chenEndtoEndMultiTaskLearning2021, liuMultiTaskDeepModel2020, joshiLungNoduleMalignancy2021}. 

Tab.~\ref{tab:res_knn} summarises the overall prediction performance and compares it with the state-of-the-art deep-learning methods. 
The results show that by simply assigning labels of nearest training samples, MERA reaches over $90\%$ accuracy simultaneously in predicting all nodule attributes, which outperforms all previous works whilst using only $730$ nodules and no additional information. 
Meanwhile, in predicting nodule malignancy, MERA achieves competitive accuracy and even outperforms other methods that rely on a larger dataset and more extensive supervision.
Notably, when reducing the annotated training samples to just $10\%$, MERA's performance remains largely unaffected. 
In addition, the consistent decent prediction performance also indicates that our approach is reasonably robust w.r.t. to the value $k$ in k-NN classifiers.

The demonstrated instance-level case-based explanation performance, both qualitatively and quantitatively, indicates that by providing diagnostic context that enhances the interpretability and reliability of the model’s predictions, MERA offers a transparent and robust support in clinical decision-making.

\subsection{Local visual explanation: case study}

This section delves into the nuanced capabilities of the proposed method in discerning and visualising the intricate features of lung nodules through the local visual explanations. We critically assess the model's performance across a diverse spectrum of cases, categorising them into correctly predicted benign, malignant, and incorrectly predicted samples, fostering a deeper understanding of its diagnostic and explanatory power in a clinical context. 
Each subsection showcases how the model pinpoints and interprets various critical diagnostic features of lung nodules, highlighting its effectiveness in distinguishing between benign and malignant characteristics, as well as identifying areas requiring improvement. Through this comprehensive case analysis, we aim to demonstrate the model's capabilities and limitations, providing insights into its potential applications in clinical practice and its utility in aiding radiologists in making informed decisions.

More non-cherry-picked samples are included in \ref{sec:appendix}.

\subsubsection{Benign nodules}

\begin{figure*}[tbp]
    \centering
    \subfloat[Our method highlights the morphological features and fissure association, distinguishing it from other methods that fail to capture these diagnostic details.]{
        \includegraphics[width=0.9\textwidth]{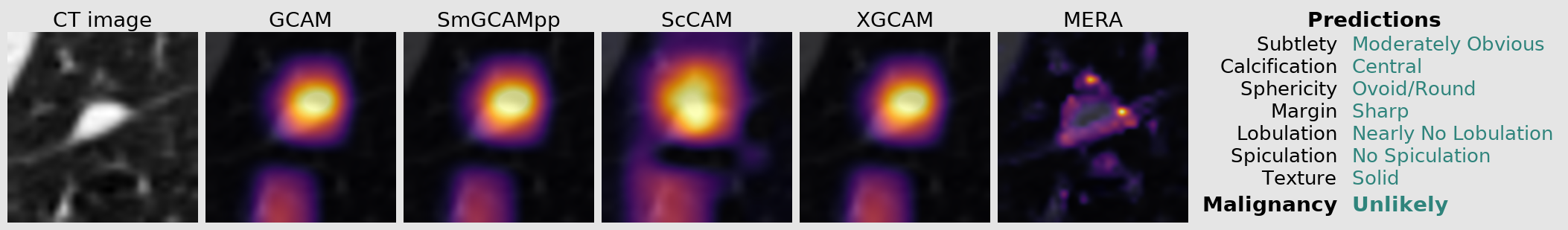}
        \label{fig:benign_1}
        }
        

    \subfloat[The proposed method uniquely emphasise the granuloma’s calcified nature and pleural connection.]{
        \includegraphics[width=0.9\textwidth]{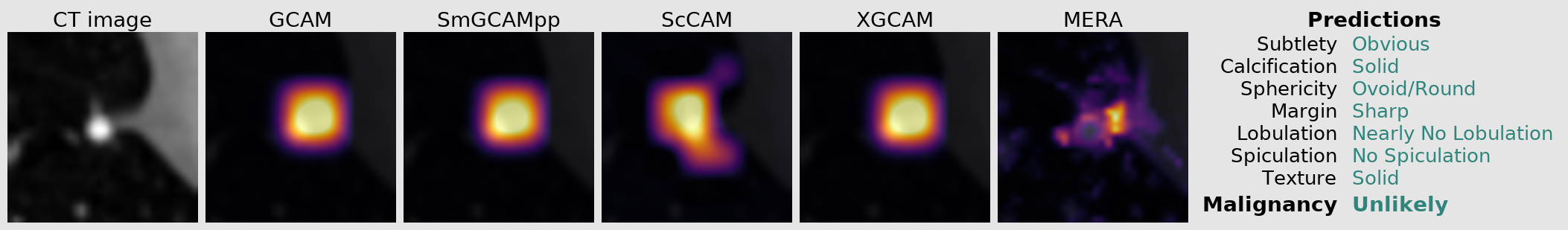}
        \label{fig:benign_2}
        }

    \subfloat[Our method focuses on the shape and pleural connection of the calcified granuloma provides more accurate and contextually relevant information compared to competitors.]{
        \includegraphics[width=0.9\textwidth]{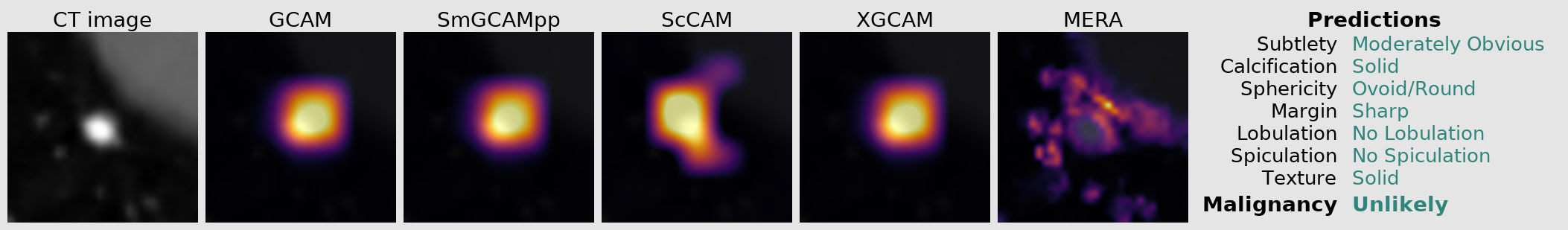}
        \label{fig:benign_3}
        }\\[-0.7ex]

    \caption{\textbf{Local visual explanation of benign nodules.} From left to right: original lung nodule image patch on an axial chest CT, visual explanation of 4 competitor methods, visual explanation of our proposed method, our predicted nodule attributes and malignancy (green font indicates correct and red font indicates wrong, and only $1\%$ annotated data is used for the predictions).
    Each subfigure demonstrates the model's superior ability to focus on key morphological and textural features that confirm the benign nature of the nodules, such as calcifications, smooth margins, and connections with pleura. The proposed method effectively distinguishes these benign characteristics, which are often overlooked by other methods.
    }
    \label{fig:res_benign}
\end{figure*}

Firstly, we explore cases where the proposed method correctly identified benign lung nodules. Fig.~\ref{fig:res_benign} features three distinct samples where the model's diagnostic accuracy is highlighted through its ability to emphasise specific morphological features indicative of benignity. This section illustrates how the model differentiates between benign and malignant nodules, providing crucial visual explanations that underline its effectiveness in recognising features such as calcification, smooth margins, and pleural connections, which are pivotal for accurate benign diagnoses.

The presented case in Fig.~\ref{fig:benign_1} involves a lung nodule with characteristics typical of an intrapulmonary lymph node, as seen in the axial chest CT image (lung window). The proposed model successfully identifies and emphasises the key features of the nodule: its association with the fissure, sphericity, and sharp margins. This can be observed in the attention map where specific areas are highlighted to indicate the relationship with the fissure (left linear highlighted area), the sphericity of the lesion (anterior highlighted area), and the sharply defined contour of the nodule (highlighted contour). These characteristics are crucial in distinguishing benign from malignant nodules.

In contrast, the attention maps from the competitor methods, although capturing the lesion, do not emphasise these defining features distinctly. They adequately detect the presence of the lesion but fail to highlight the morphological details that support the benign nature of the nodule, such as the sharp margins and the association with the fissure. 

This comparative analysis showcases the superiority of the proposed model in focusing on diagnostically relevant features, which supports its utility in clinical settings for enhancing diagnostic accuracy and reducing unnecessary interventions. The accurate prediction of ``Unlikely Malignancy'' aligns with the radiological assessments, confirming the effectiveness of the model in clinical inference.

The two cases presented in Fig.~\ref{fig:benign_2} and Fig.~\ref{fig:benign_3} feature calcified nodules associated with the pleura, suggestive of benign calcified granulomas. The original CT images, when analysed with the proposed model, demonstrate a clear focus on both the shape of the nodules and their anatomical relationships with the pleura, which are crucial aspects used by radiologists for diagnosis.

The attention maps from the proposed model distinctly highlight these areas, contrasting sharply with the competitor models that primarily focus on the existence of the lesion. This specific attention to both morphological features and contextual anatomical relationships enhances the interpretative power of the proposed model, aligning it closely with expert radiological assessment practices. The successful identification of these nodules as benign by focusing on relevant diagnostic cues not only corroborates the radiologist's interpretation but also underscores the model's potential in reducing false positives in pulmonary nodule diagnosis.

In summary, the analysis of correctly predicted benign nodules underscores the proposed method's effectiveness in identifying and highlighting essential features that indicate a nodule’s benign nature. The discussed cases in Fig.~\ref{fig:res_benign} reveal the model's ability to focus on critical diagnostic features such as fissure associations, sphericity, and calcification, distinguishing these nodules from malignant ones. 
Each example illustrated the model’s precision in detecting characteristics that are pivotal for ruling out malignancy, aligning with radiological best practices \cite{macmahonGuidelinesManagementIncidental2017}.
Through detailed visual explanations consistently distinguishing these features more effectively than competitor methods, the proposed method demonstrates its robustness in avoiding false positives, thereby enhancing its utility in clinical settings.

\subsubsection{Malignant nodules}

\begin{figure*}[tbp]
    \centering
    \subfloat[MERA uniquely highlights the cystic components, in addition to spiculations and irregular margins.]{
        \includegraphics[width=0.9\textwidth]{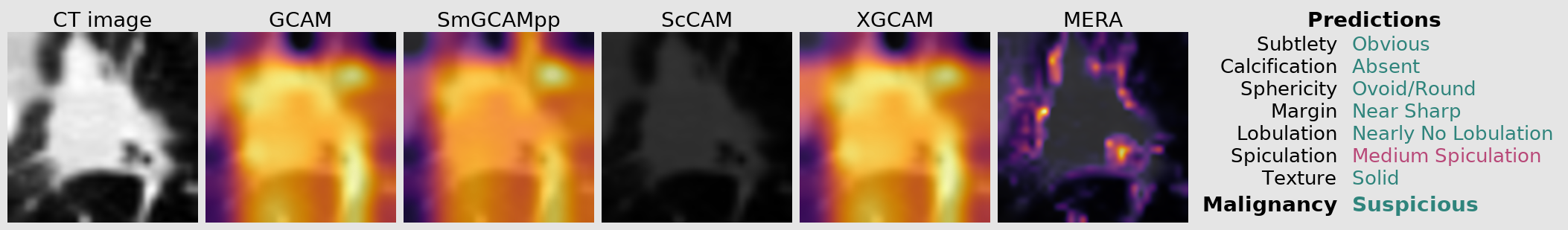}
        \label{fig:malig_1}
        }

        
    \subfloat[Compared to other algorithms which focus inadequately on only parts of the lesion, MERA effectively highlights the irregular margin, lobulation and spiculation, providing clear visual cues for malignancy.]{
        \includegraphics[width=0.9\textwidth]{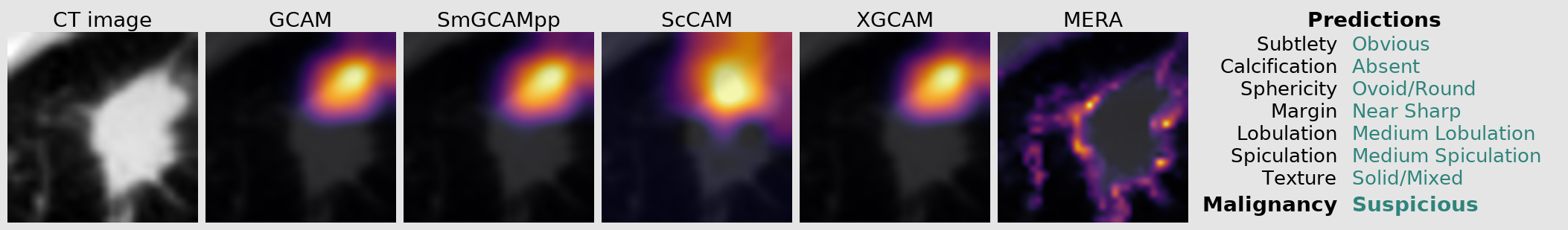}
        \label{fig:malig_5}
        }
        
    \subfloat[While other methods focus ambiguously on the high-intensity rib showing no abnormality, MERA accurately highlights the ground glass texture and irregular margins of the nodule.]{
        \includegraphics[width=0.9\textwidth]{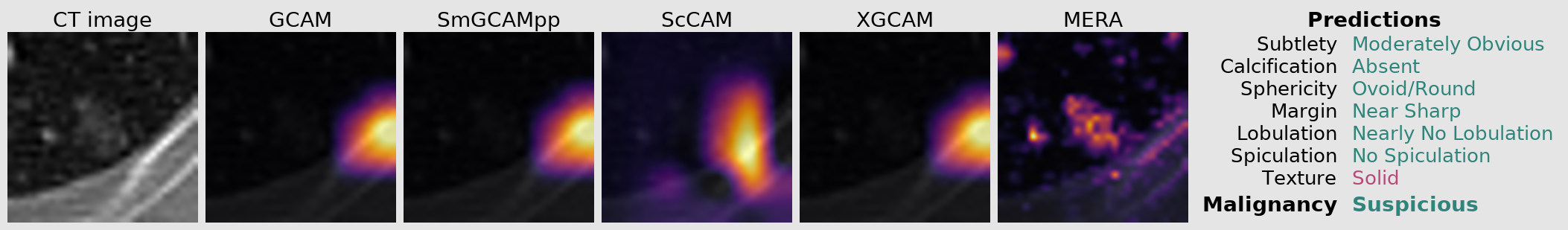}
        \label{fig:malig_3}
        }

    \subfloat[MERA proficiently highlights the heterogeneous texture and irregular margins, key indicators of malignancy.]{
        \includegraphics[width=0.9\textwidth]{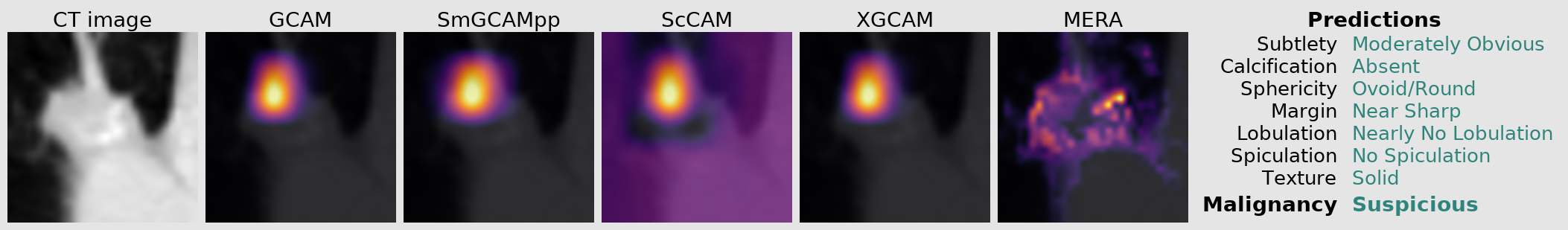}
        \label{fig:malig_4}
        }\\[-0.7ex]

    \caption{\textbf{Local visual explanation of malignant nodules.} From left to right: original lung nodule image patch on an axial chest CT, visual explanation of 4 competitor methods, visual explanation of our proposed method, our predicted nodule attributes and malignancy (green font indicates correct and red font indicates wrong, and only $1\%$ annotated data is used for the predictions). 
    Compared to competitor methods, the proposed model consistently emphasises the most diagnostically significant features indicative of malignancy.
    }
    \label{fig:res_malig}
\end{figure*}

Transitioning from the examination of benign nodules, this section delves into the visual explanation analysis of some correctly predicted malignant lung nodules. Fig.~\ref{fig:res_malig} presents cases where the proposed method successfully identifies and highlights key diagnostic features indicative of malignancy. Each example illustrates the model's capability to focus on critical characteristics such as irregular margins, heterogeneous texture, and cystic components, confirming its effectiveness and precision in malignancy detection. By exploring these correctly classified cases, we aim to demonstrate the robustness and diagnostic accuracy of the proposed model in identifying malignant nodules.

The presented case in Fig.~\ref{fig:malig_1} displays a partly solid nodule characterised by an increase in the density of the solid component, without calcification, making it a subject of malignancy suspicion. 
The proposed model not only effectively highlights areas of the spiculations or the irregularity of the margins (which are also generally indicated by GCAM\cite{selvarajuGradCAMVisualExplanations2019}, SmGCAMpp\cite{omeizaSmoothGradCAMEnhanced2019}, and XGCAM \cite{fuAxiombasedGradCAMAccurate2020}), but also uniquely pinpoints the cystic components of the nodule, evident in the posterior left highlighted area of the visual explanation map. Both of these features are critical indicators of malignant potential.
The model's demonstrated ability to focus on subtle yet significant morphological details such as the cystic components presents a more comprehensive and nuanced analysis, aligning closely with diagnostic criteria used in radiology. 


In Fig.~\ref{fig:malig_5}, the CT image illustrates a partly solid nodule with increased density in the solid component and absent calcification. The lesion displays irregular margins, medium lobulation and spiculation, all of which are associated with malignancy.
Unlike the other methods whose focus areas remain ambiguous, our MERA model comprehensively highlights the irregular margins, lobulation, and spiculation of the nodule. 
Our model's ability to distinctly emphasise and visually explain such malignancy-indicative features like directly supports the clinical decision-making process. 

Fig.~\ref{fig:malig_3} presents an axial chest CT image (bone window) of a ground glass nodule without a solid component or calcifications, raising suspicion of malignancy due to the nodule's distinctive texture and irregular margins. 
Unlike other methods which remain ambiguous and mostly focus on the nearby rib showing no abnormality (probably misled by the high intensity), the proposed model distinctly highlights the ground glass texture and the irregular margins of the nodule, which are essential markers for identifying potential malignancies. 
The ability of the proposed model to selectively emphasise these features demonstrates its practical applicability in clinical settings, providing radiologists with a targeted visual explanation.

The case in Fig.~\ref{fig:malig_4} showcases a solid spiculated nodule with a heterogeneous texture. The nodule's characteristics such as its irregular margins and potential invasion of the adjacent pleura contribute significantly to the suspicion of malignancy, making this case particularly noteworthy.
The visual explanation produced by other methods display lack of specificity and unclear focus on the nodule’s critical features. In contrast, our proposed method effectively identifies the heterogeneity and the irregular margins of the lesion, although it does not clearly demarcate the pleural invasion.
This focus on heterogeneous texture and irregular margins is vital, as these features are highly indicative of malignant processes. The ability of our model's visual explanation to highlight these aspects, despite the limitation in visualising pleural invasion, underscores its value in aiding radiologists with essential diagnostic information.

In conclusion, the analysis of correctly predicted malignant samples demonstrates the proposed model's adeptness at identifying and emphasising diagnostically significant features within lung nodules. The cases discussed in Fig.~\ref{fig:res_malig} highlight the model’s exceptional ability to discern subtle yet critical morphological details, substantially outperforming the competitor methods \cite{selvarajuGradCAMVisualExplanations2019, omeizaSmoothGradCAMEnhanced2019, wangScoreCAMScoreWeightedVisual2020, fuAxiombasedGradCAMAccurate2020}, enhancing its utility in the clinical diagnostic process. Each case confirmed the model’s precision in detecting features strongly associated with malignancy, which aligns closely with the diagnostic criteria used by radiologists \cite{macmahonGuidelinesManagementIncidental2017}. This reinforces the model’s potential as a valuable tool in the accurate diagnosis of malignant lung nodules.

\subsubsection{Incorrect predictions}

\begin{figure*}[tbp]
    \centering
    \subfloat[In an inappropriate bone window setting, while other methods are off-focus, the proposed method attempts to highlight the triangular-shaped associated lesion, possibly either due to anatomical relationship with a fissure or suspicion of lobulation, despite of the final incorrect malignancy prediction.]{
        \includegraphics[width=0.9\textwidth]{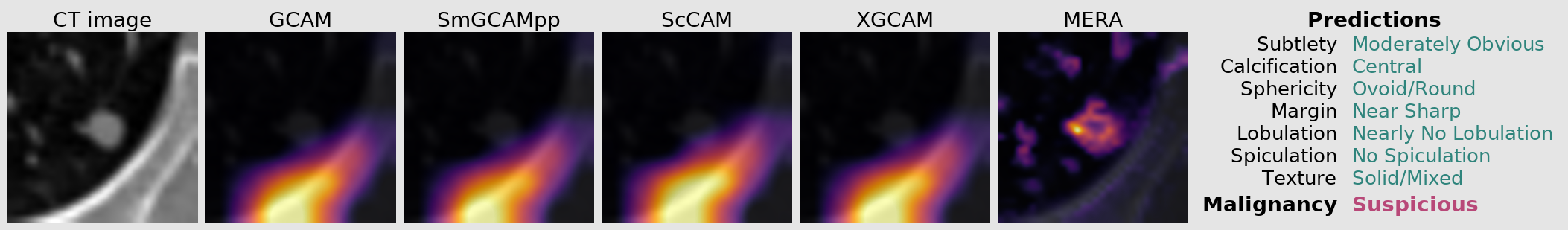}
        \label{fig:wrong_1}
        }

    \subfloat[Despite of the incorrect malignancy prediction, the proposed model accurately highlights the irregular margins and subtle lobulation of a solid nodule.]{
        \includegraphics[width=0.9\textwidth]{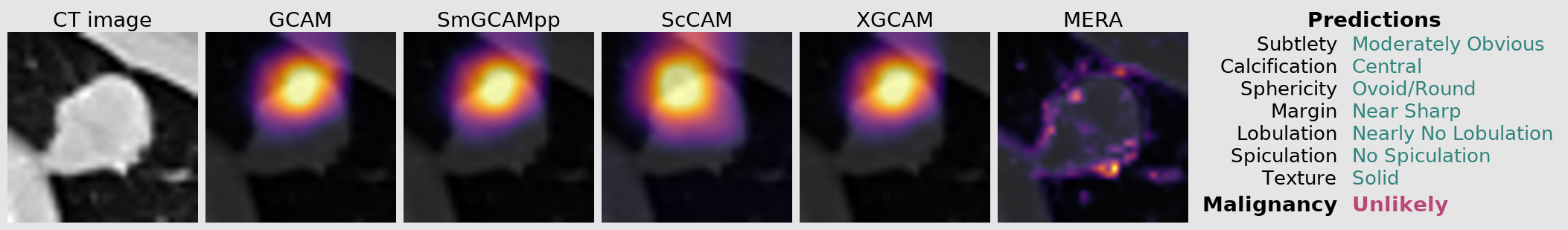}
        \label{fig:wrong_2}
        }
        
    \subfloat[The proposed model accurately focuses on the central cystic component of the nodule while neglecting the irregular margins, leading to an incorrect malignancy prediction.]{
        \includegraphics[width=0.9\textwidth]{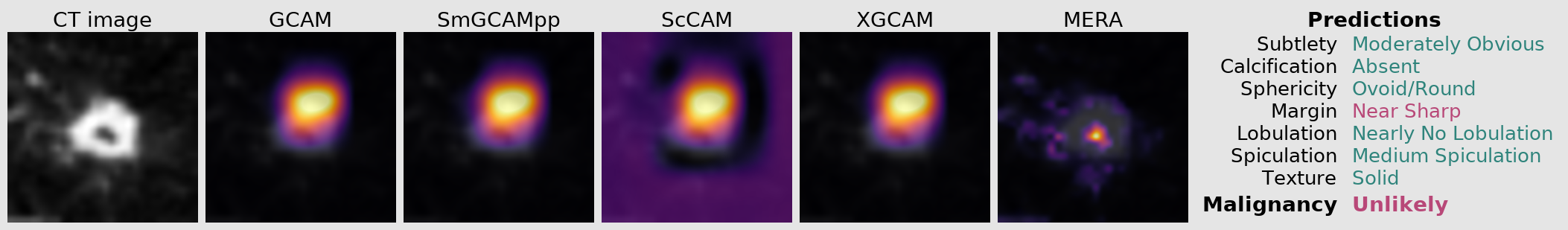}
        \label{fig:wrong_3}
        }

    \subfloat[Despite the incorrect malignancy prediction due to an erroneous assessment of lobulation, the proposed model effectively highlights the nodule’s irregular margins and bilateral linear features.]{
        \includegraphics[width=0.9\textwidth]{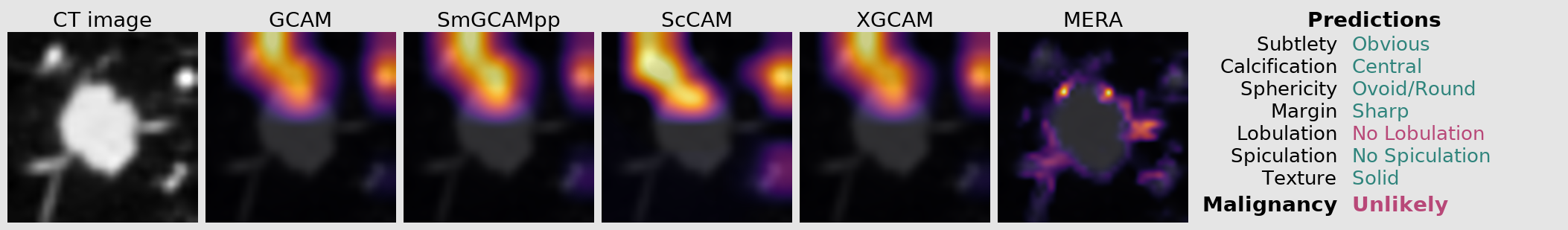}
        \label{fig:wrong_4}
        }\\[-0.7ex]

    \caption{\textbf{Local visual explanation of incorrect predictions.} From left to right: original lung nodule image patch on an axial chest CT, visual explanation of 4 competitor methods, visual explanation of our proposed method, our predicted nodule attributes and malignancy (green font indicates correct and red font indicates wrong, and only $1\%$ annotated data is used for the predictions). 
    Each subfigure underscores the model's capability to effectively highlight key morphological features, such as irregular margins, textural variations and cystic components, demonstrating its value in providing detailed visual explanations despite the incorrect final malignancy predictions.
    }
    \label{fig:res_wrong}
\end{figure*}

Following the discussion on correctly predicted malignant lung nodules, we also present cases where the proposed method provides incorrect malignancy prediction. 
This subset of samples in Fig.~\ref{fig:res_wrong}, detailed below, showcases instances where the model effectively highlighted key diagnostic features but ultimately failed in the final malignancy classification. By analysing these discrepancies, we aim to unearth insights into the model's diagnostic limitations and strengths. 

The case in Fig.~\ref{fig:wrong_1} displays a small pulmonary nodule with sharp margins and subtle intranodular calcifications, visualised in a bone window setting. The associated lesion's triangular shape may suggest a connection with a fissure, indicating benign characteristics that are challenging to assess accurately without the appropriate lung window setting \cite{macmahonGuidelinesManagementIncidental2017}.
While other methods predominantly focus on the rib bone, our method distinguishes itself by attempting to highlight features traditionally associated with malignancy, such as the suspected lobulation in the right-hand part of the nodule. However, this is mistakenly flagged due to the model interpreting the image in a non-standard bone window setting, deviating from the standard lung window typically used for assessing lung lesions.
This example underscores the model’s sensitivity to subtle features, which, although potentially leading to incorrect malignancy predictions under inappropriate conditions, still provides valuable insights through its visual explanation.

Fig.~\ref{fig:wrong_2} illustrates an axial chest CT image of a solid nodule with subtle calcified components. Notable features include irregular margins and slight lobulation, which generally raise suspicions of malignancy.
Unlike other algorithms that show unclear attention, our proposed method demonstrates a precise capability to detect and visually emphasise the nodule’s irregular margins and textural variations. 
Despite the incorrect classification of the nodule as unlikely malignant, the accurate depiction of the nodule's attributes by our method is of significant diagnostic value. It supports radiologists by providing clear visual cues about the nature of the nodule, aiding in a more informed analysis. 

The case in Fig.~\ref{fig:wrong_3} depicts a partially solid nodule with a central cystic component, suggestive of necrosis. The lesion's irregular margins, indicative of malignancy, are also a notable feature.
Our method successfully identifies and highlights the central cystic component of the nodule, a critical marker often associated with malignant lesions. In spite of the limitation in neglecting the irregular margins, this contrasts with other methods that primarily focus on the heterogeneous texture of the nodule, missing crucial details.

Fig.~\ref{fig:wrong_4} showcases an axial chest CT image of a solid nodule characterised by its significant size and irregular margins and lobulation, features that typically suggest malignancy.
Our method excels in detailing the irregular margins and bilateral linear features that might suggest infiltrative behaviour, unlike other competitor methods that generally focus on only parts of the nodule. 
Although the final classification by our model as unlikely malignant is incorrect, the ability of the model to distinctly highlight important features provides valuable visual information that can support clinical diagnosis. Such detailed visualisation can aid radiologists in scrutinising the nodule more thoroughly, potentially prompting further investigation that could lead to a correct diagnosis.

In summary, the exploration of incorrectly predicted malignant samples underscores the proposed method's ability to detail critical and substantial diagnostic features, which, although not always leading to correct malignancy predictions, significantly enrich the assistance during the diagnostic process. Each case demonstrated the model’s proficiency in highlighting relevant features such as irregular margins, lobulation, and specific patterns indicative of potential malignancies. These findings stress the importance of multimodal and multiscale explainability of the model, aiming to better integrate the visual explanations with clinical outcomes. 

\subsection{Concept explanation: from nodule attributes to malignancy diagnostics}

\begin{table*}[t]
    \centering
    \caption{\textbf{Prediction accuracy ($\%$) of nodule attributes and malignancy.}
    The best in each column is \textbf{bolded} for full/partial annotation respectively.
    Dashes (-) denote values not reported by the compared methods. 
    Results of our proposed \colorbox[HTML]{E2EFD9}{MERA} are highlighted.
    Observe that with $1\%$ annotations, MERA reaches competitive accuracy in malignancy prediction and over $90\%$ accuracy simultaneously in predicting all nodule attributes, meanwhile using the fewest nodules and no additional information.
    }
    \label{tab:res}
    \resizebox{\textwidth}{!}{%
    \begin{threeparttable}
        \begin{tabular}{lcccccccccc}
            \hline
             & \multicolumn{7}{c}{\textbf{Nodule attributes}} & \multicolumn{1}{l}{} & \multicolumn{1}{l}{} & \multicolumn{1}{l}{} \\ \cline{2-8}
             & \textbf{Sub} & \textbf{Cal} & \textbf{Sph} & \textbf{Mar} & \textbf{Lob} & \textbf{Spi} & \textbf{Tex} & \multicolumn{1}{l}{\multirow{-2}{*}{\textbf{Malignancy}}} & \multicolumn{1}{l}{\multirow{-2}{*}{\textbf{\#nodules}}} & \multicolumn{1}{l}{\multirow{-2}{*}{\begin{tabular}[c]{@{}c@{}}\textbf{No additional} \\ \textbf{information}\end{tabular}}} \\ \hline

            \multicolumn{11}{l}{Full annotation} \\
            \textbf{HSCNN}\cite{shenInterpretableDeepHierarchical2019} & 71.90 & 90.80 & 55.20 & 72.50 & - & - & 83.40 & 84.20 & 4252 & \xmark\tnote{c} \\
            \textbf{X-Caps}\cite{lalondeEncodingVisualAttributes2020} & 90.39 & - & 85.44 & 84.14 & 70.69 & 75.23 & 93.10 & 86.39 & 1149 & \cmark \\
            \textbf{MSN-JCN}\cite{chenEndtoEndMultiTaskLearning2021} & 70.77 & 94.07 & 68.63 & 78.88 & \textbf{94.75} & 93.75 & 89.00 & 87.07 & 2616 & \xmark\tnote{d} \\
            \textbf{MTMR}\cite{liuMultiTaskDeepModel2020} & - & - & - & - & - & - & - & \textbf{93.50} & 1422 & \xmark\tnote{e} \\
            \rowcolor[HTML]{E2EFD9} 
            \textbf{MERA} & \textbf{96.32$\pm$0.61} & \textbf{95.88$\pm$0.15} & \textbf{97.23$\pm$0.20} & \textbf{96.23$\pm$0.23} & 93.93$\pm$0.87 & \textbf{94.06$\pm$0.60} & \textbf{97.01$\pm$0.26} & 87.56$\pm$0.61 & \textbf{730} & \cmark \\ \hline

            \multicolumn{11}{l}{Partial annotation} \\
            \textbf{WeakSup}\cite{joshiLungNoduleMalignancy2021} \textbf{(1:5\tnote{a} )} & 43.10 & 63.90 & 42.40 & 58.50 & 40.60 & 38.70 & 51.20 & 82.40 &  &  \\
            \textbf{WeakSup}\cite{joshiLungNoduleMalignancy2021} \textbf{(1:3\tnote{a} )} & 66.80 & 91.50 & 66.40 & 79.60 & 74.30 & 81.40 & 82.20 & \textbf{89.10} & \multirow{-2}{*}{2558} & \multirow{-2}{*}{\xmark\tnote{f}} \\
            \rowcolor[HTML]{E2EFD9} 
            \textbf{MERA (10\%\tnote{b} )} & \textbf{96.23$\pm$0.45} & 92.72$\pm$1.66 & 95.71$\pm$0.47 & 90.03$\pm$3.68 & \textbf{93.89$\pm$1.41} & \textbf{93.67$\pm$0.64} & 92.41$\pm$1.05 & 87.86$\pm$1.99 & \cellcolor[HTML]{E2EFD9} & \cellcolor[HTML]{E2EFD9} \\
            \rowcolor[HTML]{E2EFD9} 
            \textbf{MERA (1\%\tnote{b} )} & 95.84$\pm$0.34 & 92.67$\pm$1.24 & \textbf{95.97$\pm$0.45} & 91.03$\pm$4.65 & 93.54$\pm$0.87 & 92.72$\pm$1.19 & 92.67$\pm$1.50 & 86.22$\pm$2.51 & \multirow{-2}{*}{\cellcolor[HTML]{E2EFD9}\textbf{730}} & \multirow{-2}{*}{\cellcolor[HTML]{E2EFD9}\cmark} \\ \hline
        \end{tabular}%
        \begin{tablenotes}
            \item[a] $1:N$ indicates that $\frac{1}{1+N}$ of training samples have annotations on nodule attributes. (All samples have malignancy annotations.)
            \item[b] The proportion of training samples that have annotations on nodule attributes and malignancy.
            \item[c] 3D volume data are used.
            \item[d] Segmentation masks and nodule diameter information are used. Two other traditional methods are used to assist training.
            \item[e] All 2D slices in 3D volumes are used.
            \item[f] Multi-scale 3D volume data are used.
        \end{tablenotes}
    \end{threeparttable}
    }
    \vspace{-0.7ex}
\end{table*}

\begin{figure}[tbp]
    \vspace{-1ex}
    \centering
    \subfloat[Full annotation]{
        \includegraphics[width=0.47\textwidth]{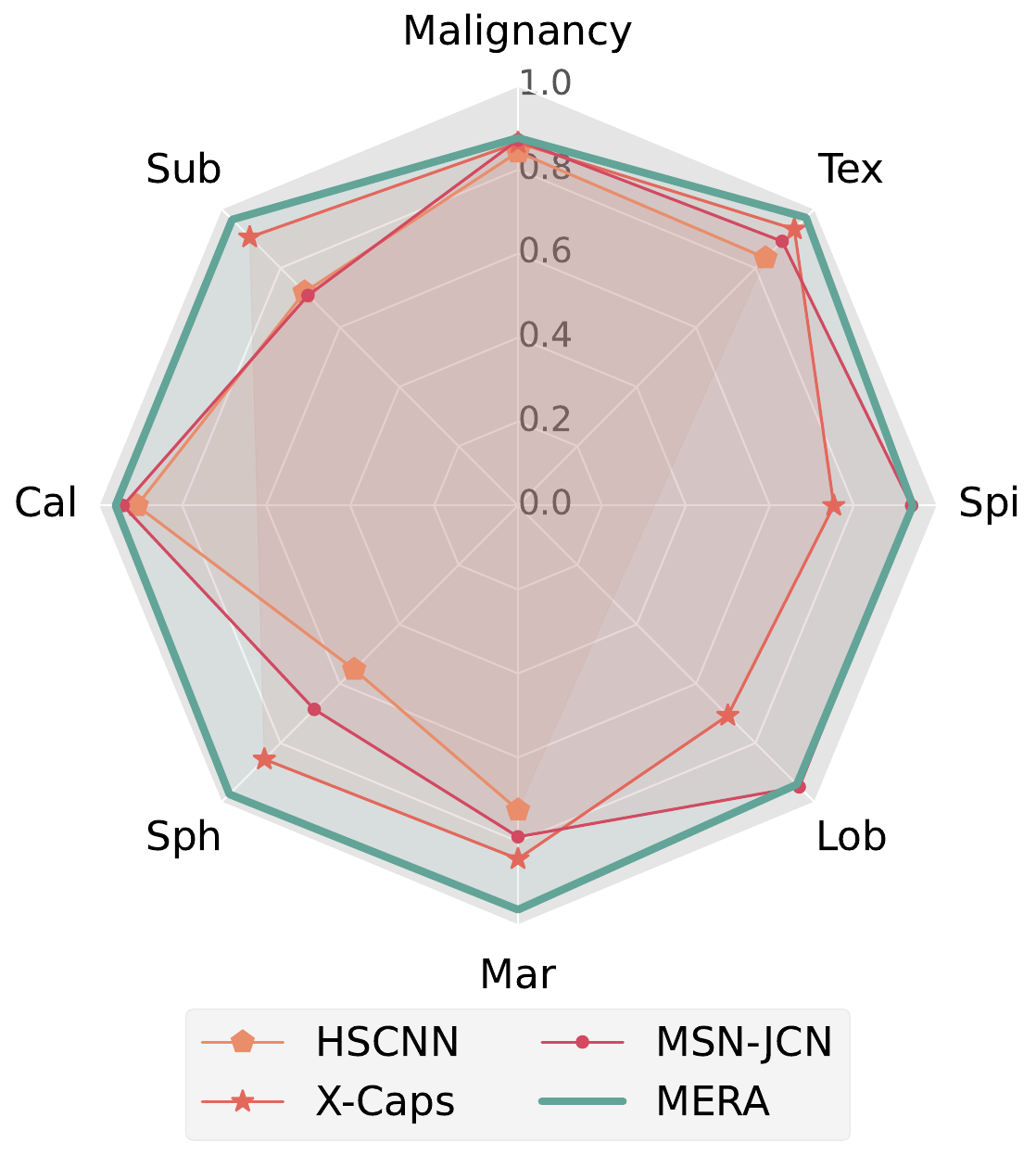}
        \label{fig:radar_a}}\hfil
    \subfloat[Partial annotation]{
        \includegraphics[width=0.47\textwidth]{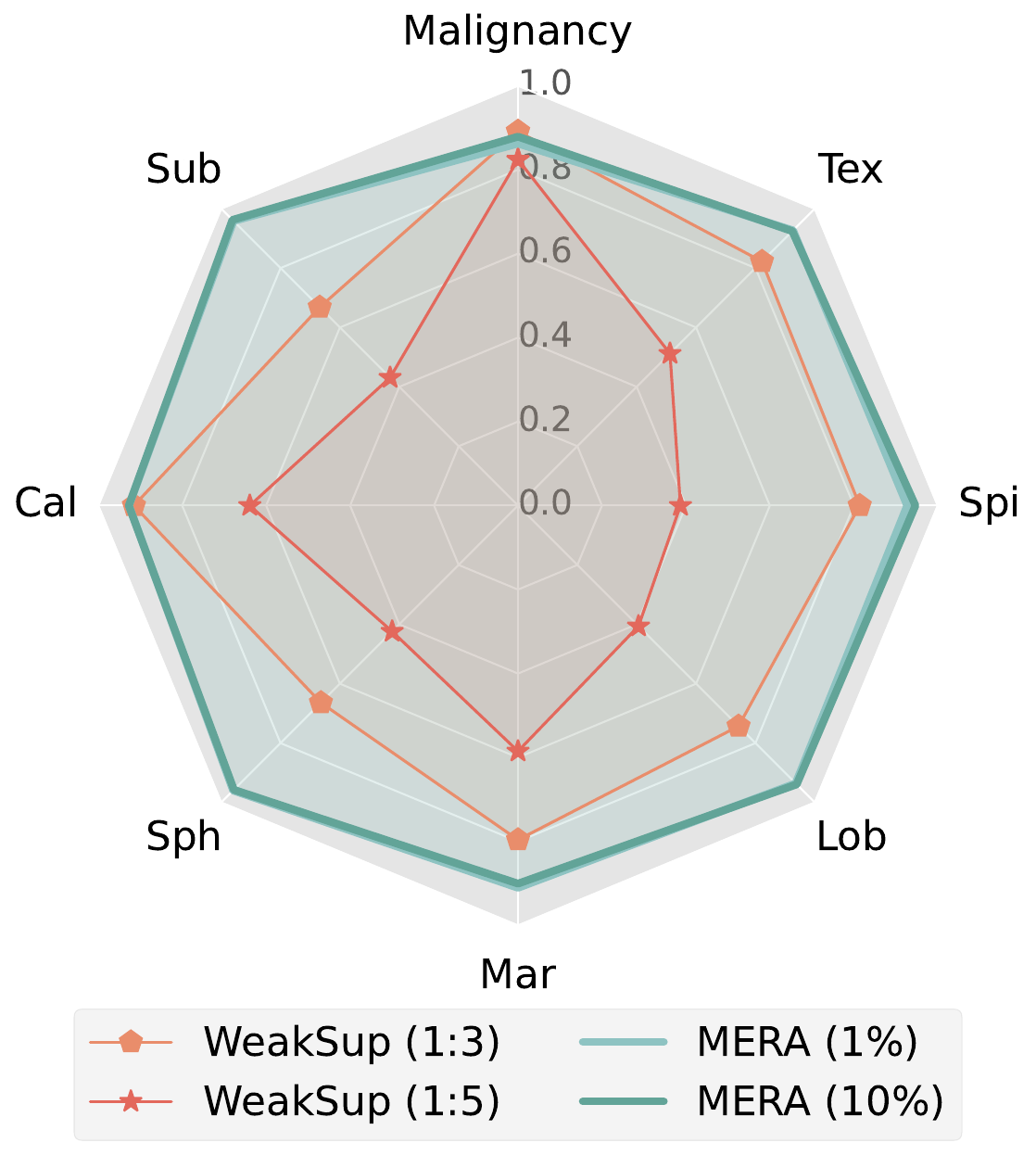}
        \label{fig:radar_b}}\\[-0.7ex]
    \caption{\textbf{Performance comparison}, in terms of prediction accuracy ($\%$) of nodule attributes and malignancy.
    Observe that \textcolor[HTML]{2F847C}{MERA} achieves simultaneously high accuracy in predicting malignancy and all nodule attributes, regardless of using either full or partial annotations.
    }
    \label{fig:radar}
\end{figure}

\begin{figure}[tbp]
    \centering
    \includegraphics[width=0.8\textwidth]{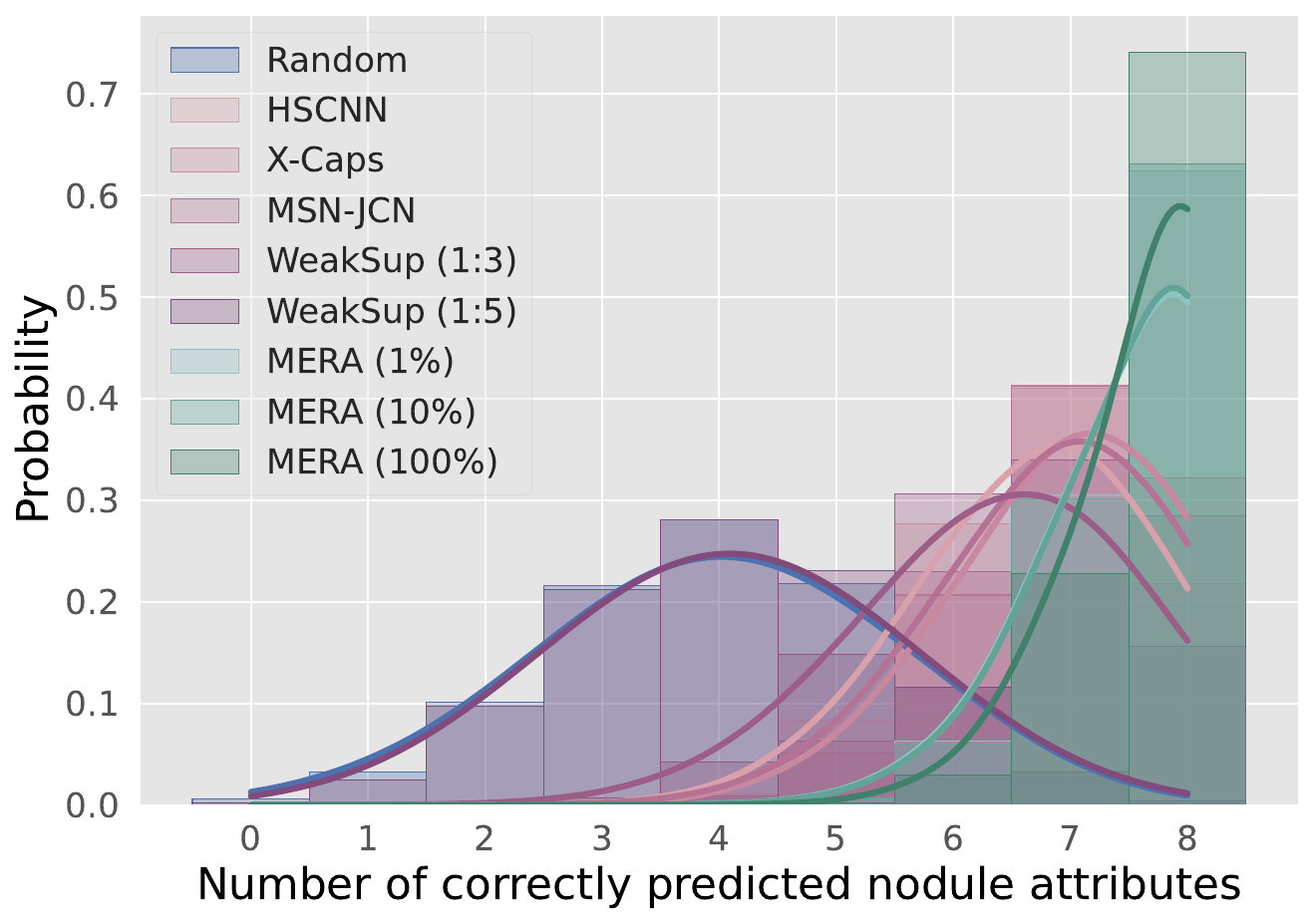}

    \caption{\textbf{Probabilities of correctly predicting a certain number of attributes for a given nodule sample.}
    Observe that \textcolor[HTML]{2F847C}{MERA} shows a more prominent probability of simultaneously predicting all $8$ nodule attributes correctly.
    }
    \label{fig:res_prob_ftr}
\end{figure}

The performance is evaluated quantitatively in terms of prediction accuracy of nodule attributes and malignancy, 
with the evaluation procedure as in Sec.~\ref{sec:res_knn}.

Tab.~\ref{tab:res} summarises the overall prediction performance and compares it with the state-of-the-art. 
The results show that when using only $518$ among the $730$ nodule samples and $1\%$ of their annotations for training, MERA reaches over $90\%$ accuracy simultaneously in predicting all nodule attributes, which outperforms all previous works. 
Meanwhile, regarding nodule malignancy prediction accuracy, MERA performs comparably with X-Caps\cite{lalondeEncodingVisualAttributes2020} and already better than HSCNN \cite{shenInterpretableDeepHierarchical2019}, which uses 3D volume data.
When using $10\%$ annotations, our malignancy prediction accuracy surpasses all other explainable competitors using full annotations, among which MSN-JCN\cite{chenEndtoEndMultiTaskLearning2021} is heavily supervised by additional information. 

The visualisation of the performance comparison is shown in Fig.~\ref{fig:radar}. It can be observed that our approach demonstrates simultaneously high prediction accuracy in malignancy and all nodule attributes. This substantially increases the model's trustworthiness and has not been achieved in previous works. 

In addition, we also calculate the probabilities of correctly predicting a certain number of attributes for a given nodule sample, as shown in Fig.~\ref{fig:res_prob_ftr}. The probabilities are calculated from Tab.~\ref{tab:res}. To not underestimate the performance of other compared methods, their not reported values are all assumed to be $100\%$ accuracy. 
It can be seen that MERA demonstrates a more prominent probability of simultaneously predicting all $8$ nodule attributes correctly. The probability of correctly predicting at least $7$ attributes is higher than $90\%$, even under the extreme $1\%$ annotation condition. In contrast, WeakSup(1:5)\cite{joshiLungNoduleMalignancy2021}, despite achieving $82.4\%$ accuracy in malignancy prediction, shows no significant difference in predicting nodule attributes compared to random guessing, which we consider to be the opposite of trustworthiness.

\subsection{Ablation study}
\label{sec:res_ablation}

We validate our proposed training strategy by the following ablation experiments.  

\subsubsection{Validation of unsupervised feature extraction}

\begin{table}[tbp]
    \vspace{-1ex}
    \centering
    \caption{\textbf{Validation of Stage 1 training}, evaluated by malignancy prediction accuracy ($\%$). All annotations are used during training. The result of our proposed setting is \colorbox[HTML]{DFE7F3}{highlighted}.
    Only our proposed setting and conventional end-to-end trained CNN achieve higher than $85\%$ accuracy. 
    }
    \label{tab:res_ablation_vit}
    \resizebox{\textwidth}{!}{%
    \begin{threeparttable}
        \begin{tabular}{ccccc}
        \hline
        \textbf{Arch} &
          \textbf{\#params} &
          \textbf{\begin{tabular}[c]{@{}c@{}}Training\\ strategy\end{tabular}} &
          \textbf{\begin{tabular}[c]{@{}c@{}}ImageNet\\ pretrain\end{tabular}} &
          \textbf{Acc} \\ \hline
        \multirow{3}{*}{ResNet-50} & \multirow{3}{*}{23.5M} & \textcolor[HTML]{B84878}{end-to-end} & \textcolor[HTML]{B84878}{\xmark} & \textcolor[HTML]{B84878}{88.08\tnote{$\ast$}} \\
                                   &                        & two-stage  & \xmark & 70.48 \\
                                   &                        & two-stage  & \cmark & 70.48 \\
        \cellcolor[HTML]{DFE7F3}   & \cellcolor[HTML]{DFE7F3} & end-to-end & \xmark & 64.24 \\
        \cellcolor[HTML]{DFE7F3}   & \cellcolor[HTML]{DFE7F3} & two-stage  & \xmark & 79.19 \\
        \rowcolor[HTML]{DFE7F3}
        \multirow{-3}{*}{\cellcolor[HTML]{DFE7F3}{ViT}} & \multirow{-3}{*}{\cellcolor[HTML]{DFE7F3}{21.7M}} & two-stage  & \cmark & 87.56 \\ \hline
        \end{tabular}%
        \begin{tablenotes}
            \item[$\ast$] This is a representative setting and performance of previous works using CNN architecture.
        \end{tablenotes}
    \end{threeparttable}
    }
    \vspace{-0.7ex}
\end{table}

We validate our unsupervised feature extraction approach in Stage 1 (\ref{sec:method_recap}) by comparing with different architectures for encoders $E$, training strategies, and whether to use ImageNet-pretrained weights. The results in Tab.~\ref{tab:res_ablation_vit} show that ViT architecture benefits more from the self-supervised contrastive training compared to ResNet-50 as a CNN representative. This observation is in accord with the findings in \cite{chenEmpiricalStudyTraining2021, caronEmergingPropertiesSelfSupervised2021}. ViT's lowest accuracy in end-to-end training reiterates its requirement for a large amount of training data \cite{dosovitskiyImageWorth16x162020}. Starting from the ImageNet-pretrained weights is also shown to be helpful for ViT but not ResNet-50, probably due to ViT's lack of inductive bias needs far more than hundreds of training samples to compensate \cite{dosovitskiyImageWorth16x162020}, especially for medical images.
In summary, only the proposed approach and conventional end-to-end training of ResNet-50 achieve higher than $85\%$ accuracy of malignancy prediction.

\subsubsection{Validation of weakly supervised hierarchical prediction}

\begin{table}[tbp]
    \vspace{-1ex}
    \centering
    \caption{\textbf{Ablation study of proposed components in Stage 2,} evaluated by the malignancy prediction accuracy using $10\%$ and $1\%$ annotations. The best in each column is \textbf{bolded}. Settings of the \colorbox[HTML]{DFE7F3}{plain supervised} approach following Stage 1 and our proposed \colorbox[HTML]{E2EFD9}{MERA} are highlighted.
    }
    \label{tab:res_ablation_exploitation}
    \resizebox{\textwidth}{!}{%
    \begin{threeparttable}
        \begin{tabular}{cccccc}
        \hline
         &  &  &  & \multicolumn{2}{c}{\textbf{Maligancy accuracy}} \\ \cline{5-6} 
        \multirow{-2}{*}{\textbf{\begin{tabular}[c]{@{}c@{}}Seed sample\\ selection\end{tabular}}} & \multirow{-2}{*}{\textbf{\begin{tabular}[c]{@{}c@{}}Annotation \\ acquisition strategy\end{tabular}}} & \multirow{-2}{*}{\textbf{\begin{tabular}[c]{@{}c@{}}Pseudo \\ labelling\end{tabular}}} & \multirow{-2}{*}{\textbf{Quenching}} & \textbf{(10\%)} & \textbf{(1\%)}\tnote{$\ast$} \\ \hline
        \rowcolor[HTML]{DFE7F3} 
        random & \xmark & \xmark & \xmark & 86.65$\pm$1.39 & 80.02$\pm$8.56 \\
        random & \textbf{malignancy confidence} & \textbf{dynamic} & \textbf{\cmark} & 82.71$\pm$7.47 & 79.50$\pm$11.10 \\
        \textbf{sparse} & integrated entropy & \textbf{dynamic} & \textbf{\cmark} & 86.52$\pm$0.99 & \textbf{86.22$\pm$2.51} \\
        \textbf{sparse} & \textbf{malignancy confidence} & static & \xmark & 85.91$\pm$1.66 & 85.35$\pm$1.93 \\
        \rowcolor[HTML]{E2EFD9} 
        \textbf{sparse} & \textbf{malignancy confidence} & \textbf{dynamic} & \textbf{\cmark} & \textbf{87.86$\pm$1.99} & \textbf{86.22$\pm$2.51} \\ \hline
        \end{tabular}%
        \begin{tablenotes}
            \item[$\ast$] Does not contain requested annotations.
        \end{tablenotes}
    \end{threeparttable}
    }
    \vspace{-0.7ex}
\end{table}

We validate the proposed annotation exploitation mechanism in Stage 2 (Sec.~\ref{sec:method_annoexp}) by ablating each component as a row shown in Tab.~\ref{tab:res_ablation_exploitation}. The standard deviation when using $1\%$ annotations shows that sparse seeding plays a crucial role in stabilising performance. The sum entropy\cite{settlesUncertaintySampling2012} integrating malignancy and all nodule attributes was also experimented with as an alternative acquisition strategy, but exhibits impaired prediction accuracy. Quenching, which enables dynamic pseudo labelling, also proves necessary for the boosted performance. 


\begin{figure}[tbp]
    \vspace{-1ex}
    \centering
    \includegraphics[width=0.8\textwidth]{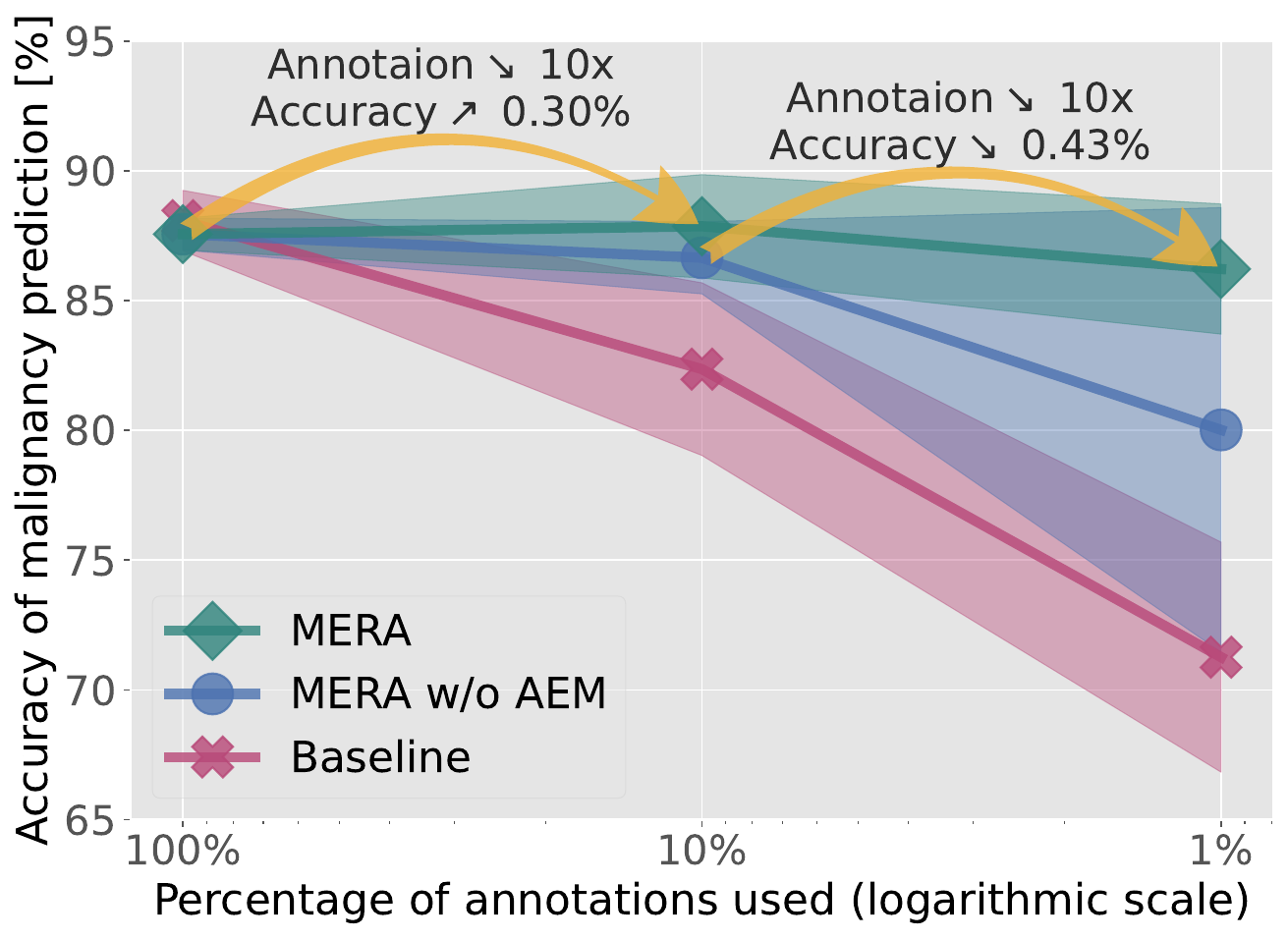}

    \caption{
    \textbf{Influence of annotation reduction} on \textcolor[HTML]{2F847C}{MERA}, \textcolor[HTML]{4C72B0}{MERA without annotation exploitation mechanism (AEM)} and a representative \textcolor[HTML]{B84878}{CNN baseline}
    , in terms of nodule malignancy prediction accuracy.
    With the proposed AEM, \textcolor[HTML]{2F847C}{MERA} achieves comparable or even higher accuracy with 10x fewer annotations, meanwhile being more robust.
    }
    \label{fig:res_annoreduce}
    \vspace{-0.7ex}
\end{figure}

To further illustrate the effect of the proposed annotation exploitation mechanism (AEM) in Stage 2 (Sec.~\ref{sec:method_annoexp}), we plot the malignancy prediction accuracy as annotations are reduced on a logarithmic scale.
As shown in Fig.~\ref{fig:res_annoreduce}, MERA demonstrates strong robustness w.r.t. annotation reduction.
When annotations reach only $1\%$, the average accuracy of the end-to-end trained CNN baseline model decreases rapidly to $71.26\%$. 
On the other hand, MERA without AEM suffer from the high instability, despite of the its higher average accuracy.
In contrast, with the proposed AEM, MERA achieves comparable or even higher malignancy prediction accuracy with $10$x fewer annotations, meanwhile being significantly more robust under the condition of $1\%$ annotations.

\section{Conclusions}
\label{sec:conclustions}

In this paper, we introduced MERA, a multimodal and multiscale self-explanatory model that significantly reduces the annotation burden required for lung nodule diagnosis. MERA not only maintains high diagnostic accuracy but also provides comprehensive, intrinsic explanations. Through rigorous experimentation and analysis, we have derived several key insights:

\begin{itemize}[label=\textbullet]
    \item MERA demonstrates superior or comparable accuracy in malignancy prediction using only $1\%$ annotated samples, compared with state-of-the-art methods using full annotation. This substantial reduction in annotation requirements establishes MERA as a practical solution in clinical settings where annotated data is scarce.
    \item The clustering in the learned latent space reveals underlying correlations between nodule attributes and malignancy. This alignment with clinical knowledge enhances the trustworthiness of model's decisions. The separability of data point in the latent space underpins the exhibited prediction performance.
    \item MERA contextualises diagnostic decisions by presenting cases similar to the current subject, aiding radiologists in understanding the model’s reasoning through comparable instances.
    \item Local visual explanations provided by MERA highlight critical diagnostic features such as irregular margins, heterogeneous texture, and cystic components. These visual insights align closely with clinical practices and diagnostic guidelines, significantly enriching the assistance during the diagnostic process, even under the circumstances of occasional incorrect malignancy prediction.
    \item MERA unprecedentedly achieves over $90\%$ accuracy simultaneously in predicting all nodule attributes when trained with only hundreds of samples and $1\%$ of their annotations, significantly enhancing the model's trustworthiness. 
    \item The SSL techniques employed in MERA effectively leverage unlabelled data, which helps in reducing the dependency on annotations. The proposed annotation exploitation mechanism, including sparse seeding and dynamic pseudo labelling with quenching, notably enhances training robustness and model performance.
    \item MERA integrates multimodal and multiscale explanations directly into its decision-making process, contrasting with traditional post-hoc methods. This intrinsic approach to explanation ensures more reliable and transparent diagnostic outcomes.
\end{itemize}

Overall, MERA represents a significant advancement in explainable artificial intelligence for lung nodule diagnosis. Its ability to achieve high diagnostic accuracy with very limited annotations, coupled with its robust and transparent decision-making process, makes it a valuable tool for enhancing early lung cancer detection and improving patient outcomes. Future efforts will focus on further optimising the model, exploring its applicability to other medical imaging tasks, and integrating it into clinical workflows for real-world validation.

\vfill\pagebreak


\section*{Declaration of generative AI and AI-assisted technologies in the writing process}

During the preparation of this work the author(s) used ChatGPT in order to polish the language for better readability. After using this tool/service, the author(s) reviewed and edited the content as needed and take(s) full responsibility for the content of the published article.

 \bibliographystyle{elsarticle-num} 
 \bibliography{XAI}




\vfill\pagebreak

\appendix

\section{Appendix}
\label{sec:appendix}


\begin{figure*}[p]
    \centering
    \includegraphics[width=0.9\textwidth]{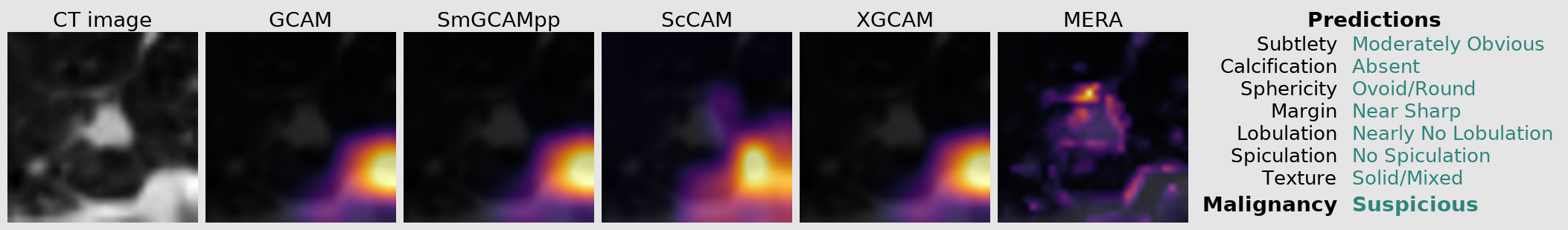} \\[-0.25ex]
    \includegraphics[width=0.9\textwidth, trim=0ex 0ex 0ex 10ex, clip]{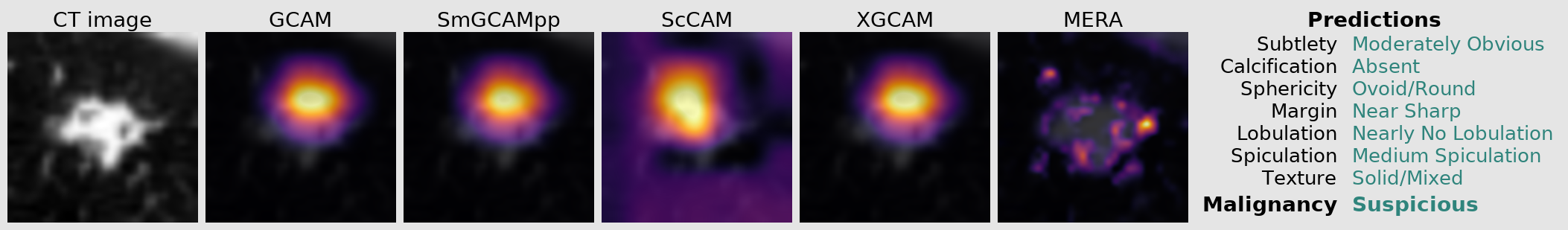} \\[-0.25ex]
    \includegraphics[width=0.9\textwidth, trim=0ex 0ex 0ex 10ex, clip]{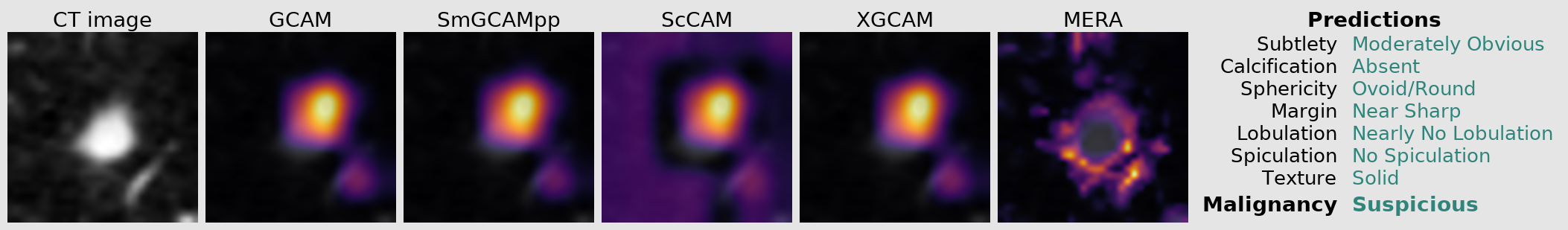} \\[-0.25ex]
    \includegraphics[width=0.9\textwidth, trim=0ex 0ex 0ex 10ex, clip]{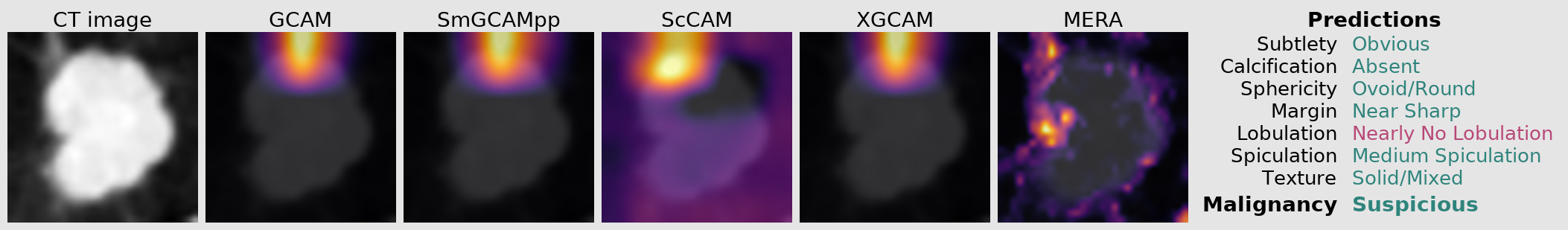} \\[-0.25ex]
    \includegraphics[width=0.9\textwidth, trim=0ex 0ex 0ex 10ex, clip]{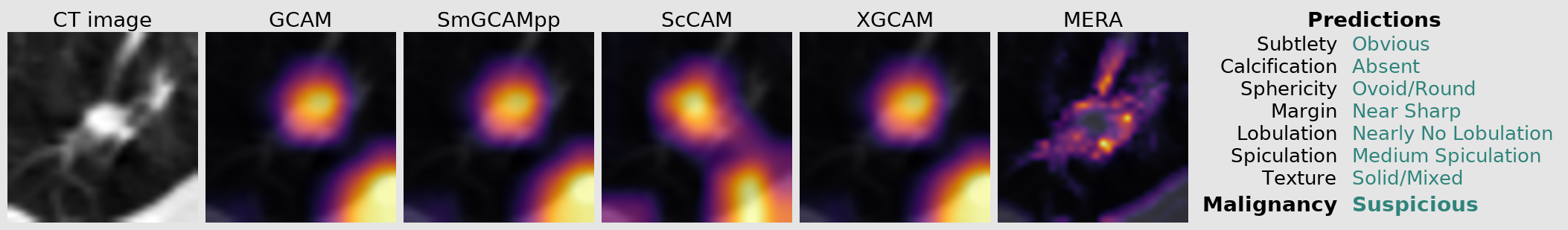} \\[-0.25ex]
    \includegraphics[width=0.9\textwidth, trim=0ex 0ex 0ex 10ex, clip]{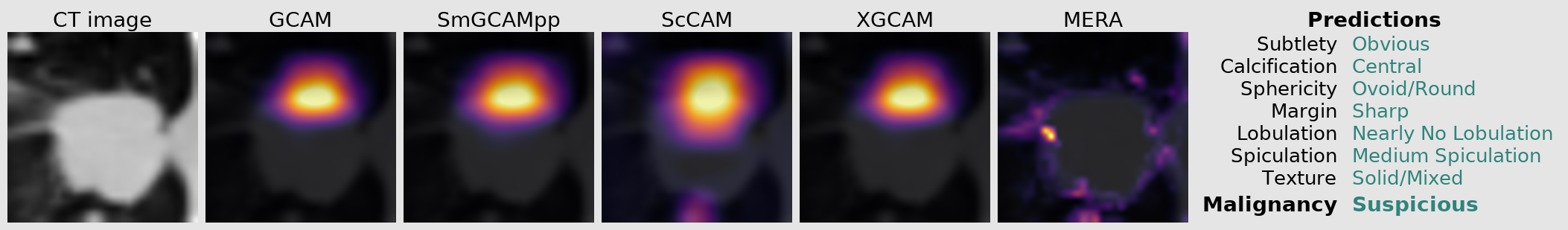} \\[-0.25ex]
    \includegraphics[width=0.9\textwidth, trim=0ex 0ex 0ex 10ex, clip]{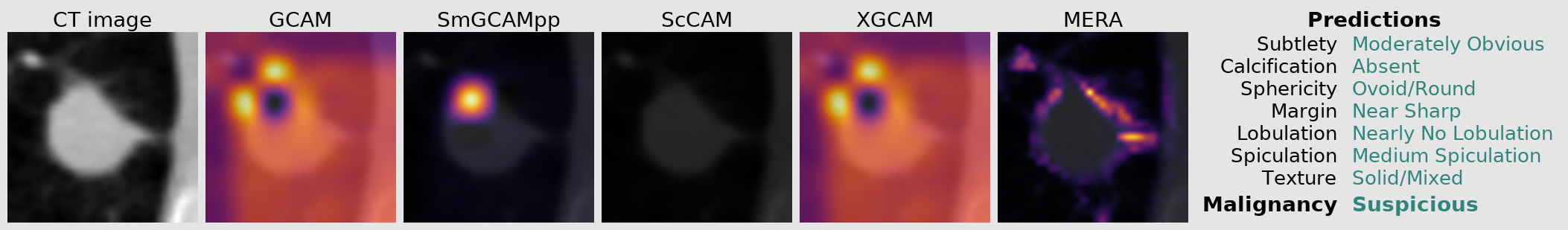} \\[-0.25ex]
    \includegraphics[width=0.9\textwidth, trim=0ex 0ex 0ex 10ex, clip]{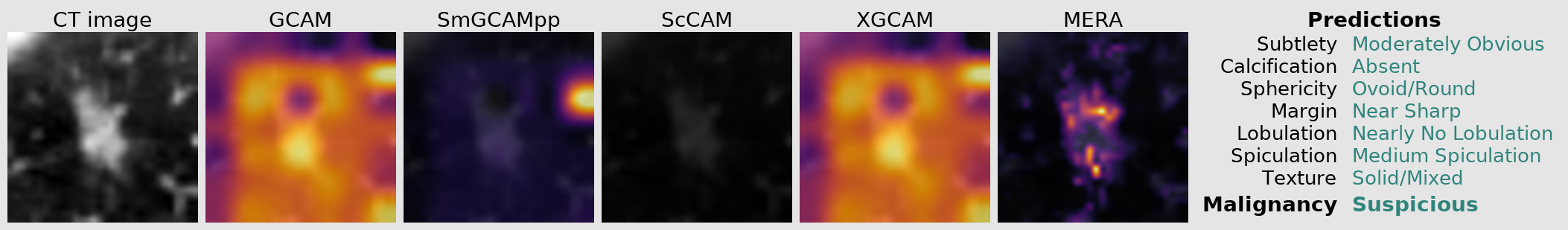} \\[-0.25ex]
    \includegraphics[width=0.9\textwidth, trim=0ex 0ex 0ex 10ex, clip]{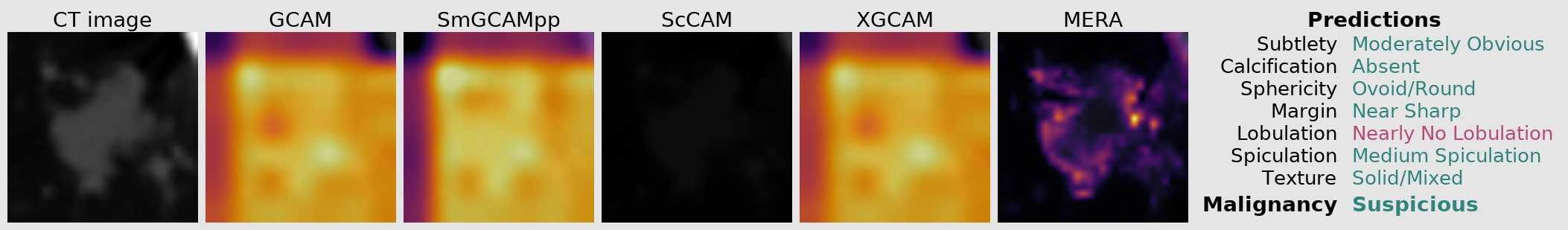} \\[-0.25ex]
    \includegraphics[width=0.9\textwidth, trim=0ex 0ex 0ex 10ex, clip]{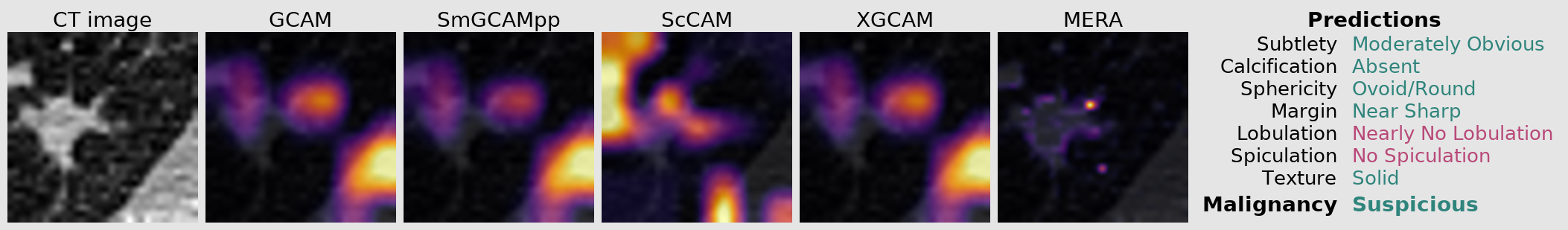}

    \caption{Samples of \textbf{true positive} malignancy predictions with local visual explanation and concept explanation \textbf{(not cherry-picked)}. 
    From left to right: original lung nodule image patch on an axial chest CT, visual explanation of 4 competitor methods, visual explanation of our proposed method, our predicted nodule attributes and malignancy (green font indicates correct and red font indicates wrong, and only $1\%$ annotated data is used for the predictions).
    }
    \label{fig:res_TP1}
\end{figure*}

\begin{figure*}[p]
    \centering
    \includegraphics[width=0.9\textwidth]{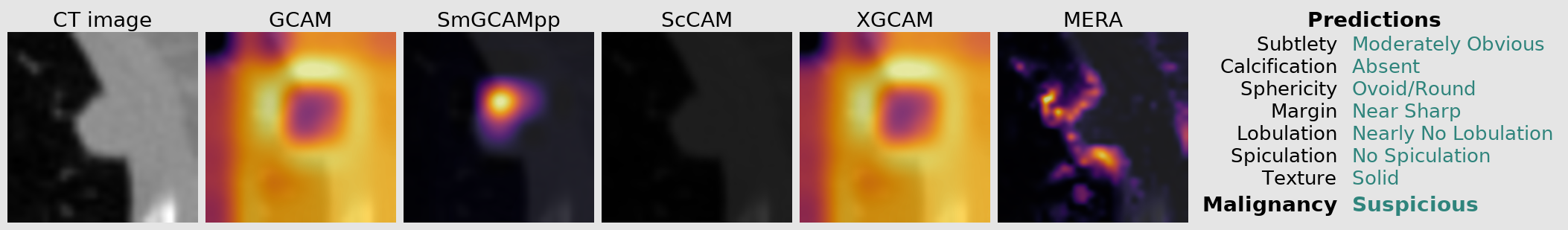} \\[-0.25ex]
    \includegraphics[width=0.9\textwidth, trim=0ex 0ex 0ex 10ex, clip]{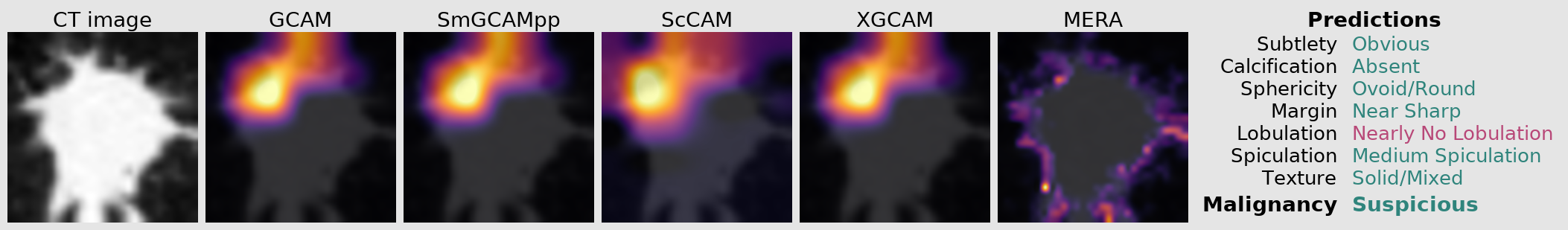} \\[-0.25ex]
    \includegraphics[width=0.9\textwidth, trim=0ex 0ex 0ex 10ex, clip]{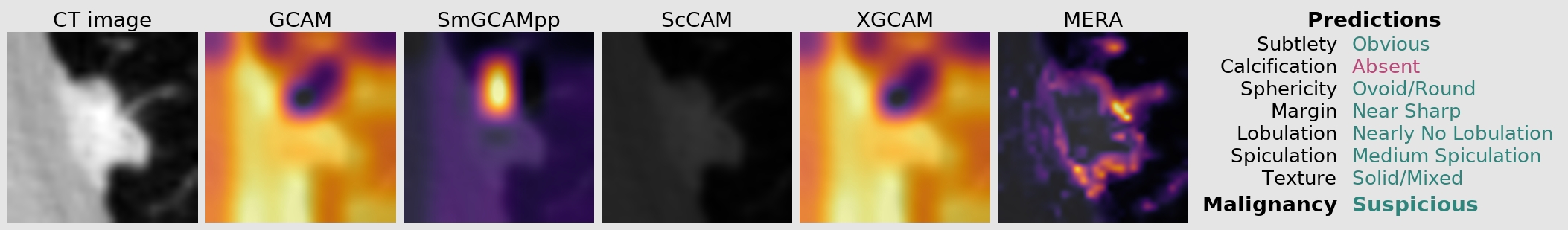} \\[-0.25ex]
    \includegraphics[width=0.9\textwidth, trim=0ex 0ex 0ex 10ex, clip]{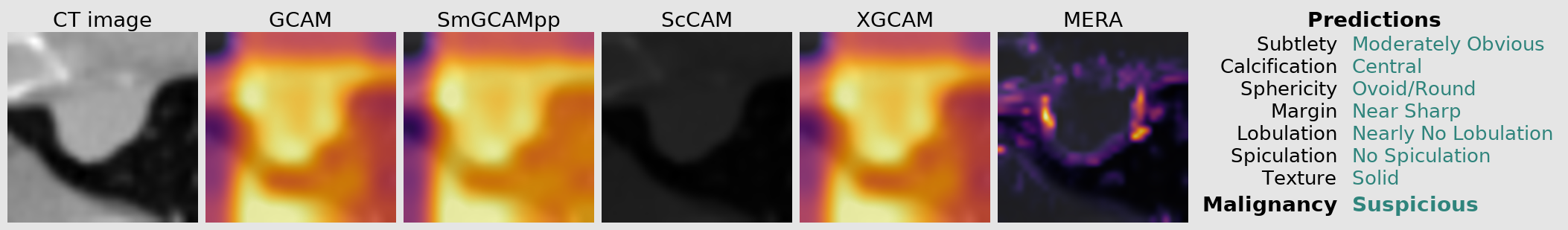} \\[-0.25ex]
    \includegraphics[width=0.9\textwidth, trim=0ex 0ex 0ex 10ex, clip]{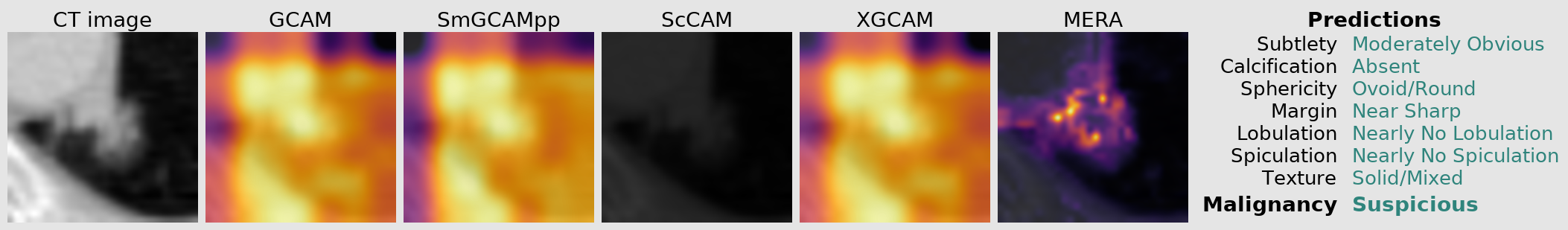} \\[-0.25ex]
    \includegraphics[width=0.9\textwidth, trim=0ex 0ex 0ex 10ex, clip]{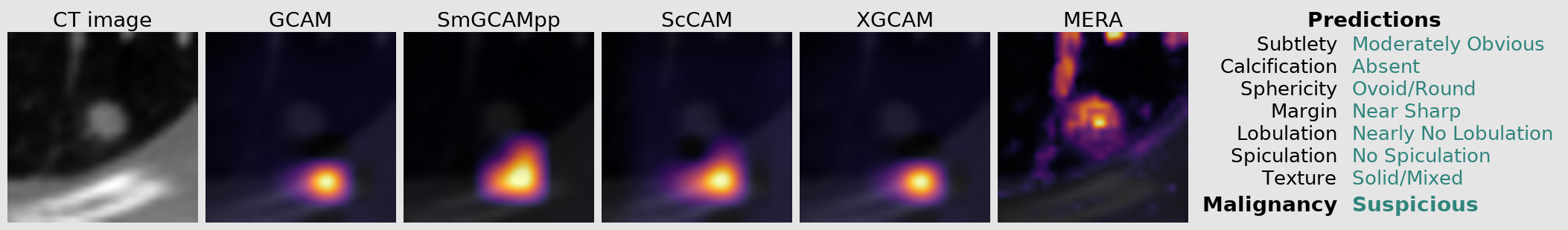} \\[-0.25ex]
    \includegraphics[width=0.9\textwidth, trim=0ex 0ex 0ex 10ex, clip]{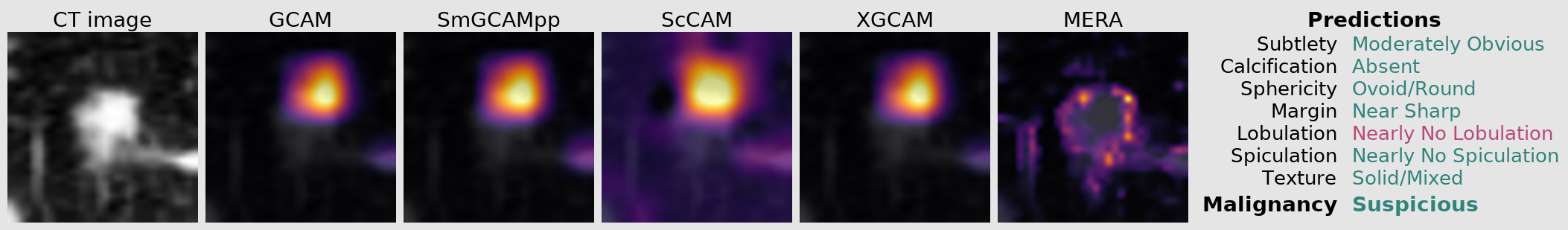} \\[-0.25ex]
    \includegraphics[width=0.9\textwidth, trim=0ex 0ex 0ex 10ex, clip]{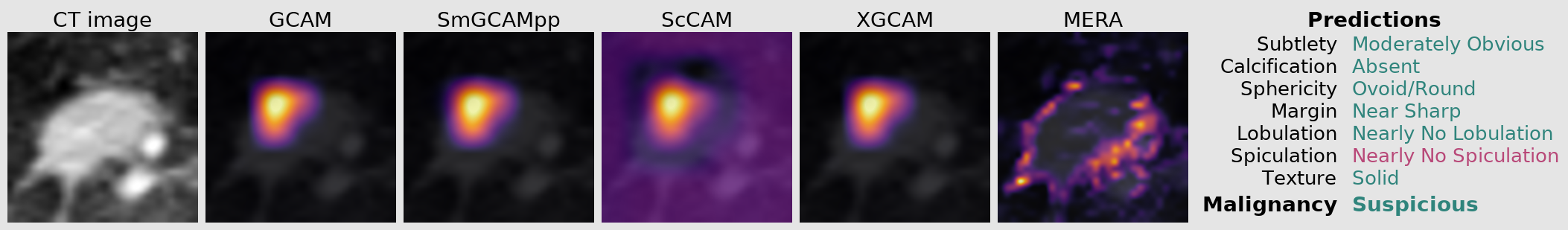} \\[-0.25ex]
    \includegraphics[width=0.9\textwidth, trim=0ex 0ex 0ex 10ex, clip]{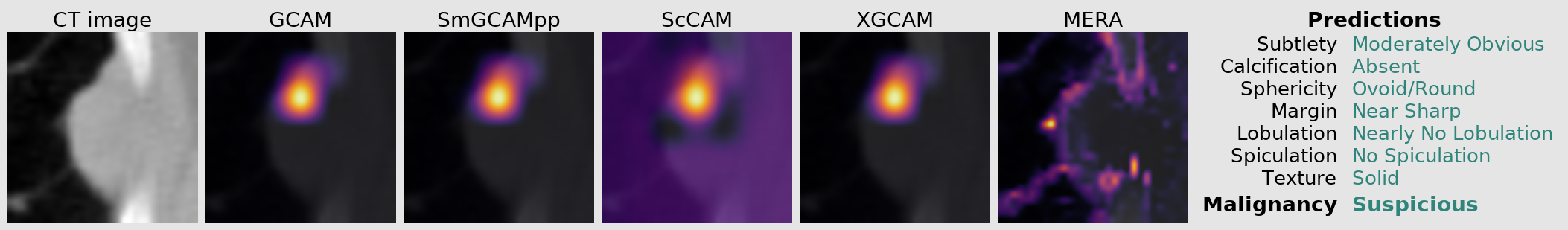} \\[-0.25ex]
    \includegraphics[width=0.9\textwidth, trim=0ex 0ex 0ex 10ex, clip]{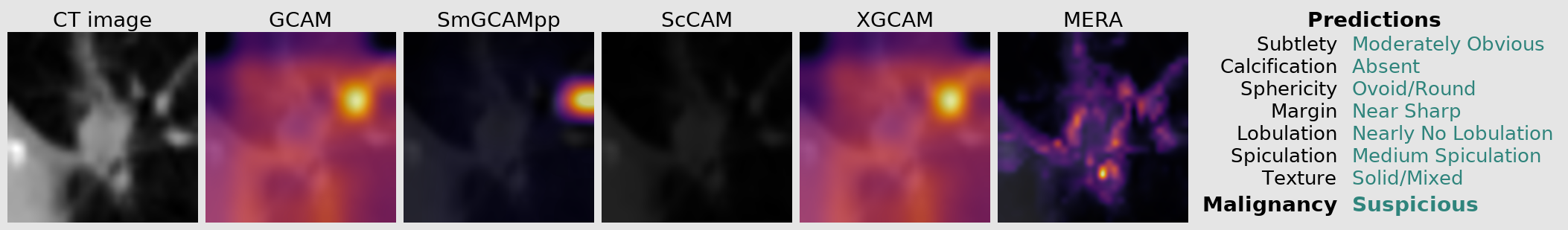}

    \caption{Samples of \textbf{true positive} malignancy predictions with local visual explanation and concept explanation \textbf{(not cherry-picked)}. 
    From left to right: original lung nodule image patch on an axial chest CT, visual explanation of 4 competitor methods, visual explanation of our proposed method, our predicted nodule attributes and malignancy (green font indicates correct and red font indicates wrong, and only $1\%$ annotated data is used for the predictions).
    }
    \label{fig:res_TP2}
\end{figure*}

\begin{figure*}[p]
    \centering
    \includegraphics[width=0.9\textwidth]{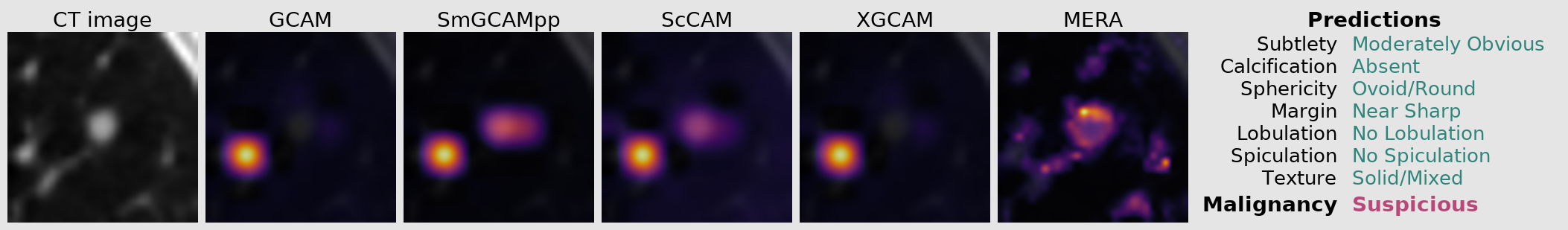} \\[-0.25ex]
    \includegraphics[width=0.9\textwidth, trim=0ex 0ex 0ex 10ex, clip]{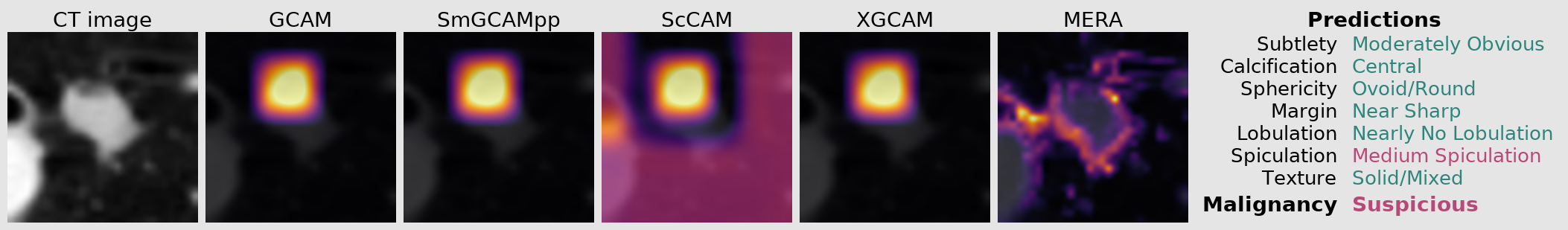} \\[-0.25ex]
    \includegraphics[width=0.9\textwidth, trim=0ex 0ex 0ex 10ex, clip]{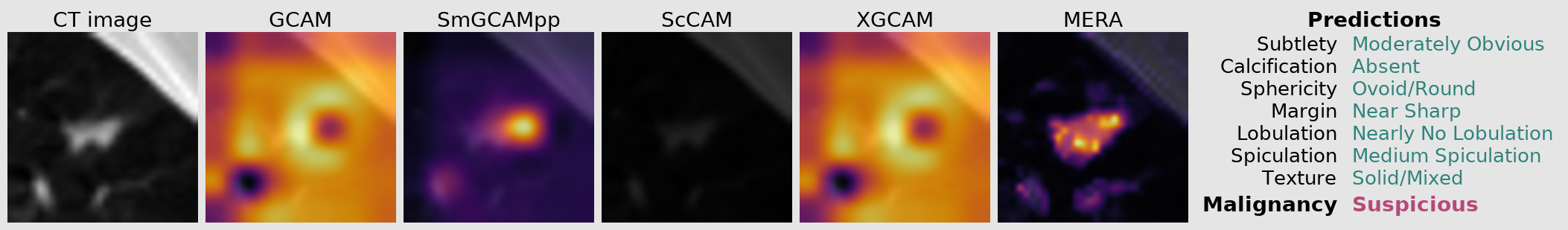} \\[-0.25ex]
    \includegraphics[width=0.9\textwidth, trim=0ex 0ex 0ex 10ex, clip]{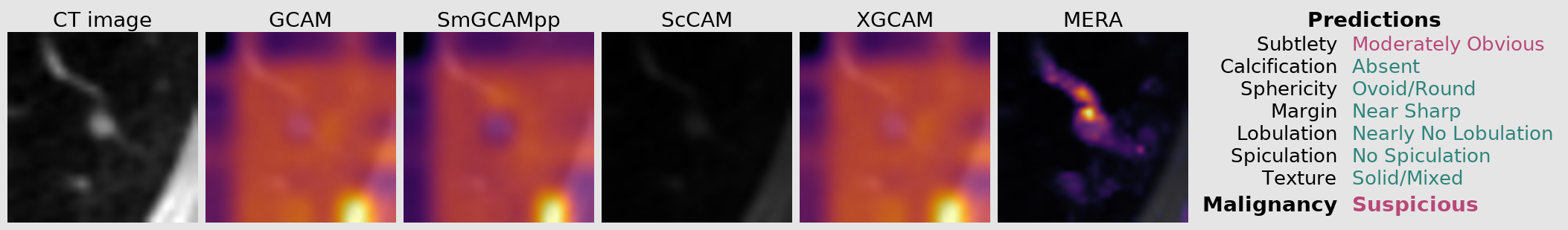} \\[-0.25ex]
    \includegraphics[width=0.9\textwidth, trim=0ex 0ex 0ex 10ex, clip]{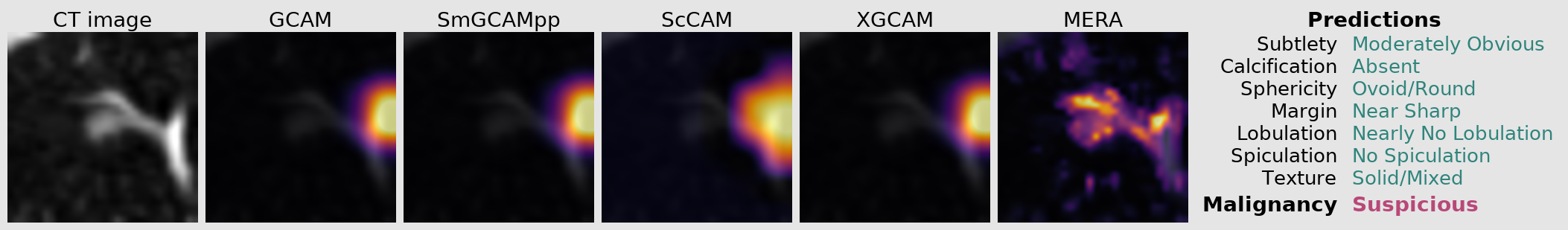} \\[-0.25ex]
    \includegraphics[width=0.9\textwidth, trim=0ex 0ex 0ex 10ex, clip]{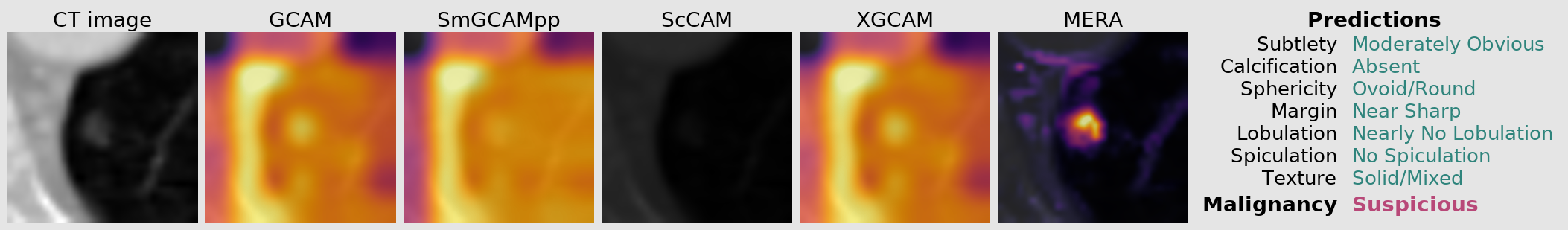} \\[-0.25ex]
    \includegraphics[width=0.9\textwidth, trim=0ex 0ex 0ex 10ex, clip]{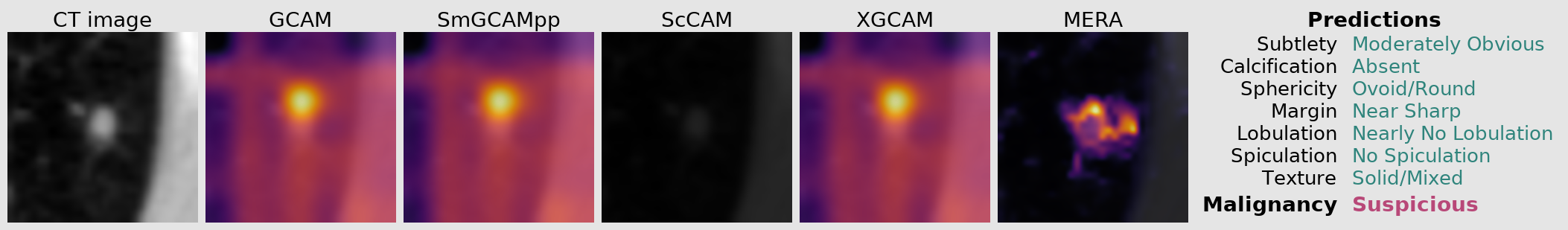} \\[-0.25ex]
    \includegraphics[width=0.9\textwidth, trim=0ex 0ex 0ex 10ex, clip]{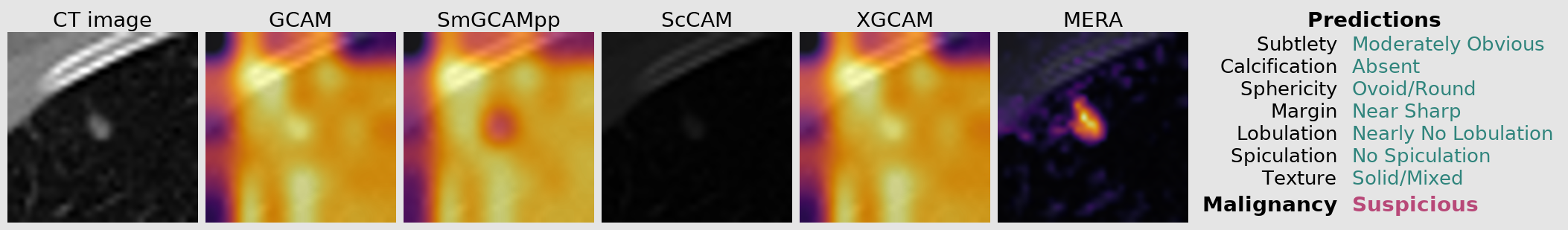} \\[-0.25ex]
    \includegraphics[width=0.9\textwidth, trim=0ex 0ex 0ex 10ex, clip]{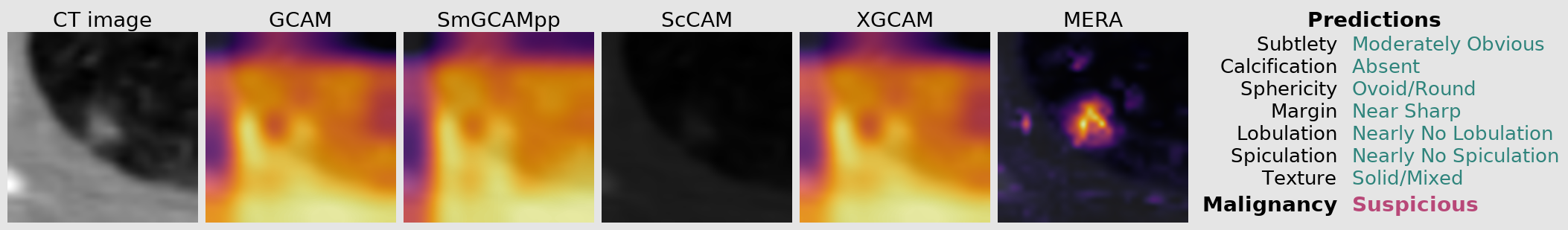} \\[-0.25ex]
    \includegraphics[width=0.9\textwidth, trim=0ex 0ex 0ex 10ex, clip]{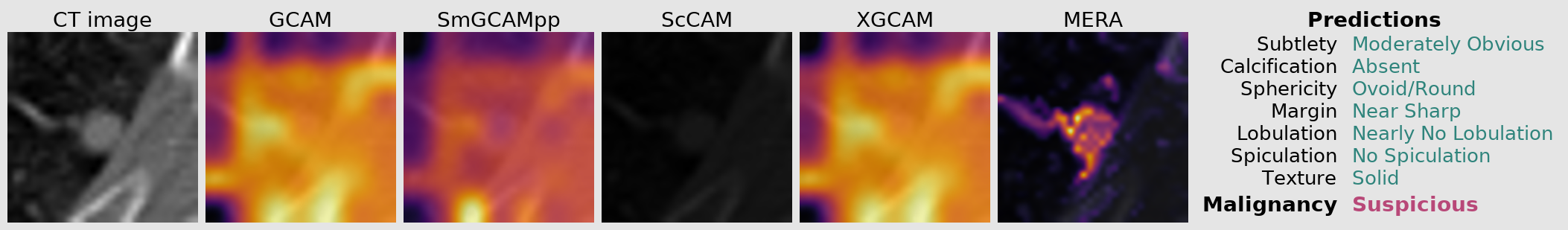}

    \caption{Samples of \textbf{false positive} malignancy predictions with local visual explanation and concept explanation \textbf{(not cherry-picked)}. 
    From left to right: original lung nodule image patch on an axial chest CT, visual explanation of 4 competitor methods, visual explanation of our proposed method, our predicted nodule attributes and malignancy (green font indicates correct and red font indicates wrong, and only $1\%$ annotated data is used for the predictions).
    }
    \label{fig:res_FP1}
\end{figure*}

\begin{figure*}[p]
    \centering
    \includegraphics[width=0.9\textwidth]{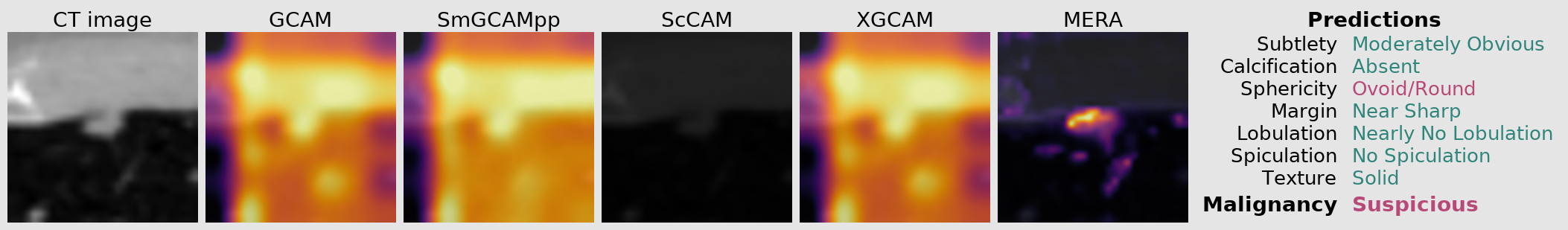} \\[-0.25ex]
    \includegraphics[width=0.9\textwidth, trim=0ex 0ex 0ex 10ex, clip]{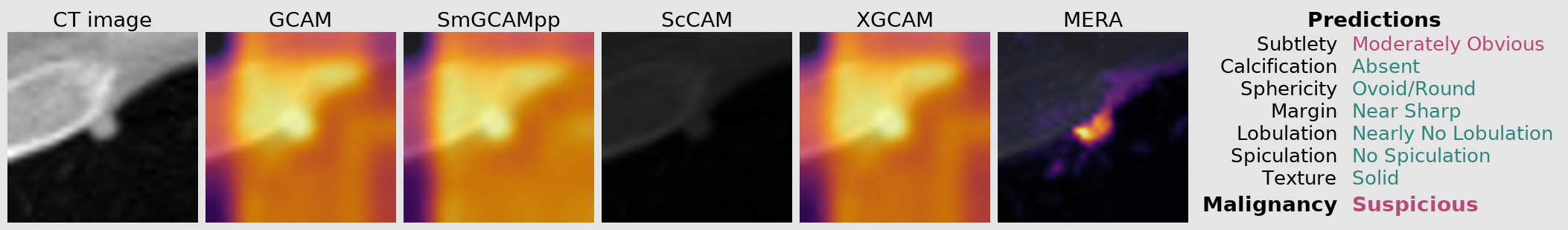} \\[-0.25ex]
    \includegraphics[width=0.9\textwidth, trim=0ex 0ex 0ex 10ex, clip]{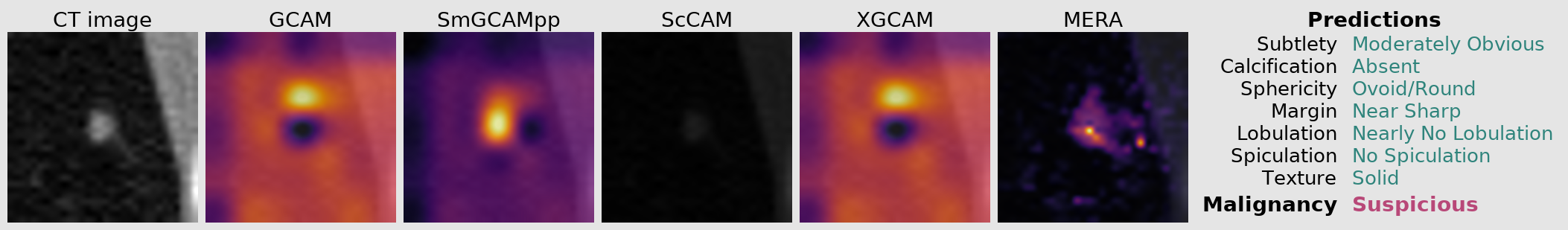} \\[-0.25ex]
    \includegraphics[width=0.9\textwidth, trim=0ex 0ex 0ex 10ex, clip]{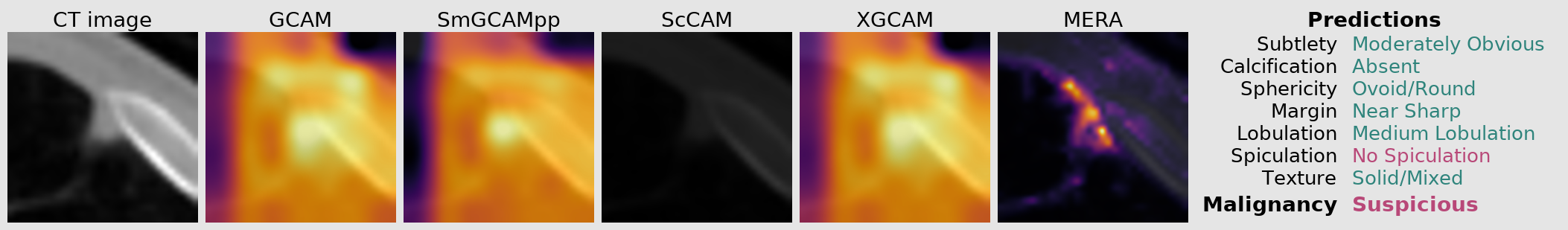} \\[-0.25ex]
    \includegraphics[width=0.9\textwidth, trim=0ex 0ex 0ex 10ex, clip]{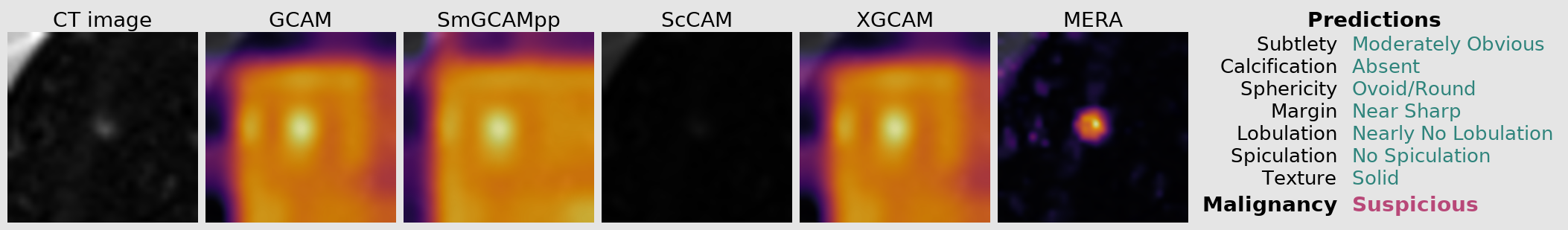} \\[-0.25ex]
    \includegraphics[width=0.9\textwidth, trim=0ex 0ex 0ex 10ex, clip]{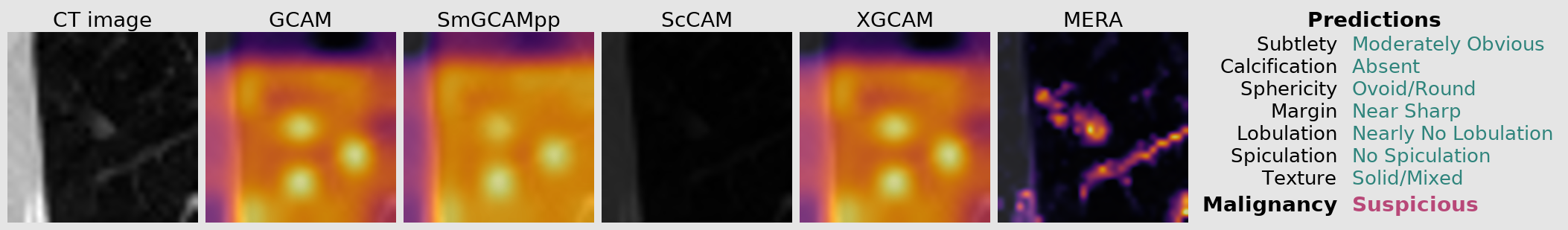} \\[-0.25ex]
    \includegraphics[width=0.9\textwidth, trim=0ex 0ex 0ex 10ex, clip]{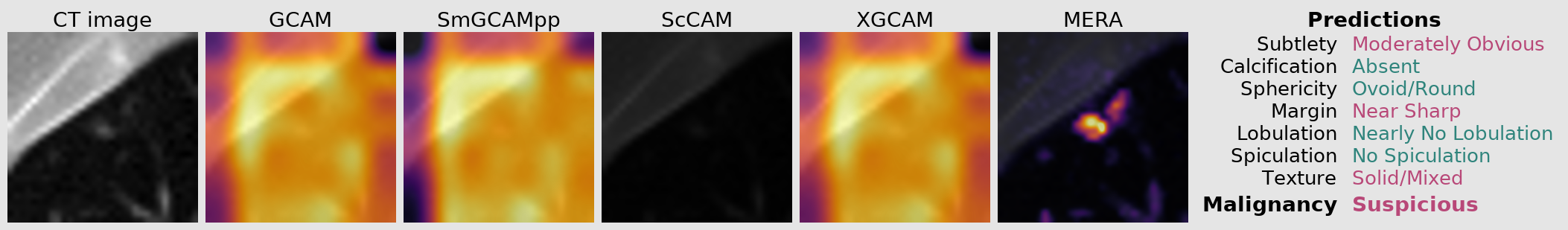} \\[-0.25ex]
    \includegraphics[width=0.9\textwidth, trim=0ex 0ex 0ex 10ex, clip]{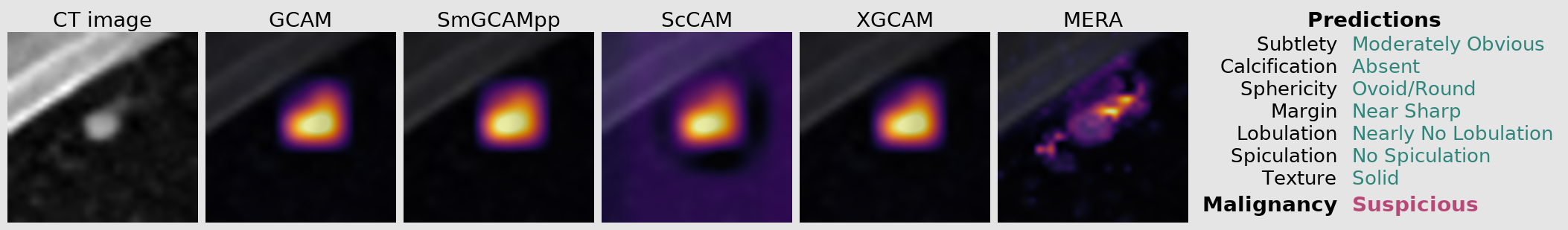} \\[-0.25ex]
    \includegraphics[width=0.9\textwidth, trim=0ex 0ex 0ex 10ex, clip]{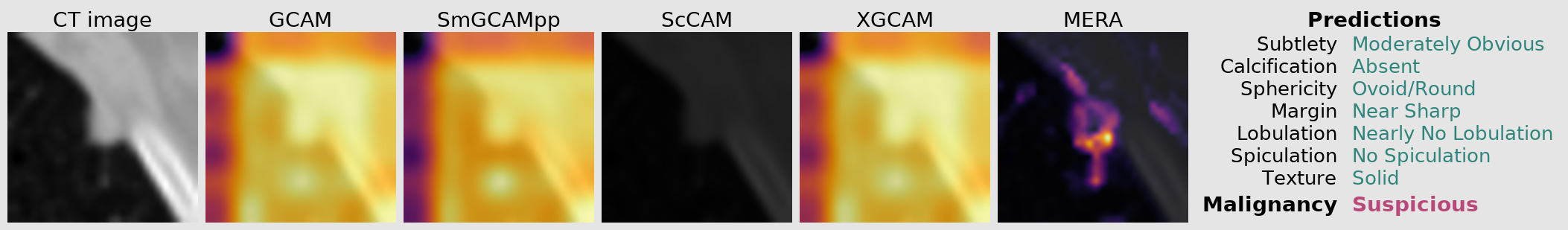} \\[-0.25ex]
    \includegraphics[width=0.9\textwidth, trim=0ex 0ex 0ex 10ex, clip]{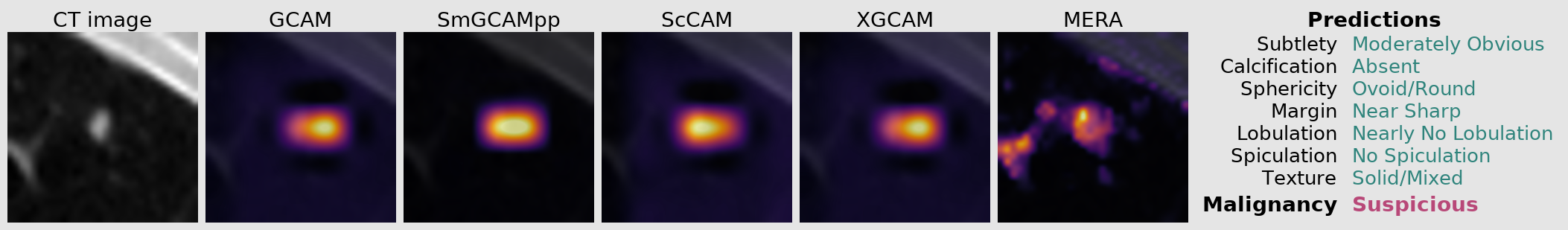}

    \caption{Samples of \textbf{false positive} malignancy predictions with local visual explanation and concept explanation \textbf{(not cherry-picked)}. 
    From left to right: original lung nodule image patch on an axial chest CT, visual explanation of 4 competitor methods, visual explanation of our proposed method, our predicted nodule attributes and malignancy (green font indicates correct and red font indicates wrong, and only $1\%$ annotated data is used for the predictions).
    }
    \label{fig:res_FP2}
\end{figure*}

\begin{figure*}[p]
    \centering
    \includegraphics[width=0.9\textwidth]{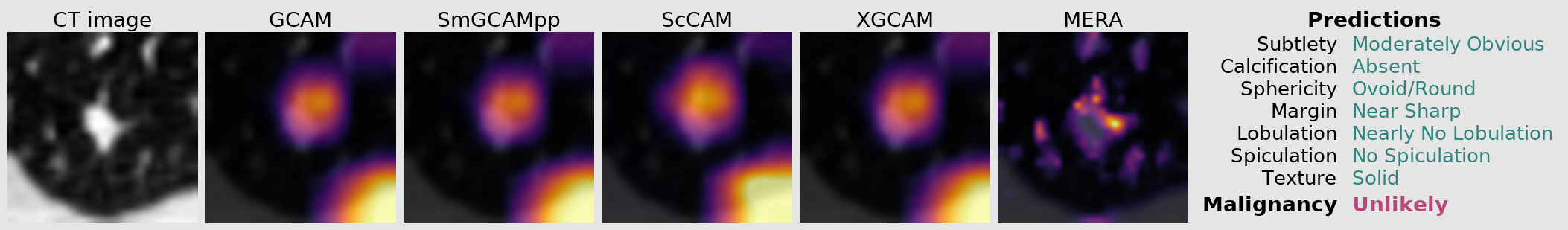} \\[-0.25ex]
    \includegraphics[width=0.9\textwidth, trim=0ex 0ex 0ex 10ex, clip]{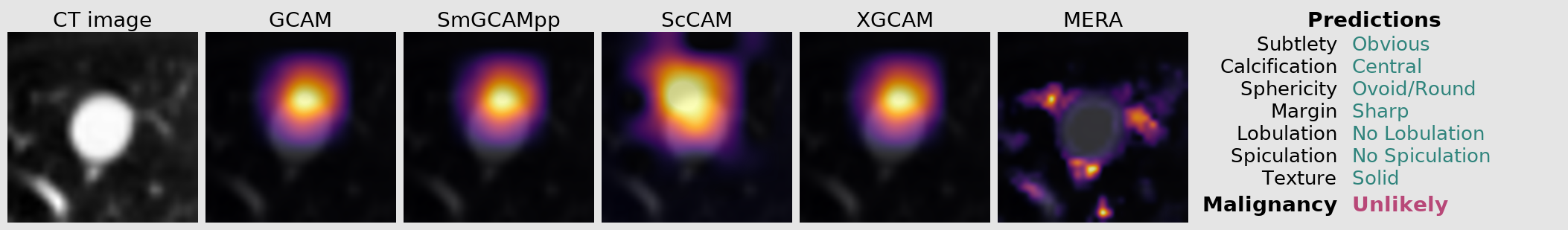} \\[-0.25ex]
    \includegraphics[width=0.9\textwidth, trim=0ex 0ex 0ex 10ex, clip]{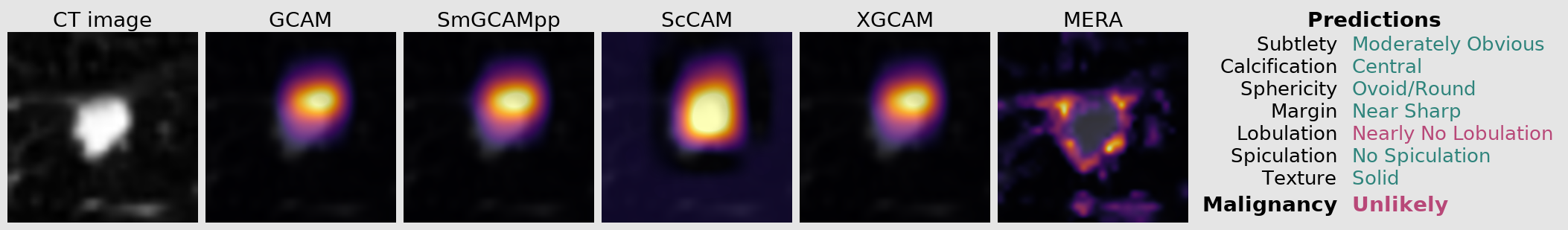} \\[-0.25ex]
    \includegraphics[width=0.9\textwidth, trim=0ex 0ex 0ex 10ex, clip]{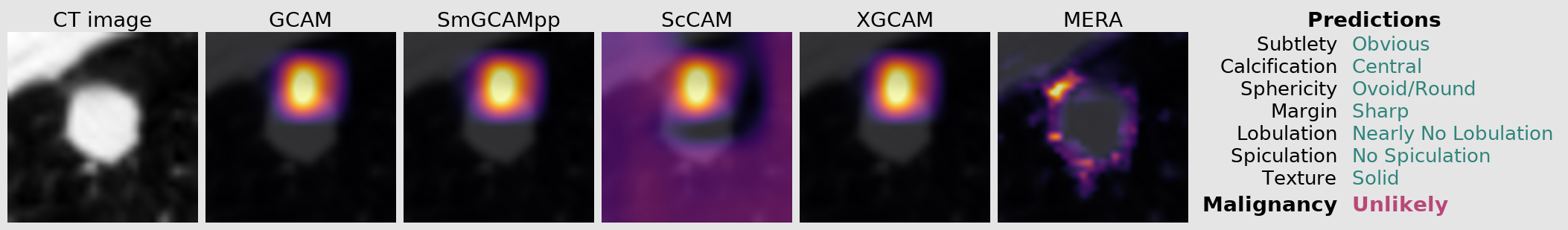} \\[-0.25ex]
    \includegraphics[width=0.9\textwidth, trim=0ex 0ex 0ex 10ex, clip]{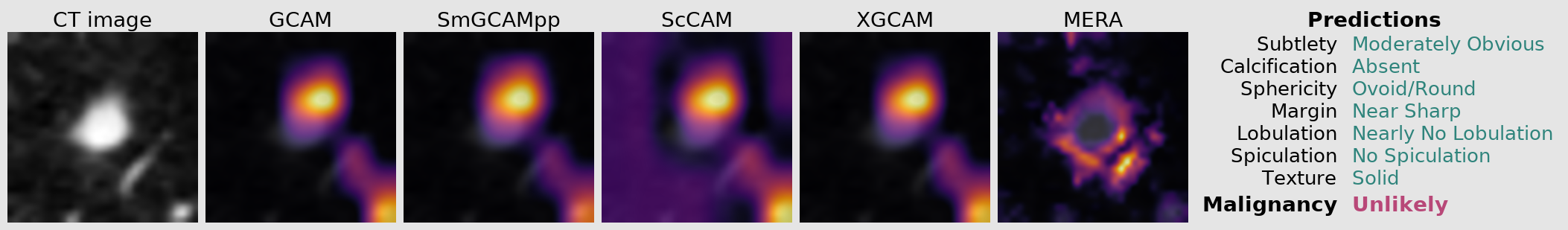} \\[-0.25ex]
    \includegraphics[width=0.9\textwidth, trim=0ex 0ex 0ex 10ex, clip]{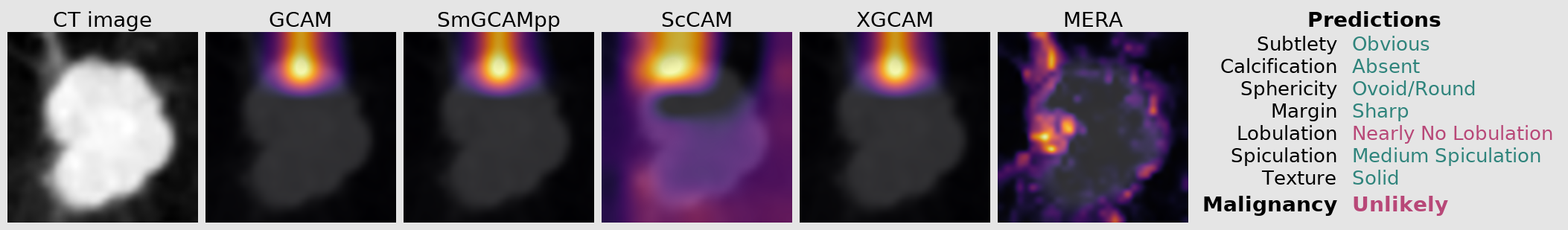} \\[-0.25ex]
    \includegraphics[width=0.9\textwidth, trim=0ex 0ex 0ex 10ex, clip]{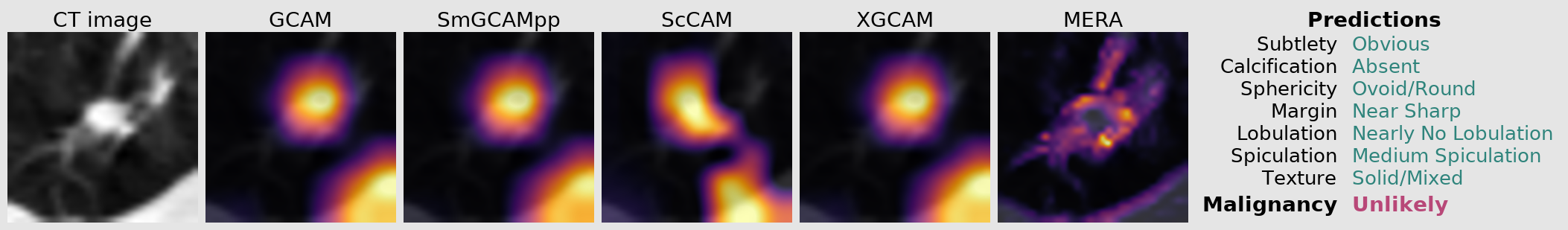} \\[-0.25ex]
    \includegraphics[width=0.9\textwidth, trim=0ex 0ex 0ex 10ex, clip]{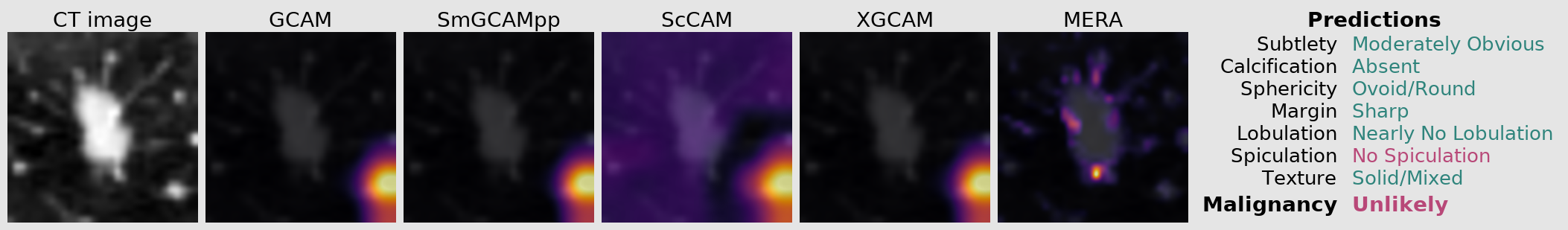} \\[-0.25ex]
    \includegraphics[width=0.9\textwidth, trim=0ex 0ex 0ex 10ex, clip]{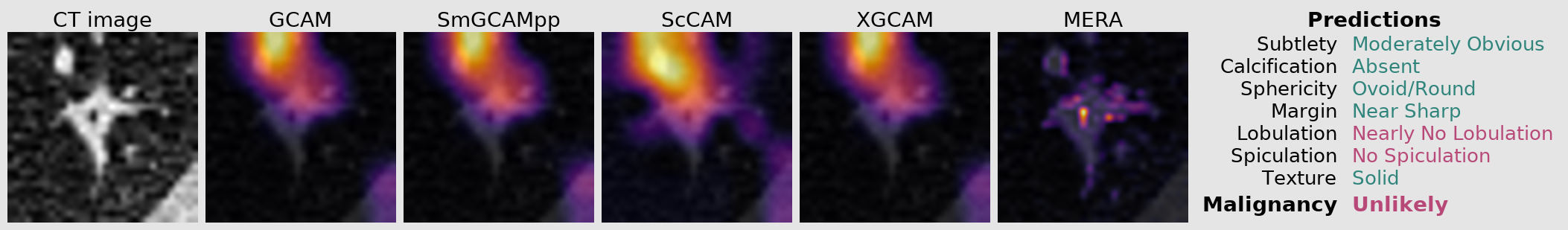} \\[-0.25ex]
    \includegraphics[width=0.9\textwidth, trim=0ex 0ex 0ex 10ex, clip]{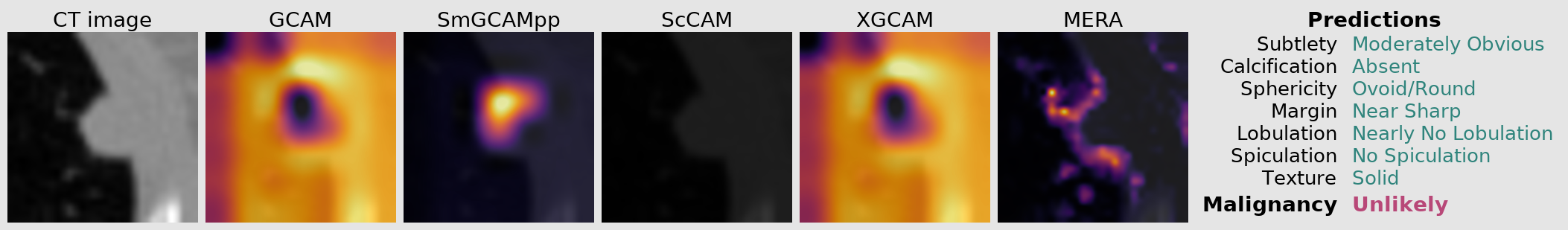}

    \caption{Samples of \textbf{false negative} malignancy predictions with local visual explanation and concept explanation \textbf{(not cherry-picked)}. 
    From left to right: original lung nodule image patch on an axial chest CT, visual explanation of 4 competitor methods, visual explanation of our proposed method, our predicted nodule attributes and malignancy (green font indicates correct and red font indicates wrong, and only $1\%$ annotated data is used for the predictions).
    }
    \label{fig:res_FN1}
\end{figure*}

\begin{figure*}[p]
    \centering
    \includegraphics[width=0.9\textwidth]{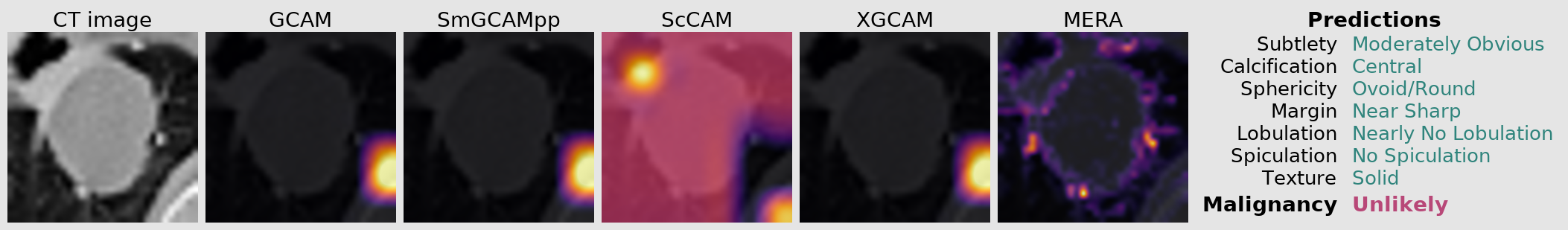} \\[-0.25ex]
    \includegraphics[width=0.9\textwidth, trim=0ex 0ex 0ex 10ex, clip]{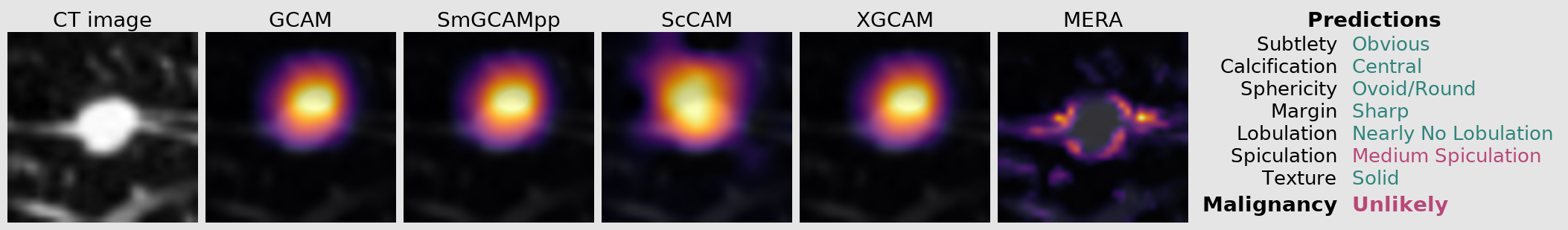} \\[-0.25ex]
    \includegraphics[width=0.9\textwidth, trim=0ex 0ex 0ex 10ex, clip]{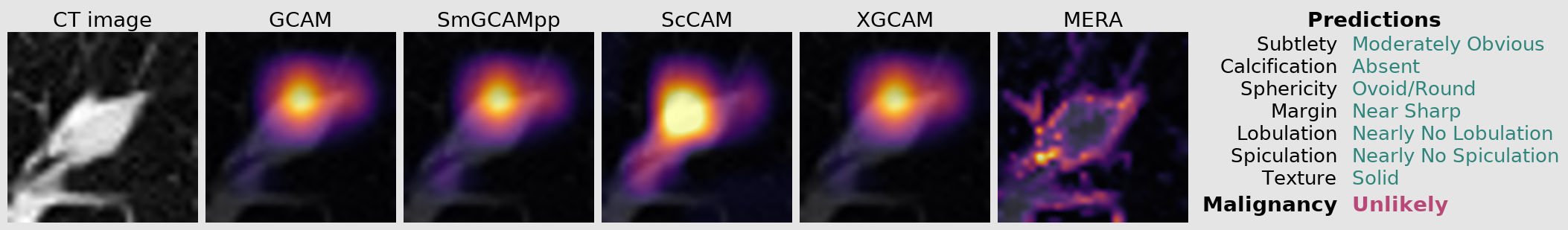} \\[-0.25ex]
    \includegraphics[width=0.9\textwidth, trim=0ex 0ex 0ex 10ex, clip]{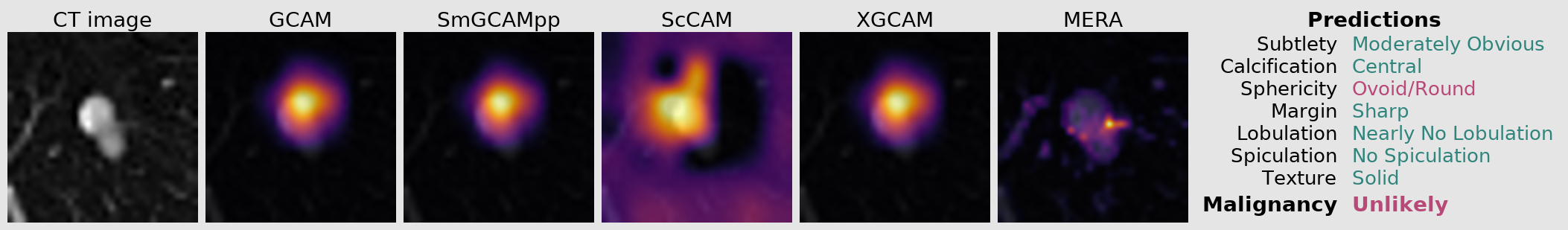} \\[-0.25ex]
    \includegraphics[width=0.9\textwidth, trim=0ex 0ex 0ex 10ex, clip]{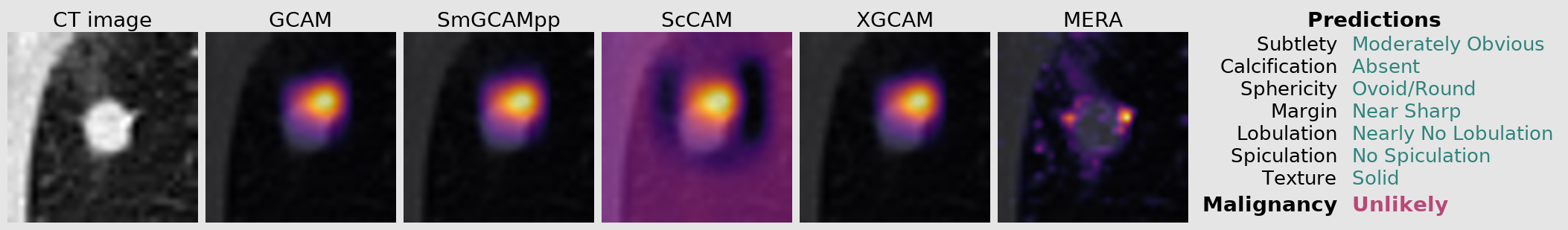} \\[-0.25ex]
    \includegraphics[width=0.9\textwidth, trim=0ex 0ex 0ex 10ex, clip]{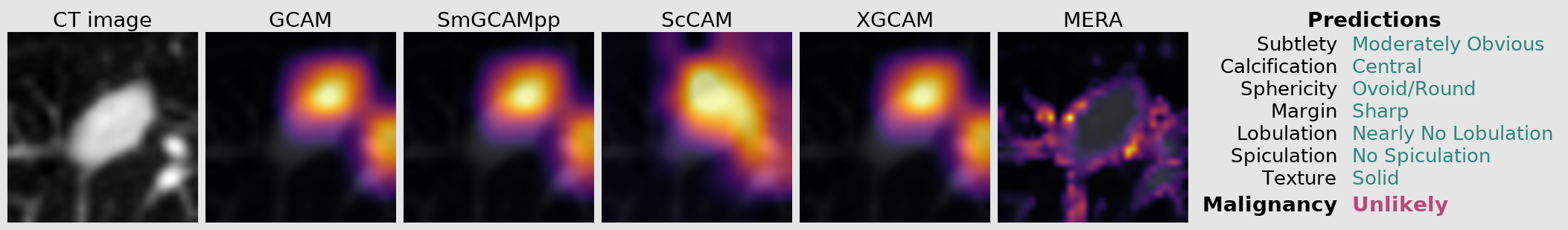} \\[-0.25ex]
    \includegraphics[width=0.9\textwidth, trim=0ex 0ex 0ex 10ex, clip]{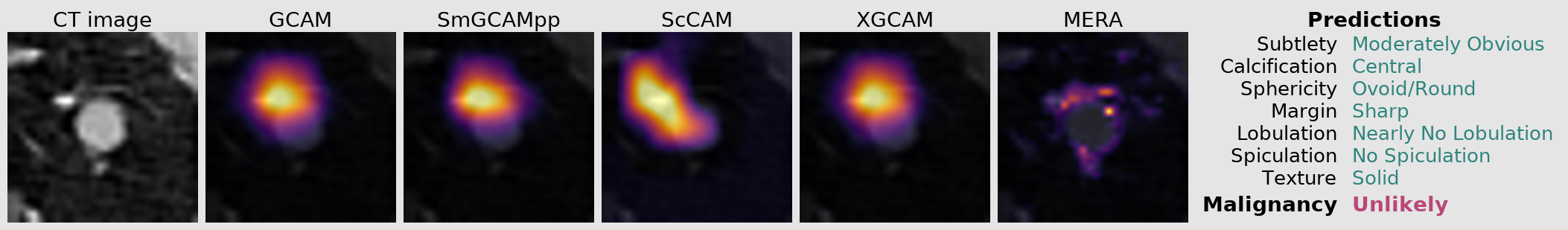} \\[-0.25ex]
    \includegraphics[width=0.9\textwidth, trim=0ex 0ex 0ex 10ex, clip]{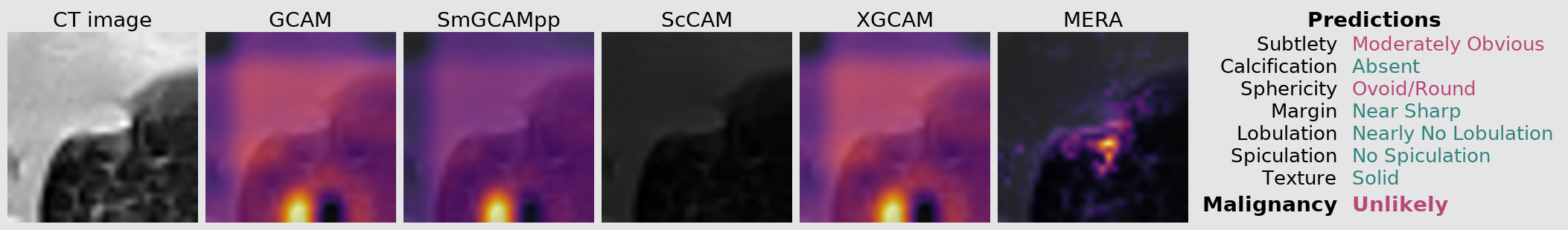} \\[-0.25ex]
    \includegraphics[width=0.9\textwidth, trim=0ex 0ex 0ex 10ex, clip]{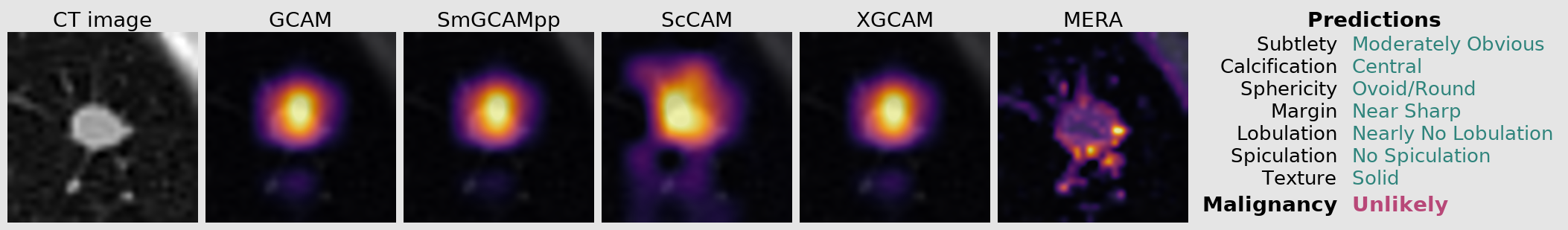} \\[-0.25ex]
    \includegraphics[width=0.9\textwidth, trim=0ex 0ex 0ex 10ex, clip]{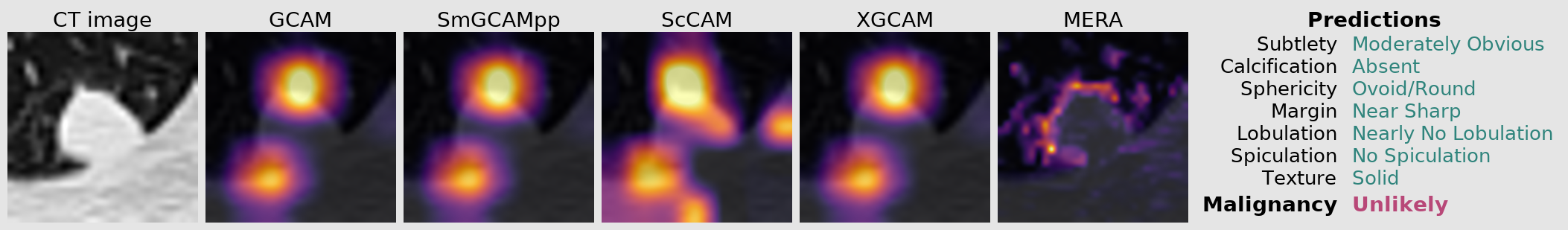}

    \caption{Samples of \textbf{false negative} malignancy predictions with local visual explanation and concept explanation \textbf{(not cherry-picked)}. 
    From left to right: original lung nodule image patch on an axial chest CT, visual explanation of 4 competitor methods, visual explanation of our proposed method, our predicted nodule attributes and malignancy (green font indicates correct and red font indicates wrong, and only $1\%$ annotated data is used for the predictions).
    }
    \label{fig:res_FN2}
\end{figure*}

\begin{figure*}[p]
    \centering
    \includegraphics[width=0.9\textwidth]{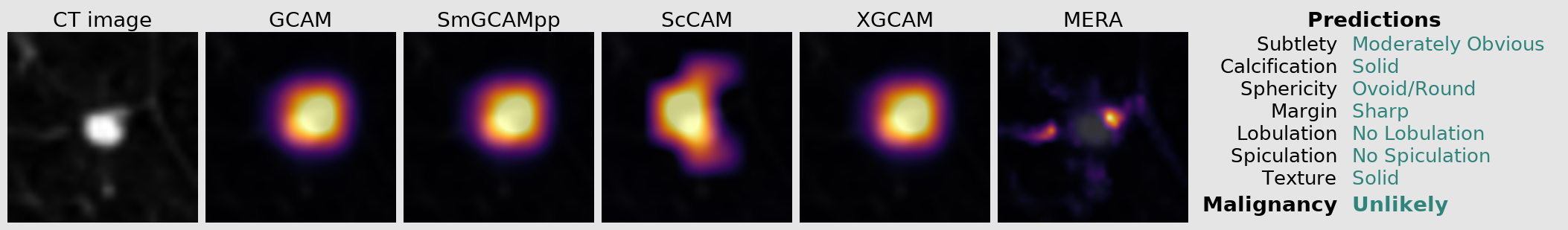} \\[-0.25ex]
    \includegraphics[width=0.9\textwidth, trim=0ex 0ex 0ex 10ex, clip]{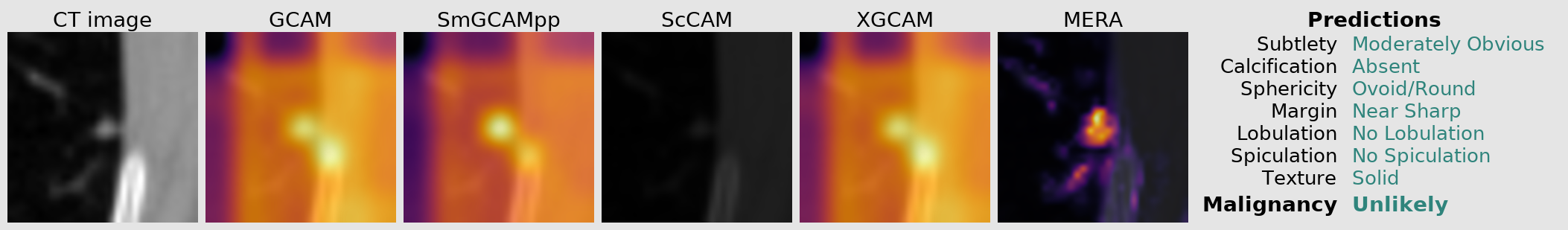} \\[-0.25ex]
    \includegraphics[width=0.9\textwidth, trim=0ex 0ex 0ex 10ex, clip]{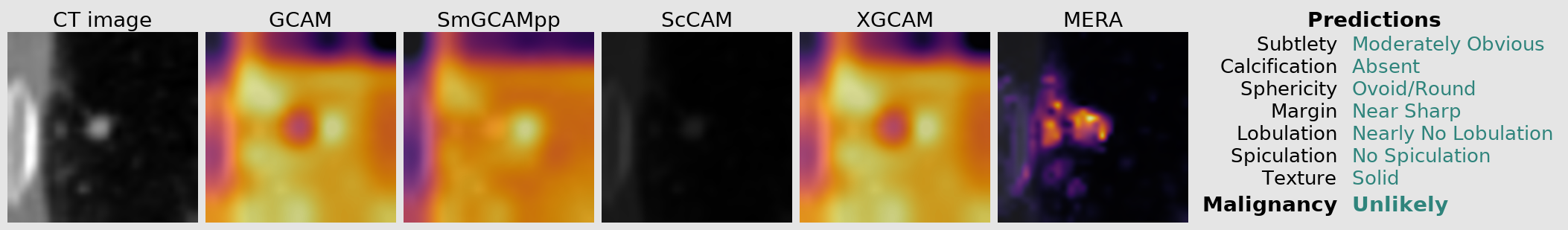} \\[-0.25ex]
    \includegraphics[width=0.9\textwidth, trim=0ex 0ex 0ex 10ex, clip]{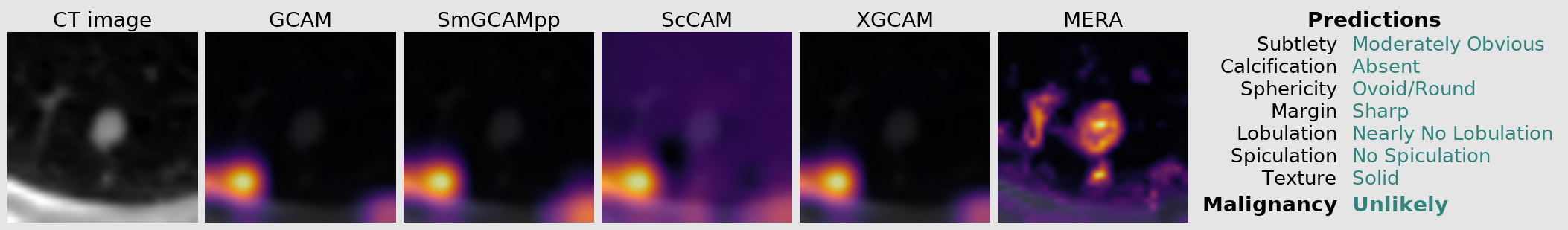} \\[-0.25ex]
    \includegraphics[width=0.9\textwidth, trim=0ex 0ex 0ex 10ex, clip]{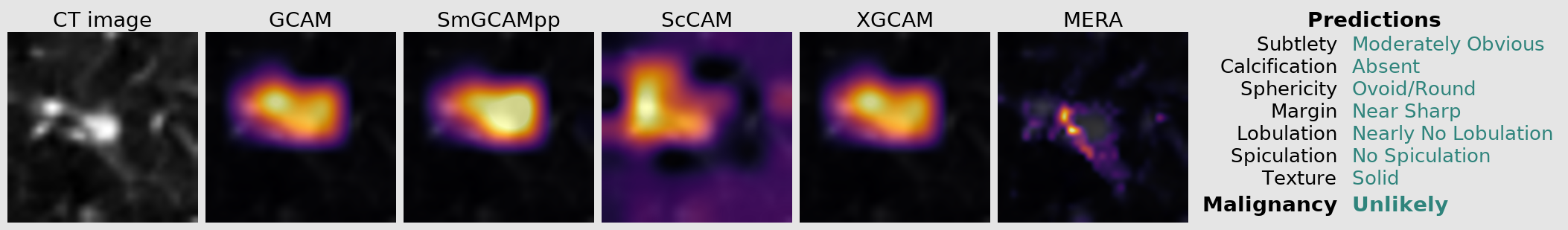} \\[-0.25ex]
    \includegraphics[width=0.9\textwidth, trim=0ex 0ex 0ex 10ex, clip]{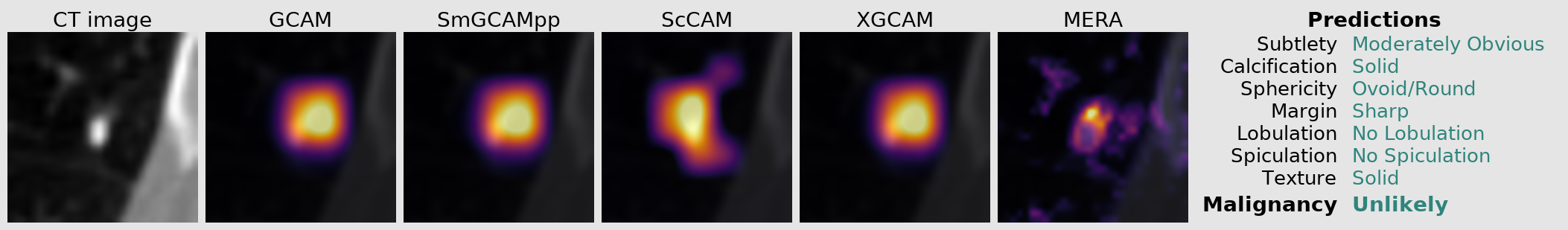} \\[-0.25ex]
    \includegraphics[width=0.9\textwidth, trim=0ex 0ex 0ex 10ex, clip]{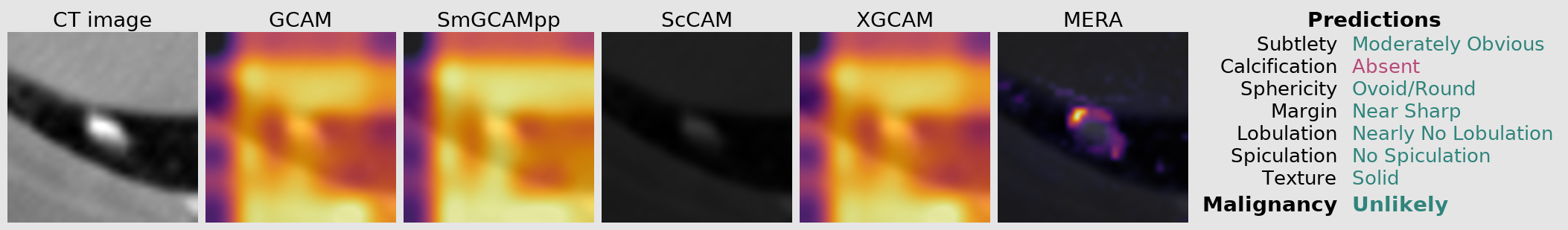} \\[-0.25ex]
    \includegraphics[width=0.9\textwidth, trim=0ex 0ex 0ex 10ex, clip]{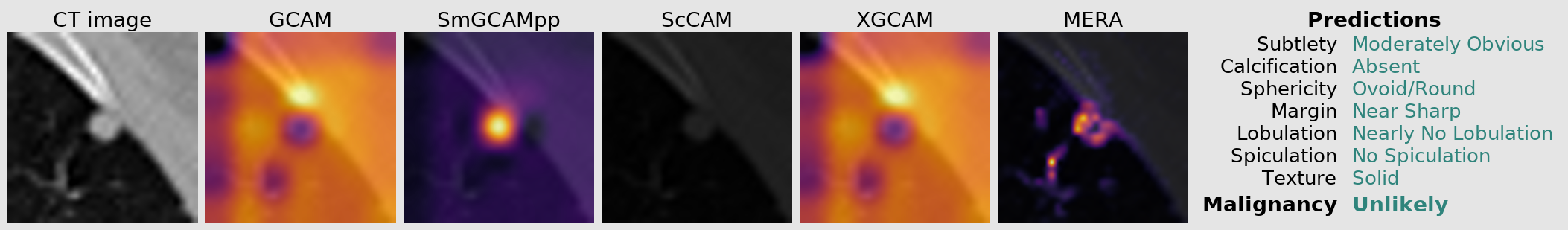} \\[-0.25ex]
    \includegraphics[width=0.9\textwidth, trim=0ex 0ex 0ex 10ex, clip]{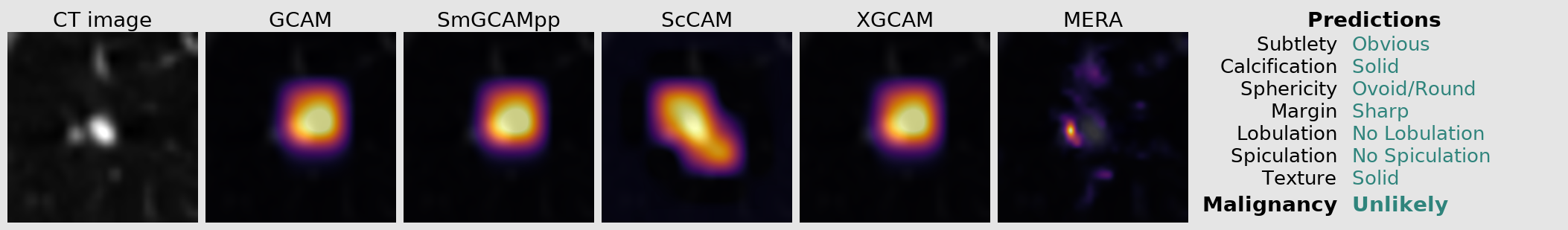} \\[-0.25ex]
    \includegraphics[width=0.9\textwidth, trim=0ex 0ex 0ex 10ex, clip]{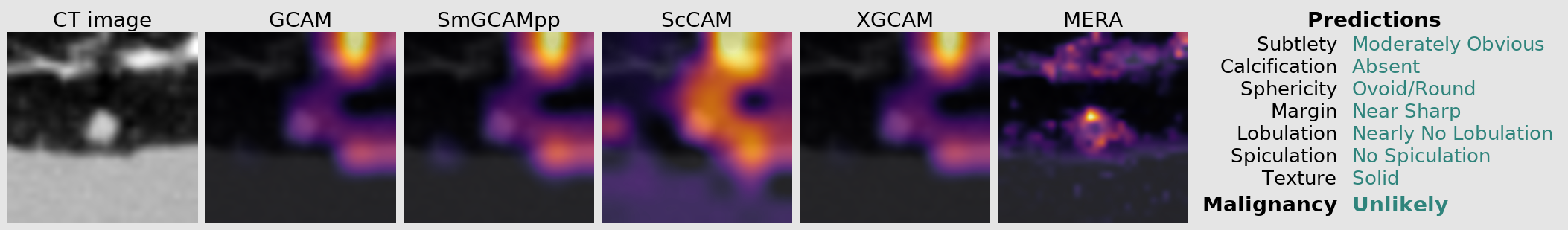}

    \caption{Samples of \textbf{true negative} malignancy predictions with local visual explanation and concept explanation \textbf{(not cherry-picked)}. 
    From left to right: original lung nodule image patch on an axial chest CT, visual explanation of 4 competitor methods, visual explanation of our proposed method, our predicted nodule attributes and malignancy (green font indicates correct and red font indicates wrong, and only $1\%$ annotated data is used for the predictions).
    }
    \label{fig:res_TN1}
\end{figure*}

\begin{figure*}[p]
    \centering
    \includegraphics[width=0.9\textwidth]{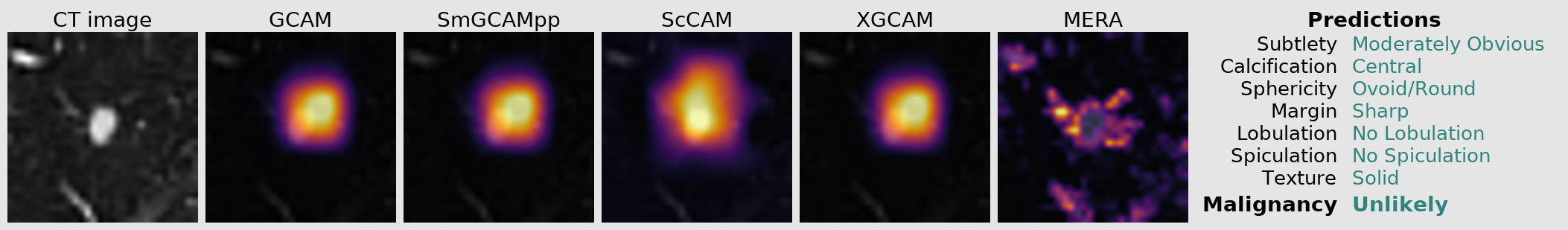} \\[-0.25ex]
    \includegraphics[width=0.9\textwidth, trim=0ex 0ex 0ex 10ex, clip]{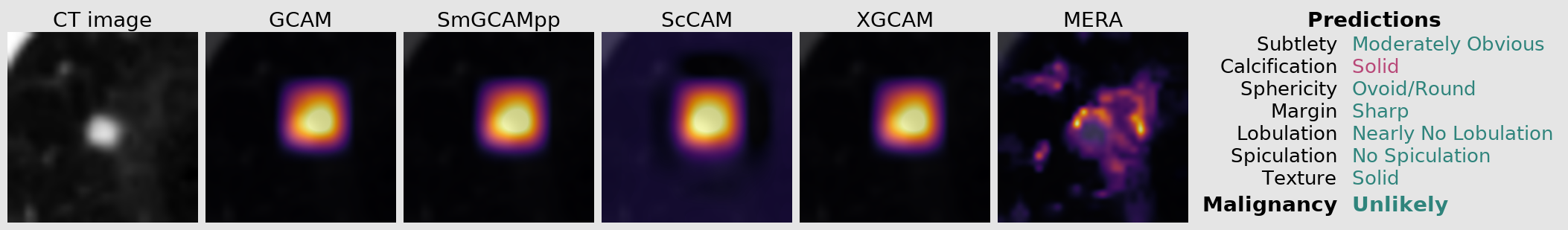} \\[-0.25ex]
    \includegraphics[width=0.9\textwidth, trim=0ex 0ex 0ex 10ex, clip]{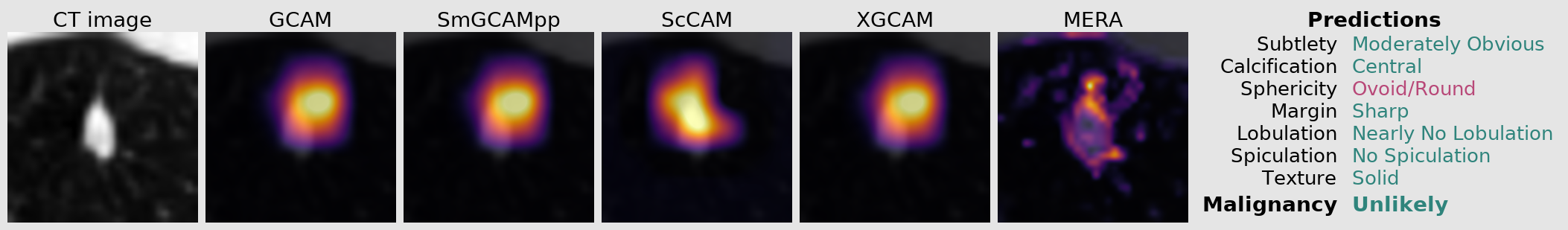} \\[-0.25ex]
    \includegraphics[width=0.9\textwidth, trim=0ex 0ex 0ex 10ex, clip]{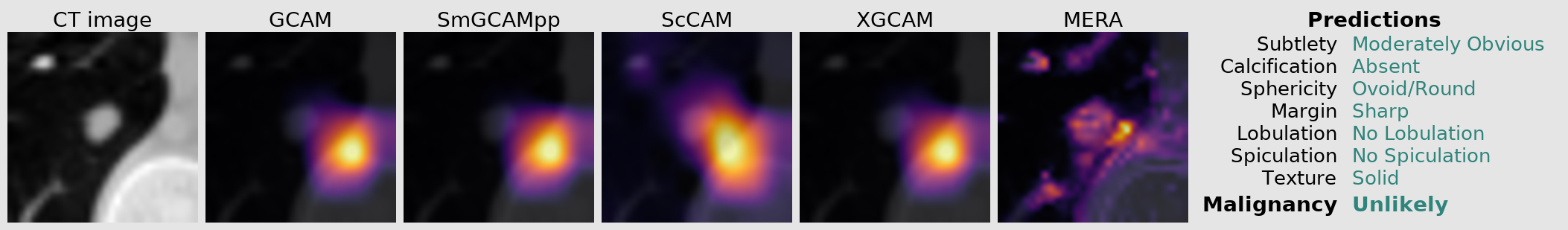} \\[-0.25ex]
    \includegraphics[width=0.9\textwidth, trim=0ex 0ex 0ex 10ex, clip]{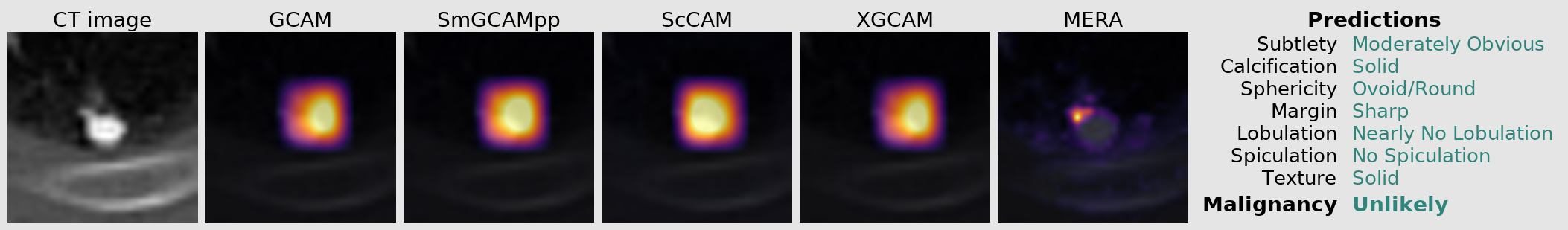} \\[-0.25ex]
    \includegraphics[width=0.9\textwidth, trim=0ex 0ex 0ex 10ex, clip]{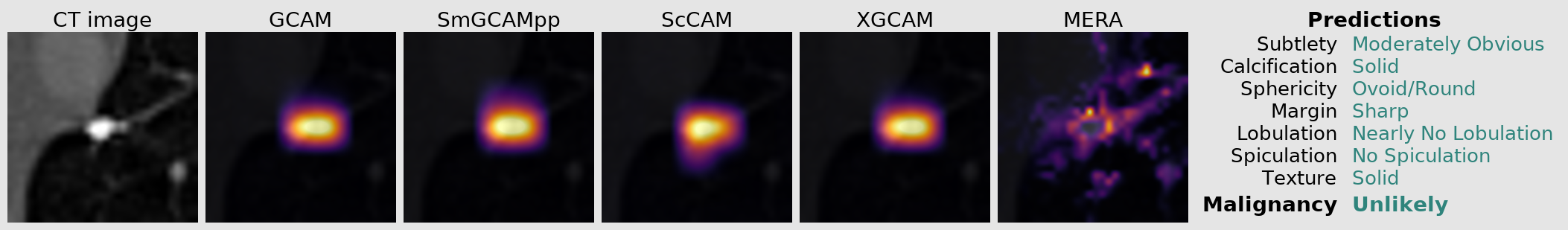} \\[-0.25ex]
    \includegraphics[width=0.9\textwidth, trim=0ex 0ex 0ex 10ex, clip]{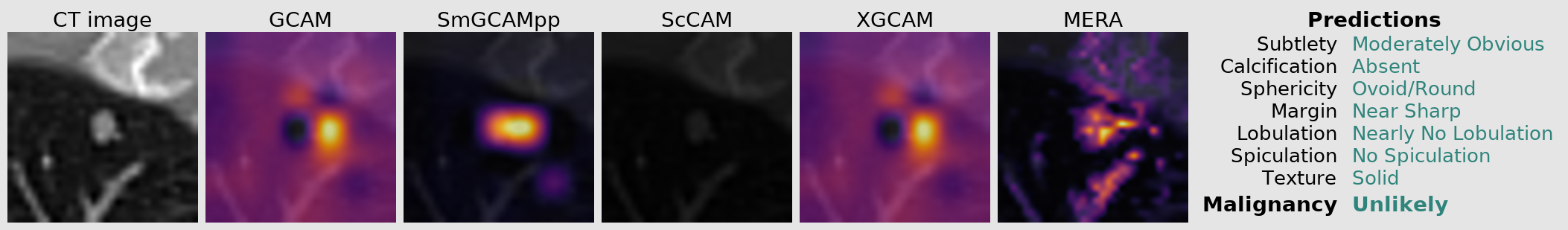} \\[-0.25ex]
    \includegraphics[width=0.9\textwidth, trim=0ex 0ex 0ex 10ex, clip]{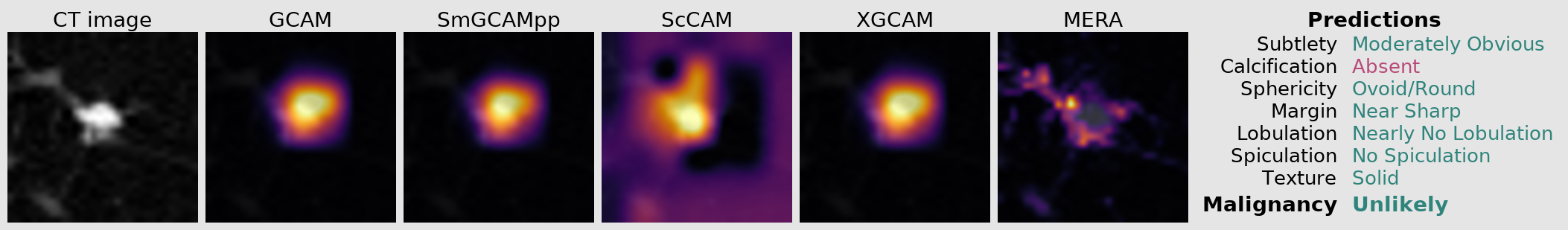} \\[-0.25ex]
    \includegraphics[width=0.9\textwidth, trim=0ex 0ex 0ex 10ex, clip]{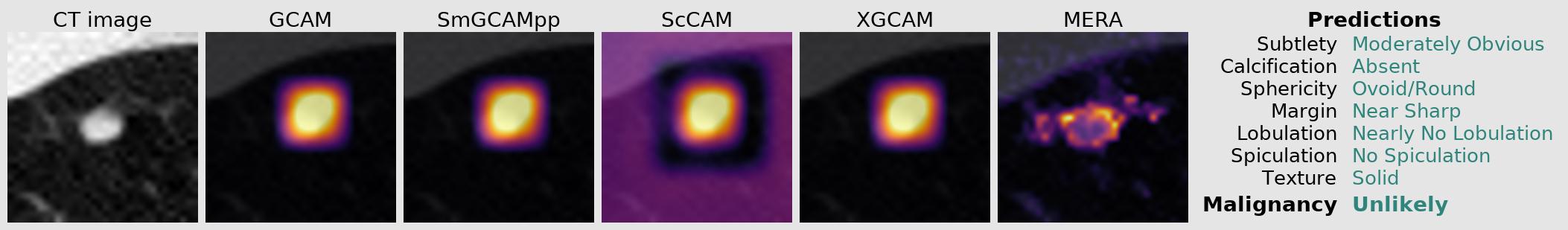} \\[-0.25ex]
    \includegraphics[width=0.9\textwidth, trim=0ex 0ex 0ex 10ex, clip]{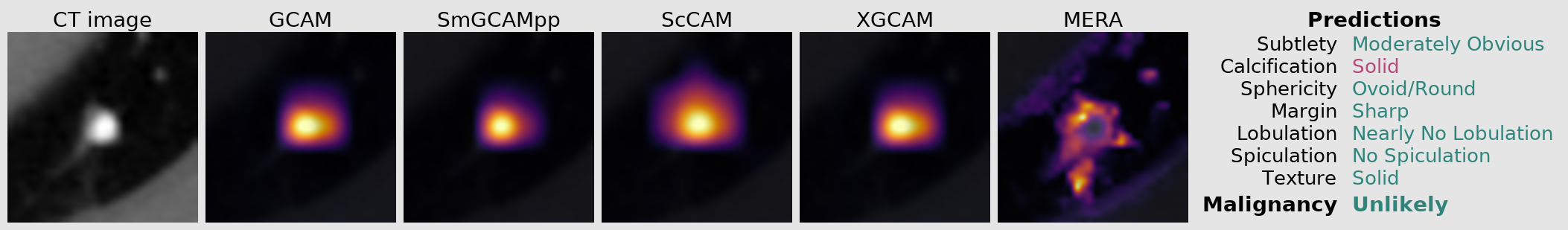}

    \caption{Samples of \textbf{true negative} malignancy predictions with local visual explanation and concept explanation \textbf{(not cherry-picked)}. 
    From left to right: original lung nodule image patch on an axial chest CT, visual explanation of 4 competitor methods, visual explanation of our proposed method, our predicted nodule attributes and malignancy (green font indicates correct and red font indicates wrong, and only $1\%$ annotated data is used for the predictions).
    }
    \label{fig:res_TN2}
\end{figure*}

\end{document}